\documentclass[10pt,twocolumn,letterpaper]{article}

\usepackage[pagenumbers]{cvpr} % To force page numbers, e.g. for an arXiv 
\usepackage[dvipsnames]{xcolor}

\newcommand{\e}[1]{   {\scriptsize\ \textcolor[RGB]{180,180,180}{($\pm$#1)}}     }
\newcommand{\p}[1]{   {\scriptsize\ \textcolor[RGB]{12,180,22}{($+$#1)}}     }
\newcommand{\m}[1]{   {\scriptsize\ \textcolor[RGB]{180,12,22}{($-$#1)}}     }

\newcommand{\du}{360$^{\circ}$}

\newcommand{\cmark}{\ding{51}}
\newcommand{\xmark}{\ding{55}}

\newcommand{\cleanfootnote}[1]{ \let\thefootnote\relax\footnotetext{#1} }

\usepackage{graphicx}
\usepackage{booktabs}
\usepackage{amssymb}   % http://ctan.org/pkg/amssymb
\usepackage{pifont}    % http://ctan.org/pkg/pifont
\usepackage{wrapfig}
\usepackage{makecell}
\usepackage{amsmath}
\usepackage{bm}
\usepackage{multirow}

\usepackage{colortbl}
\usepackage{xcolor}

\usepackage{floatrow}
\usepackage{caption}
\usepackage{subcaption}
\usepackage{caption}
\usepackage[dvipsnames]{xcolor}
\usepackage{amssymb}   % http://ctan.org/pkg/amssymb
\usepackage{pifont}    % http://ctan.org/pkg/pifont
\usepackage{wrapfig}
\usepackage{makecell}
\usepackage{amsmath}
\usepackage{bm}
\usepackage{multirow}
\usepackage{scalerel}  % Scale in Height

\usepackage{cuted}   % Teaser
\usepackage{capt-of}
\usepackage[accsupp]{axessibility}

\newcommand{\customlabel}[1]{%
\protected@write \@auxout {}{\newlabel {#1}{{}{}}}}
\makeatother

\newcommand{\verticalphotos}[5]{%
{
    \begin{figure*}[t]  % htbp
    \vspace{-1.25cm}
        \centering
        \begin{subfigure}[b]{0.79\textwidth}
            \centering
            \includegraphics[width=\textwidth]{#1}
            \caption{\du panoramic video frame example}
        \end{subfigure}
        
        \vspace{0.2em}
        
        \begin{subfigure}[b]{0.79\textwidth}
            \includegraphics[width=\textwidth]{#2}
            \caption{Third-person front view video frame example}
        \end{subfigure}
        
        \vspace{0.2em}
        
        \begin{subfigure}[b]{0.79\textwidth}
            \includegraphics[width=\textwidth]{#3}
            \caption{Binocular video frame example}
        \end{subfigure}

        \vspace{0.2em}
        \begin{subfigure}[b]{0.79\textwidth}
            \includegraphics[width=\textwidth, trim=0 10 0 23, clip]{#4}
            \caption{Stereo Waveform Difference Figure}
        \end{subfigure}
        
        \vspace{-0.6em}
        \caption{Frame examples in the category of #5}
    \end{figure*}
}
}

\newcommand{\verticalphotosHead}[5]{%
{
    \begin{figure*}[t]  % htbp
    \vspace{-1.25cm}
        \centering
        \begin{subfigure}[b]{0.79\textwidth}
            \centering
            \includegraphics[width=\textwidth]{#1}
            \caption{\du panoramic video frame example}
        \end{subfigure}
        
        \vspace{0.2em}
        
        \begin{subfigure}[b]{0.79\textwidth}
            \includegraphics[width=\textwidth]{#2}
            \caption{Third-person front view video frame example}
        \end{subfigure}
        
        \vspace{0.2em}
        
        \begin{subfigure}[b]{0.79\textwidth}
            \includegraphics[width=\textwidth]{#3}
            \caption{Binocular video frame example}
        \end{subfigure}

        \vspace{0.2em}
        \begin{subfigure}[b]{0.79\textwidth}
            \includegraphics[width=\textwidth, trim=0 10 0 23, clip]{#4}
            \caption{Stereo Waveform Difference Figure}
        \end{subfigure}
        
        \vspace{-0.6em}
        \caption{Frame examples in the category of #5}
        \label{fig:head}
    \end{figure*}
}
}

\newcommand{\verticalphotosEnd}[5]{%
{
    \begin{figure*}[t]  % htbp
    \vspace{-1.25cm}
        \centering
        \begin{subfigure}[b]{0.79\textwidth}
            \centering
            \includegraphics[width=\textwidth]{#1}
            \caption{\du panoramic video frame example}
        \end{subfigure}
        
        \vspace{0.2em}
        
        \begin{subfigure}[b]{0.79\textwidth}
            \includegraphics[width=\textwidth]{#2}
            \caption{Third-person front view video frame example}
        \end{subfigure}
        
        \vspace{0.2em}
        
        \begin{subfigure}[b]{0.79\textwidth}
            \includegraphics[width=\textwidth]{#3}
            \caption{Binocular video frame example}
        \end{subfigure}

        \vspace{0.2em}
        \begin{subfigure}[b]{0.79\textwidth}
            \includegraphics[width=\textwidth, trim=0 10 0 23, clip]{#4}
            \caption{Stereo Waveform Difference Figure}
        \end{subfigure}
        
        \vspace{-0.6em}
        \caption{Frame examples in the category of #5}
        \label{fig:end}
    \end{figure*}
}
}

\definecolor{cvprblue}{rgb}{0.21,0.49,0.74}
\usepackage[accsupp]{axessibility} 
\usepackage{url}
\usepackage[pagebackref]{hyperref}
\hypersetup{
	final,
	breaklinks,
	colorlinks,
	linkcolor={Mahogany},
	citecolor={cvprblue},
	urlcolor={CadetBlue}
}
\usepackage{pbalance}

\usepackage[labelformat=simple,labelsep=period]{subcaption}  
\def\confName{CVPR}
\def\confYear{2024}

\title{\textit{360+x}: A Panoptic Multi-modal Scene Understanding Dataset}

\author{Hao Chen\hspace{4mm} Yuqi Hou\hspace{4mm}  Chenyuan Qu\hspace{4mm} Irene Testini\hspace{4mm}
Xiaohan Hong\hspace{4mm} Jianbo Jiao\\[2mm]
The \href{https://mix.jianbojiao.com/}{Machine Intelligence + \textit{x}} Group, 
University of Birmingham, UK\hspace{4mm} \\[2mm]
 Project page: \href{https://x360dataset.github.io/}{https://x360dataset.github.io/}
}

\begin{document}
\maketitle

\begin{strip}
\centering
\vspace{-1.5cm}
\vstretch{1.1}{\includegraphics[width=\textwidth]{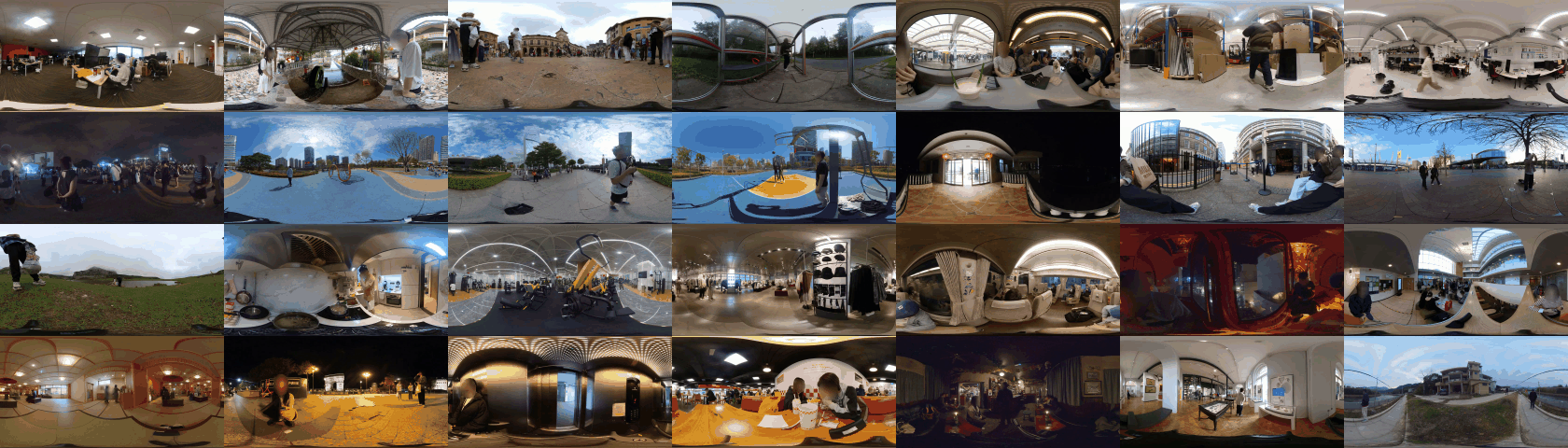}}
\captionof{figure}{Example \du panoramics videos from all 28 scene categories.}
\label{fig:teasor}
\vspace{1mm}
\end{strip}

\begin{abstract}
% 
% To answer Why doing that
Human perception of the world is shaped by a multitude of viewpoints and modalities. While many existing datasets focus on scene understanding from a certain perspective (e.g. egocentric or third-person views), our dataset offers a panoptic perspective (i.e. multiple viewpoints with multiple data modalities). Specifically, we encapsulate third-person panoramic and front views, as well as egocentric monocular/binocular views with rich modalities including video, multi-channel audio, directional binaural delay, location data and textual scene descriptions within each scene captured, presenting comprehensive observation of the world. Figure \ref{fig:teasor} offers a glimpse of all 28 scene categories of our \textit{360+x} dataset.  To the best of our knowledge, this is the first database that covers multiple viewpoints with multiple data modalities to mimic how daily information is accessed in the real world. 
Through our benchmark analysis, we presented 5 different scene understanding tasks on the proposed 360+x dataset to evaluate the impact and benefit of each data modality and perspective in panoptic scene understanding.  We hope this unique dataset could broaden the scope of comprehensive scene understanding and encourage the community to approach these problems from more diverse perspectives.

% categories (Figure \ref{fig:teasor}) and

% Extensive experimental analysis reveals the effectiveness of each data modality and perspective 

% \let\thefootnote\relax\footnotetext{ Project page: \href{https://x360dataset.github.io/}{https://x360dataset.github.io/}} 

% g.  .e impact and benefit of each individual data
% 017 modality and viewpoints on various visual-audio tasks

% presented 

% For an overview of the dataset, please refer to: \href{https://x360dataset.github.io/}{https://x360dataset.github.io/}.

% A benchmark analysis is presented over this dataset to evaluate the impact of individual data modality and viewpoint on the performance of models trained on them. This analysis encompasses several visual-audio scene understanding tasks, including video activity classification, temporal activity localisation and self-supervised representation learning. Extensive experimental analysis shows the effectiveness of each data modality/perspective in contributing to panoptic scene understanding.

% real-world video and audio data from a \du panoramic camera and a stereo glasses-like camera,. 

 % including indoor, outdoor, campus, shopping centres, cafes, train stations, etc

\end{abstract}    
\vspace{-0.1cm}
\section{Introduction}
\label{sec:intro}

Scene understanding is crucial for robotics and artificial intelligent systems to perceive the environment around them. As humans, we intuitively understand the world through primarily visual inputs, as well as auditory and other sensory inputs (\eg touch and smell). The community has made remarkable progress in mimicking human perception with contributions from various datasets and benchmarks~\cite{grauman2022ego4d,damen2018scaling,kay2017kinetics,audioset,kuehne2011hmdb,soomro2012ucf101,caba2015activitynet}. These efforts have approached scene understanding from a diverse range of perspectives, such as normal frontal-view vision~\cite{caba2015activitynet,kay2017kinetics,soomro2012ucf101}, panoramic view~\cite{song2018im2pano3d,xiao2012recognizing},  binocular/stereo view~\cite{scharstein2014high,yang2019drivingstereo}, egocentric monocular view~\cite{grauman2022ego4d,damen2018scaling}, and audio~\cite{audioset,chen2020vggsound}.

While there has been exciting progress in understanding scenes from a limited number of perspectives, it is notable that humans understand the world by incorporating a combination of viewpoints, in a holistic manner. This includes an egocentric view for activities we are involved in and a third-person view for activities we are observing. 
In addition to visual cues, we also rely on a range of modalities, including hearing and binaural delay, to fully comprehend our surroundings and track movements. Our prior knowledge of the scene, such as localisation information and scene descriptions, has also supported our understanding of the environment (\eg the cafe in the city centre may be different from a similar cafe on a university campus). 

\begin{figure*}[!t]
\begin{center}

\includegraphics[width=1\textwidth]{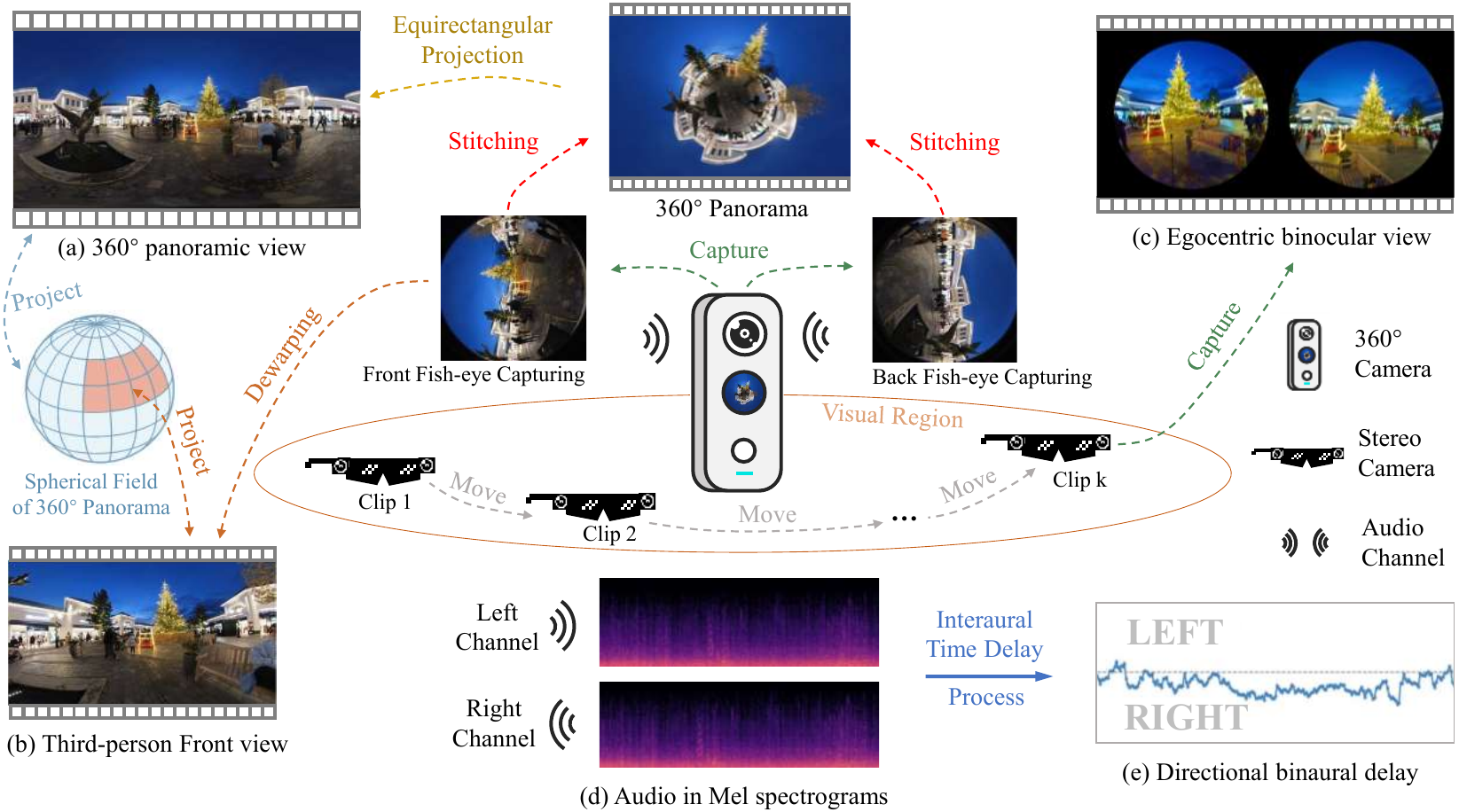}

\end{center}
% \vspace{-2mm}

\caption{\small \textbf{Illustration of the proposed \textit{360+x} dataset.} The \du~camera records fish-eye raw videos with front and back lenses. These videos are merged to create a spherical \du panorama (middle-up figure, zoom in for details), which is then transformed to (a) \du panoramic data using equirectangular projection. The (b) third-person front view is obtained by de-warping the rich movements region highlighted red in the spherical field of \du~panorama (the middle-left figure). By wearing stereo cameras, the capturers record (c) egocentric clips while staying visible to the fixed \du camera (central ellipse). (e) Directional audio time delay data is generated from left and right audio inputs (d) from the \du camera by \textit{interaural time delay} process \cite{chen2022sound}. This helps locate sound sources in the \du~panorama.}

\label{fig:data}

% \vspace{-1em}
\end{figure*}

Taking the above observations into consideration, a new dataset covering all these aforementioned aspects is presented in this work, to provide a panoptic scene understanding, termed \emph{360+x} dataset. This new dataset offers a diverse selection of perspectives, including a \du~panoramic view providing a complete panoptic view of the environment, and a third-person front view that highlights the region of interest that has the most movements in front of the camera. Additionally, we have included egocentric monocular and binocular videos to capture the first-person perspective of individuals in the environment. These viewpoints are complemented by aligned multi-channel audio with directional binaural delay information, as well as location information and scene descriptions as metadata. 
An illustration of the presented dataset collection system is shown in Figure~\ref{fig:data}.

% \end{wrapfigure}

Based on this newly collected dataset, we perform 5 visual-audio scene understanding tasks to analyse the contribution and effectiveness of each data viewpoint and modality. Particularly, we look at video classification, temporal action localisation, self-supervised representation learning, cross-modality retrieval and pre-training model migration for dataset adaptation, with interesting findings and insights from extensive experimental analysis.
The main contributions of this work are summarised as follows:
\begin{itemize}
    \item We propose to our knowledge the first and probably the most authentic panoptic scene understanding dataset covering multiple viewpoints and data modalities \textit{in the wild}.
    \item We perform extensive experimental analysis to validate the effectiveness of the proposed dataset on different tasks from various perspectives and modalities.
    \item Interesting findings are derived from the analysis, suggesting the effectiveness of each viewpoint and data modality. Learning from this new dataset without supervision even shows a better performance than that from a model trained in a supervised manner.
\end{itemize}

% \rv{\du panoramic video and paired front videos example clips are shown in Figure \ref{fig:teasor}.}

\section{Related Works}
\label{sec:related}

% \vspace{-3mm}  action recognition and 
\paragraph{Video understanding and analysis.}
% \vspace{-3mm}
Video analysis has been extensively studied in the literature. Existing datasets such as UCF101~\cite{soomro2012ucf101}, ActivityNet~\cite{caba2015activitynet} and Kinetics~\cite{kay2017kinetics} have provided large-scale video data for activity understanding tasks. However, these datasets often exhibit lower complexity compared to real-world scenes. Some datasets, like MultiThumos~\cite{yeung2018every}, aim to increase complexity but are limited to specific scenarios with domain-specific actions, deviating from real-life daily activities. In contrast, our dataset builds upon the activity labels from ActivityNet~\cite{caba2015activitynet} and strives to capture data that closely simulates real-life scenarios.
Apart from that, we also include multiple data viewpoints and modalities as compared to existing datasets.

\vspace{-3mm}
\paragraph{Panoramic scene understanding.}
In recent years, panoramic scene understanding has gained significant attention due to its holistic reflection of the environment. Several datasets have been introduced to facilitate research in this area. For instance, the KITTI-360~\cite{liao2022kitti} provides a collection of panoramic images for urban scene analysis. EGOK360~\cite{bhandari2020egok360} has been introduced to address the need for video data with a panoramic view. Im2Pano3D~\cite{song2018im2pano3d} presents a panoramic dataset for indoor scenarios with semantic segmentation and focuses on the prediction from a partial observation. However, these datasets primarily focus on panoramic visual data while lacking the incorporation of other viewpoints (\eg egocentric) and data modalities (\eg audio), limiting their potential for comprehensive scene understanding and analysis.

\vspace{-3mm}
\paragraph{Egocentric video analysis.}
Focusing on understanding scenes from a first-person perspective, existing datasets such as EPIC-Kitchens~\cite{damen2018scaling} and Ego4D~\cite{grauman2022ego4d} provide egocentric video data collected during daily activities. They have contributed to research on activity recognition and object detection in egocentric scenes. Unlike these datasets focusing on egocentric views, our dataset also covers other viewpoints and modalities aiming at supporting scene understanding research in a more panoptic manner.

\vspace{-3mm}
\paragraph{Visual-audio analysis.}
Integrating visual and audio information often enhances the performance of models in scene understanding tasks, as it provides richer contextual information. There are some existing datasets available to support research in audio-visual analysis, \eg AVA~\cite{gu2018ava}, AudioSet~\cite{7952261} and VGGSound~\cite{chen2020vggsound}, to name a few. However, these datasets are lacking in multiple viewpoints and the directional property of audio signals, which are provided in the proposed new dataset.

% By introducing the 360+X dataset, we address the limitations of existing datasets by incorporating multiple data modalities and viewpoints, mimicking the access to daily information in the real world.

\section{\textit{360+x} Dataset}
% \vspace{-3mm}
\label{collection}
\subsection{Data Acquisition and Alignment}
% \vspace{-3mm}
Two main devices were used for our data collection: 
the \textit{Insta 360 One X2} and \textit{Snapchat Spectacles 3} cameras. The \textit{360 One X2} has two fish-eye cameras that collect $360^{\circ}$ panoramic visual information in the scene with $5760\times2880$ resolution and a frame rate of 25 FPS. Additionally, directional audio was recorded using four microphones in directional audio mode. While the \textit{Spectacles 3} has a stereo camera attached to a pair of glasses used to capture the egocentric binocular vision within the scene at a resolution of $2432\times1216$ and a frame rate of 60 FPS.

\begin{figure*}
    \centering 
   % L D R U
  \begin{subfigure}[b]{0.33\textwidth}   % 0.385
     \hspace{-0.9cm}
     \includegraphics[width=1.25\textwidth]{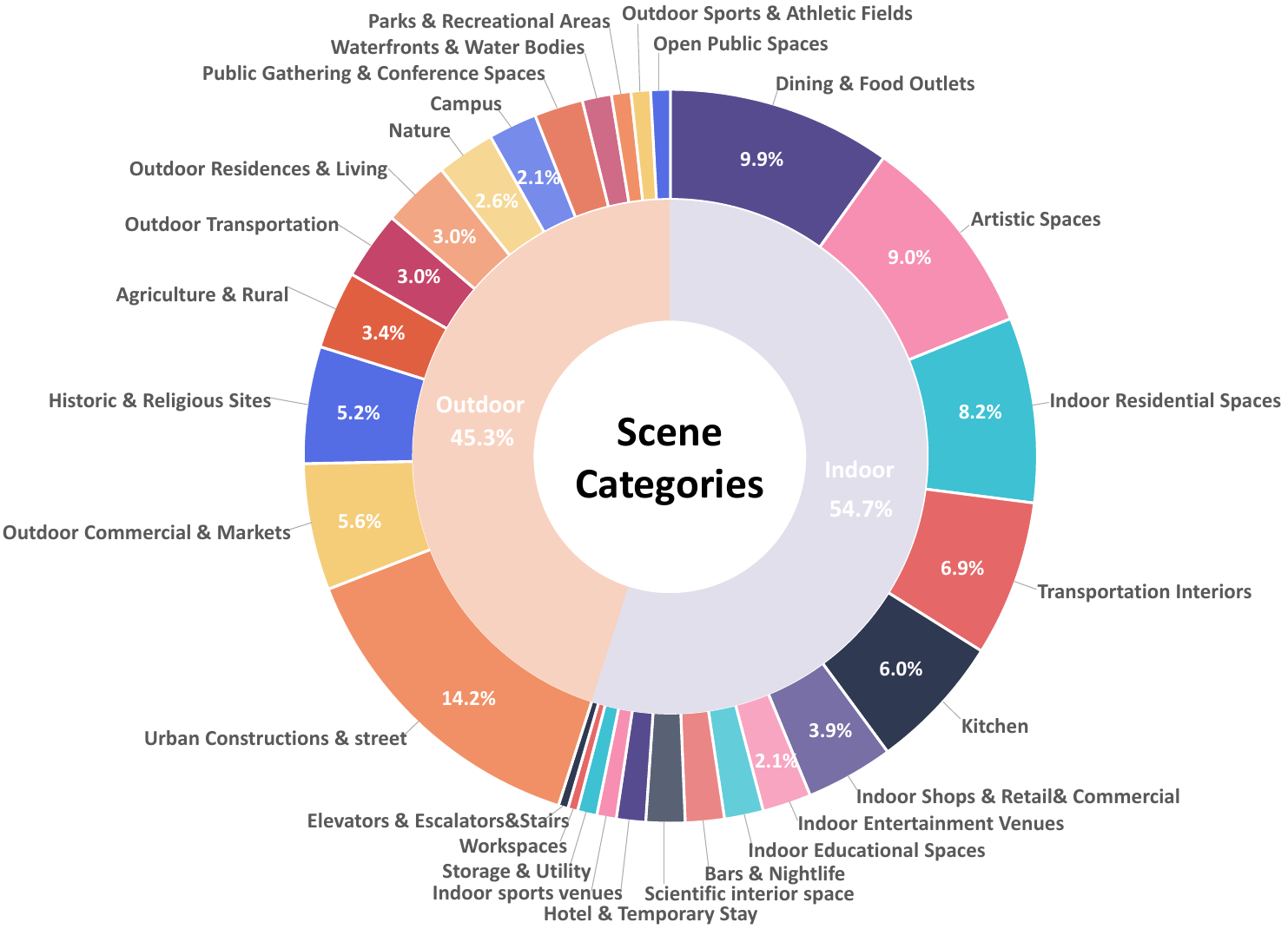}
    % \hspace{-0.3cm}
    \caption{Distribution of the scene categories (number).}
    \label{fig:categories}
  \end{subfigure}
% --------------------------
 \hfill
    \begin{subfigure}[b]{0.33\textwidth}  % 25   0.27
    \vspace*{-20cm}
    % \vspace*{-2cm}
      % \begin{minipage}{\textwidth}
        \includegraphics[width=1.1\textwidth, clip]{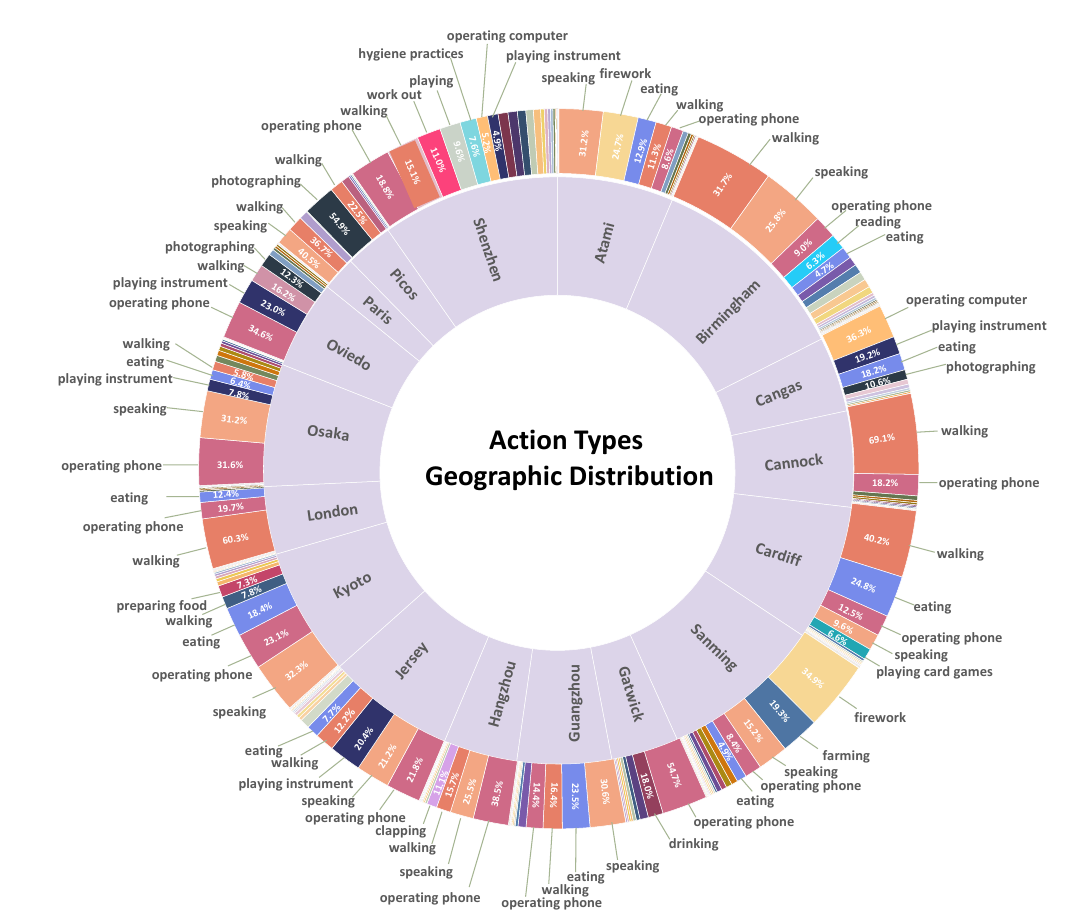}
            \centering 
\caption{Geographical distribution of actions.}
    \label{fig:geometry}
  \end{subfigure}
% ----------------------------------------
  \hfill
  \begin{subfigure}[b]{0.33\textwidth}   % 34
   % \vspace{-5cm}
   \hspace{0.5cm}
   % L D R U
    \includegraphics[width=0.97\textwidth, trim=10 25 88 27, clip]{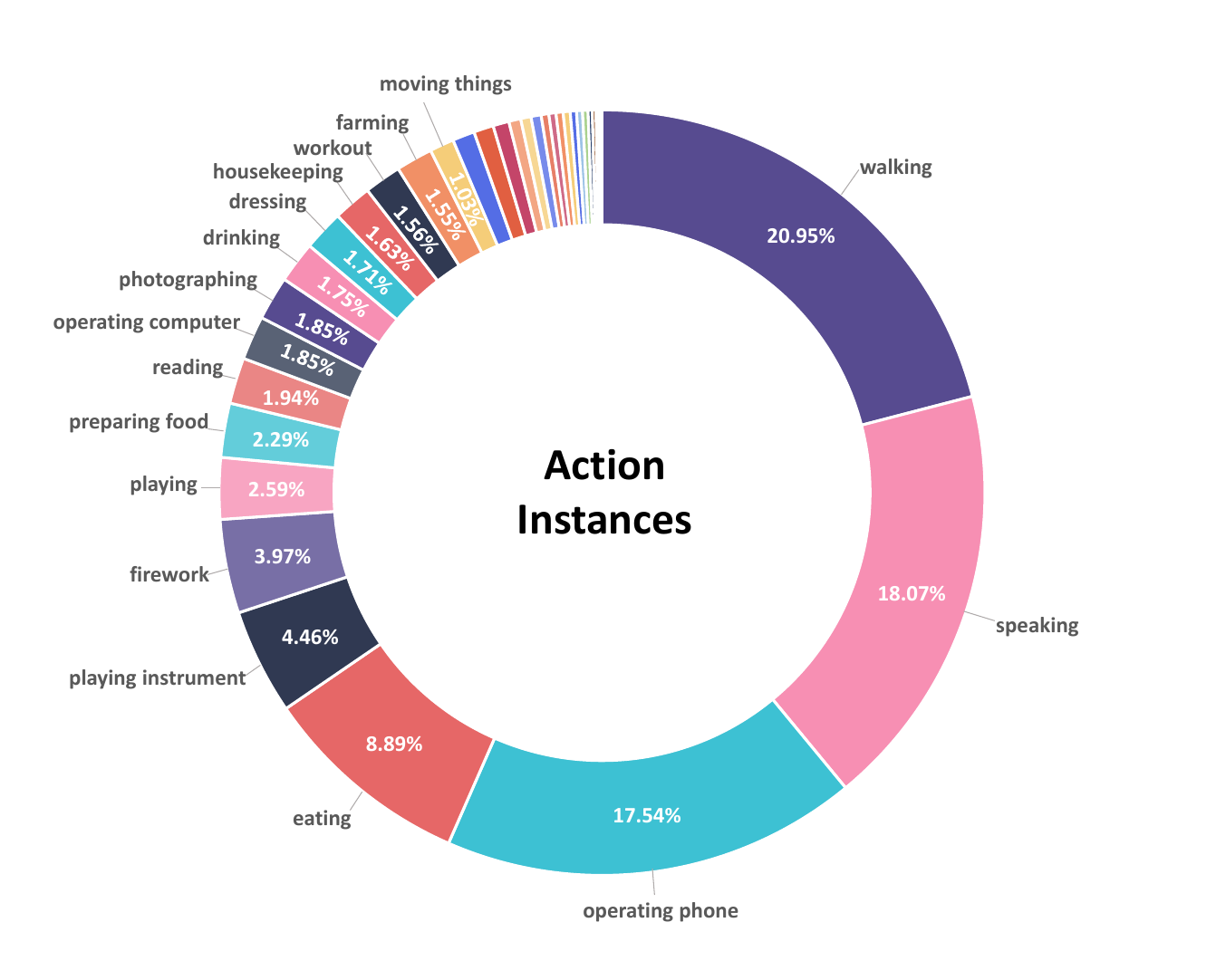}
     % \centering 
     { \flushright
    \caption{Overall distribution of actions duration.}
        \label{fig:labels}
    }
  \end{subfigure}
 % ---------------------------------------- 
    \centering
    \hfill
  \begin{subfigure}[b]{0.33\textwidth}
    \includegraphics[width=\textwidth]{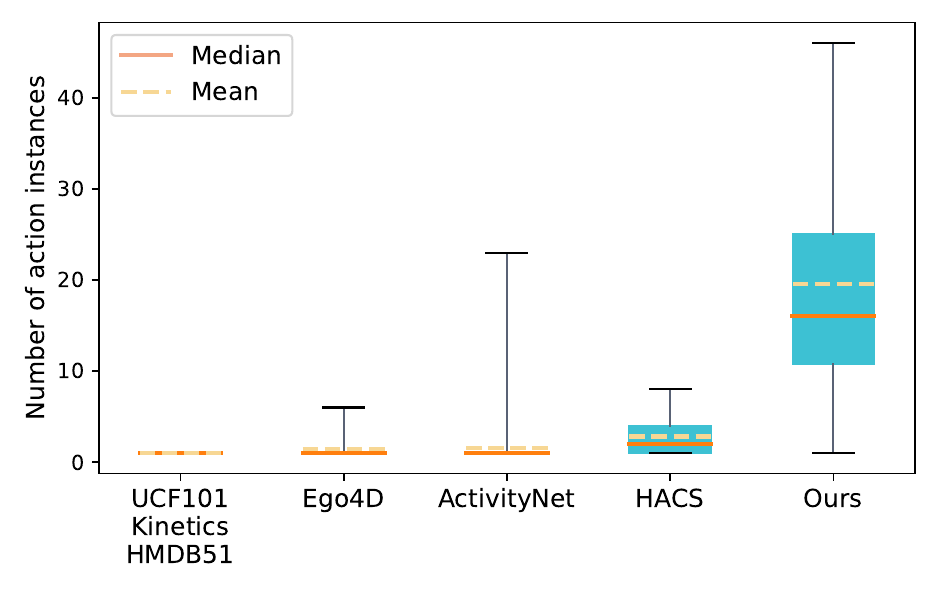}   
    \caption{Distribution comparison of the number of action instances per video.}
    % between existing datasets (e.g. UCF101 \cite{soomro2012ucf101}, Kinetics \cite{kay2017kinetics}, HMDB51 \cite{Kuehne11}, Ego4D \cite{grauman2022ego4d}, ActivityNet \cite{caba2015activitynet}, HACS \cite{zhao2019hacs}) and Ours.}
    \label{fig:comparison_dataset}
  \end{subfigure}
      \hfill
  \begin{subfigure}[b]{0.225\textwidth}
   \vspace*{-2mm}
   % L D R U  0 20
         % \includegraphics[width=\textwidth, trim=-10 7 7 0, clip]{asset/img/analysis/time1.pdf}

    \includegraphics[width=\textwidth, trim=-5 -6 7 -2, clip]{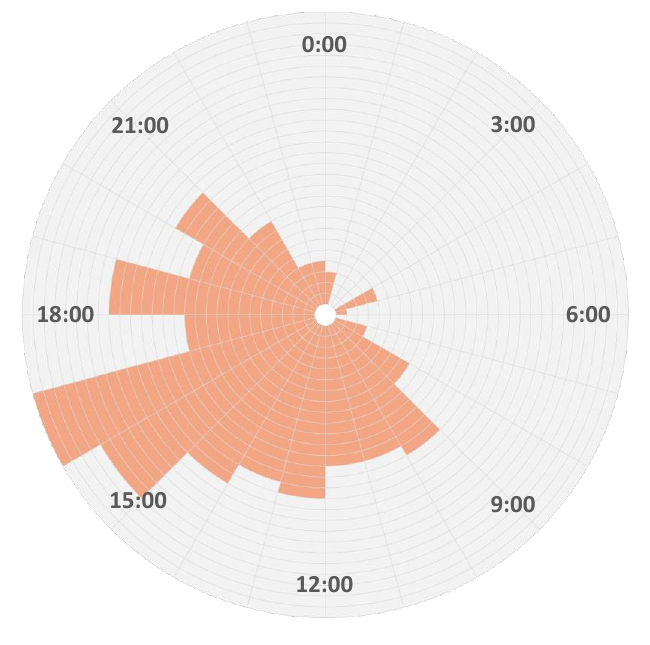}

        \caption{Capture time of the day.}
        \label{fig:timeofday}
  \end{subfigure}
  \hfill
  \begin{subfigure}[b]{0.39\textwidth} 
     \vspace*{-2mm}
     \includegraphics[width=\textwidth]{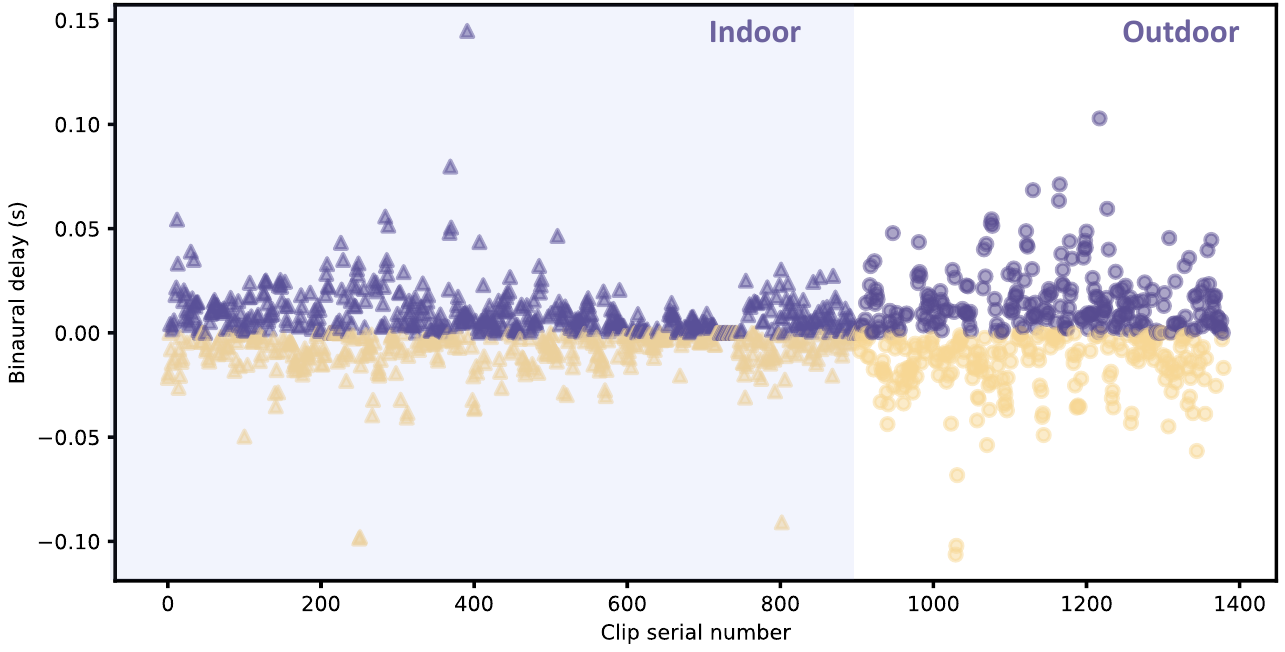}
    \vspace{0.5mm}
    \caption{Binaural delay per clip.}
    \label{fig:at_statistic}
    % \end{minipage}
     % \includegraphics[width=\textwidth]{asset/img/analysis/at_statistics_distribution.pdf}
    % \caption{Overall binaural delay histogram}
    % \label{fig:at_statistic_distribution}
  \end{subfigure}
  \caption{\textbf{Dataset statistics analysis}, on the distributions of (a) the scene category, (b) action distribution per cities, (c) temporal action instance duration, and (d) number of actions per video, (e) capturing time, (f) binaural delay per clip.}
  \label{fig:TSL and complexity}
  \vspace{-1mm}
\end{figure*}

Once we obtained the raw data, we aligned the different viewpoints and modalities through a specific process. The initial raw footage captured by the two fish-eye cameras on the \textit{$360^{\circ}$ camera} was in the form of two circular videos, which were then stitched and de-warped into a spherical panorama. This panorama can be projected into an equirectangular format to produce a panoramic video. However, this direct compression of the spherical view into a rectangular format can introduce unnatural distortions. In order to provide a more natural and informative view, we inversely project a rectangular region into equirectangular space and use it to crop the spherical panorama. We use optical flow to determine the crop region with the most motion activity in the spherical panorama field. This crop region is then projected back to rectangular, resulting in an informative video view with minimal distortions.

Egocentric binocular videos, as shown in Figure \ref{fig:data}{\color{Mahogany}(c)}, were captured ranging from approximately 30 seconds to 1 minute in duration for each clip.  A total of 1 to 5 stereo clips were recorded, scattered throughout the duration of the average 6 mins \du~video. In addition to stereo videos, we also provide the corresponding monocular videos for the egocentric view.

The audio recordings were temporally aligned with their corresponding videos with left/right channel modality. The four-channel audios with the  \du~panoramic video are provided as well for further exploration. Moreover, we also provide the directional information of the audio which was presented using the estimated interaural time delay of the sound obtained from the method introduced in \cite{chen2022sound}.  The GPS information and weather information were also provided.

Given the possibility of occlusions in regions visible to the egocentric camera but not to the \textit{$360^{\circ}$ camera}, we ensured during data collection that the cameras were positioned in close proximity. This setup, with clear mutual visibility, allowed both cameras to capture a similar overall scene.

\subsection{Scene Selection}
To broaden scene coverage and promote multi-modal collaborative learning, we integrated a strategic selection process for captured scenes, governed by three key criteria:

i) Scene categories must be carefully crafted to be comprehensive, yet concise, while also being authoritative and reflective of everyday life. The location where a scene unfolds plays a crucial role in providing essential environmental context to the activities within it~\cite{meagher2020ecologizing}. Distinct scenes can impart unique meanings or emotional nuances to identical events. For instance, the act of chatting could convey divergent implications in a school setting as compared to a home environment. Such nuances are critical as they offer deeper insights into the contextual interpretation of behaviours and interactions in varied settings.

ii) The data should ideally span a wide array of weather and lighting conditions. This criterion aims to ensure the inclusion of both indoor and outdoor activities under various environmental scenarios. Such diversity is important in accurately representing the multifaceted nature of daily life and the various conditions in which these activities occur.

iii) Our third criterion is the inclusion of scenarios rich in distinctive sound sources, particularly those where multiple activities co-occur. It is essential for the dataset to not only visually represent these activities but also to capture the corresponding auditory elements. The goal is to present the complexity and realism of real-world environments as much as possible, marked by simultaneous and various actions and behaviours.

\begin{table*}[t]
	\centering
	\caption{ \textbf{Dataset comparison.} Ego: Egocentric, V: Video, A: Audio, A+V: Audio-visual events.}
    \label{tab:dataset comparison}
    % \hspace{0.2pt}
	% \small
	\setlength{\tabcolsep}{4pt}
    \resizebox{1\textwidth}{!}{
    \begin{tabular}{l|c|c|c|c|c|c|c|c|c|c|c|c}
     \toprule
     \multirow{3}{*}{Dataset}  & \multicolumn{4}{c|}{Video Viewpoints} &\multicolumn{3}{c|}{Other Modalities} & \multicolumn{3}{c|}{Statistics} & \multicolumn{2}{c}{Attributions}  \\
     & Third-person & \du &Ego &Ego 
&Normal&Directional& GPS & Avg  & Total &  Frames & Annotations & Multiple \\
        &Front View & Panoramic & Monocular &Binocular
 &Audio&Binaural Delay& Info  & Duration & Duration(s) & Count(K) & Source & Events\\

    % cmark and xmark
     \midrule
    UCF101~\cite{soomro2012ucf101}&\cmark&\xmark&\xmark&\xmark&\cmark&\xmark&\xmark&7.21 s&  96,000  &2,400  &V&\xmark\\
    Kinetics~\cite{kay2017kinetics}&\cmark&\xmark&\xmark&\xmark&\xmark&\xmark&\xmark&10 s&2,998,800&74,970&V&\xmark\\
    
    HMDB51~\cite{Kuehne11}&\cmark&\xmark&\xmark&\xmark&\xmark&\xmark&\xmark&3 s&21,426&643&V&\xmark\\
    
    ActivityNet~\cite{caba2015activitynet}&\cmark&\xmark&\xmark&\xmark&\xmark&\xmark&\xmark&2 min&2,332,800&11,664&V&\cmark\\
    
    EPIC-Kitchens~\cite{damen2018scaling}&\xmark&\xmark&\cmark&\xmark&\cmark&\xmark&\xmark&7.6 min&198,000&11,500&V&\xmark\\
    Ego4D~\cite{grauman2022ego4d}&\xmark&\xmark&\cmark&\xmark&\cmark&\xmark&\cmark&8 min&13,212,000&-&A+V&\cmark\\

    \midrule
    \textit{360+x} (Ours)&\cmark&\cmark&\cmark&\cmark&\cmark&\cmark&\cmark&6.2 min& 244,000 & 8,579 &A+V&\cmark\\
     \bottomrule
    \end{tabular}
}
    
% \vspace{-0.3cm}
\end{table*}

It is worth noting that our dataset was collected across several countries, including the United Kingdom (\eg London, Birmingham, Cardiff and Jersey), France (Paris), Spain (\eg Oviedo and Picos de Europa), China (\eg Guangzhou and Shenzhen), and Japan (\eg Kyoto and Osaka). During the data collection, the \textit{$360^{\circ}$ Camera} was placed statically to record the scene, while a capturer wearing the \textit{Spectacles} glasses recorded first-person interactions with the scene.

\paragraph{Sensitive data handling.} 
Our dataset was collected in a real-world setting and may contain sensitive personal information (\eg human faces). 
To ensure ethical and responsible research, the video capture was conducted with proper consent. Additionally, we have taken measures to protect privacy by anonymising the data. This includes applying a face detection mechanism to outline predicted face locations in each frame and applying blurring filters to maintain meaningful details while ensuring information security. More details on our privacy protection measures can be found in the supplementary material section \ref{ethical}.

\subsection{ Data Annotation }
\label{sec:3.3DataAnnotation}

% \paragraph{Metadata and scene label.}
\paragraph{Scene label rationale.}
% classification labels for various
% provides a diverse range of scenes, 
The \textit{360+x} dataset comprises a total of 28 scene categories (15 indoor scenes and 13 outdoor scenes), as illustrated in Figure \ref{fig:categories}. To establish comprehensive and authoritative scene categories that reflect daily life, we referred to the Places Database \cite{zhou2017places}, which is derived from WordNet \cite{miller1995wordnet}, as our primary basis. We then leverage the sophisticated semantic analysis capabilities of large language models, to conduct a thorough filtering and classification of a multitude of everyday scenes. This curation resulted in a refined set of 28 scene categories, each symbolising aspects of daily life. Simultaneously, the recordings concentrate on capturing common occurrences within conventional settings, providing a realistic depiction of everyday life. Detailed descriptions defining each category, along with discussions regarding these constraints and potential sampling biases, are presented in the supplementary material section \ref{scene} and section \ref{limit}, respectively.

% \td{ weather, gps, json is not concluded in current dataset, deleted the following paragraph}.
% The dataset consists of multiple series, including 3 weather conditions (sunny, cloudy, rainy) and 4 time periods (morning, noon, afternoon, and evening). To provide accurate location and time information, the dataset includes GPS coordinates and timestamps. Additionally, manually labelled descriptions are provided to enhance the understanding of each scene, covering weather, lighting, GPS\td{check if correct}, and time details. All relevant information, such as weather conditions, time periods, GPS coordinates, timestamps, and descriptions, is consolidated into a single JSON file for each scene, enabling convenient access and comprehensive scene analysis.

\vspace{-3mm}
\paragraph{Temporal segmentation label.}
We also provide temporal segment labelling for the understanding of activities in the shooting scenes. We follow the activity hierarchy standard defined by ActivityNet \cite{caba2015activitynet}, which provides a comprehensive categorisation of human activities, consisting of seven top-level categories (\textit{Personal Care, Eating and Drinking, Household, Caring and Helping, Working, Socialising and Leisure, and Sports and Exercises}). To capture the diversity and granularity of activities within each category, we defined a total of 38 action instances, covering specific actions and behaviours.
To ensure high-quality annotations, the temporal segmentation labelling was annotated by three experienced annotators. Each annotator independently annotated the temporal segments corresponding to the activities in the videos. To obtain a consensus, we merged the individual annotations and resolved any discrepancies according to discussion and consensus among the annotators.
% In total, 38\td{please check the number} temporal actions are annotated in the clips.

\subsection{Dataset Statistics and Analysis}
% \vspace{-3mm}

\paragraph{Overview.}
Existing publicly available datasets primarily focus on visual unimodality \cite{soomro2012ucf101, kay2017kinetics, kuehne2011hmdb, caba2015activitynet, damen2018scaling}. In contrast, our dataset introduces a novel approach by collecting different views or modalities, as presented in Table \ref{tab:dataset comparison}, including \du~panoramic video, third-person front view video, egocentric monocular video, egocentric binocular video, normal audio, directional binaural delay, location and textual scene description. 
This diverse range of modalities provides multiple dimensions and clues for understanding and analysing complex scenes. 
Our dataset consists of 2,152 videos representing 232 data examples, with 464 videos captured using the \textit{360 camera} and the remaining 1,688 recorded with the \textit{Spectacles camera}.

Figure \ref{fig:categories} presents the distribution of video counts across each of the 28 scene categories. Our dataset is characterised by a balanced distribution of data across these scenes. Notably, it diverges from conventional databases like UCF101~\cite{soomro2012ucf101}, Kinetics~\cite{kay2017kinetics}, HMDB~\cite{kuehne2011hmdb}, and ActivityNet~\cite{caba2015activitynet}, particularly in terms of average video duration, which is approximately 6.2 minutes. This longer duration is crucial for maintaining the integrity and coherence of actions within each scene, allowing for a comprehensive temporal analysis of the activities.

\vspace{-2mm}
\paragraph{Temporal segment label.}
The annotations of temporal segment labels in our dataset contribute to the fine-grained analysis of activities. We defined 38 
% \td{maybe need to check the number} 
action instances representing specific actions and behaviours. The length of each segment labelled with a specific activity varies across the dataset, as depicted in Figure \ref{fig:labels}. Note we acknowledge the significance of audio in accurately identifying certain actions, such as `\textit{coughing}' or `\textit{clapping}'. Therefore, our dataset combines audio information to enhance accuracy in action recognition \cite{soomro2012ucf101, kay2017kinetics, kuehne2011hmdb, caba2015activitynet, damen2018scaling}, as shown in Table \ref{tab:dataset comparison}.

\vspace{-2mm}
\paragraph{Comparative complexity.}
% Currently available publicly accessible datasets primarily focus on visual unimodality\cite{soomro2012ucf101,kay2017kinetics,kuehne2011hmdb,kuehne2011hmdb,caba2015activitynet,damen2018scaling}. 

Due to its realistic scene simulation, our dataset offers more complexity compared to previous datasets. This complexity arises from the diverse range of activities and interactions captured, resulting in a more challenging and realistic setting for scene understanding and activity recognition. As shown in Figure \ref{fig:comparison_dataset}, most existing datasets, such as UCF101 \cite{soomro2012ucf101}, Kinetics \cite{kay2017kinetics}, and HMDB51 \cite{Kuehne11}, typically consist of one action instance per video. While datasets like Ego4D \cite{grauman2022ego4d} and ActivityNet \cite{caba2015activitynet} have large volumes and broad coverage, they often contain a limited number of action instances per individual video. The HACS dataset \cite{zhao2019hacs} contains more multiple action instances per video but still pales in comparison to the richness of the proposed dataset. Our dataset surpasses these existing datasets in terms of the number of action instances per video, showcasing the extensive variety of activities captured. The improved complexity and richness of our dataset enable follow-up research to explore and develop more robust algorithms, pushing the boundaries of scene understanding in real-world contexts.

%%%%%%%New
\vspace{-3mm}
\paragraph{Data distribution.}

 We have ensured a balanced distribution across various dimensions, including scene categories, action instances, binaural delay, \etc. Figure \ref{fig:categories} depicts the scene number distribution across 28 scene categories, demonstrating a comprehensive coverage of scene categories. Notably, the dataset achieves an almost equal proportion of indoor and outdoor scenes, accounting for 54.7\% and 45.3\%  respectively. Our dataset allows each scene to conclude multiple diverse action instances naturally, and also enables different scenes to share common action instances. Notably, in Figure \ref{fig:geometry}, it displays the `\textit{types of action per location}' that can be observed in the geographic distribution and the diversity of the data, where the inner circle shows the location and the outer circle shows the action types captured in each location. As illustrated in Figure \ref{fig:labels}, the distribution of action duration shows our dataset has captured extensive and realistic human behaviours across natural scenes. One interesting observation from our dataset is the high-frequency occurrence of action `\textit{operating phone}', which contributes 17.54\% of the whole duration, providing a reflection of mobile usage in modern daily life. Additionally, the dataset offers valuable directional audio to supplement visual understanding. The distribution of data capture times in the dataset corresponds with natural human activities, as shown in Figure~\ref{fig:timeofday}. Human activities throughout the day are mainly concentrated during the daytime (more in the afternoon and evening). Figure \ref{fig:at_statistic} illustrates the diversity of binaural delay for each clip. The positive point means the audio is directed towards the left direction while the negative the right. 
 % Figure~\ref{fig:at_statistic_distribution} represents the balanced clip histogram distribution of binaural delay.  
 In summary, the presented \textit{360+x} dataset covers broad modalities and diversity with an authentic distribution from different perspectives, mimicking real daily life.
 
 % {\color{red} TODO：}
 % In our dataset, which encompasses broad modalities, diversity and distribution of each modality are well concerned, thus providing valuable cues for a more intuitive understanding of human activities.

\section{Benchmark and Experiments}
\label{sec:benchmark}

To establish a comprehensive benchmark for the presented \textit{360+x} dataset, we choose five visual understanding tasks to delve into the exploration of multiple viewpoints and modalities usage, including: video scene classification, temporal action localisation, cross-modality retrieval, self-supervised representation learning, and dataset adaptation.

\noindent\textbf{Remark:} Unless specifically stated otherwise, the experiments on \textit{360+x} will utilise three views: the \du~view, egocentric binocular view, and the third-person front view.

% To assess the performance of various algorithms on the \textit{360+x} dataset, three State-of-the-art machine learning models were trained using our dataset to establish a benchmark for comparison.
% widely adopted video-specific 

\subsection{Experimental Setting}
% \vspace{-3mm}
\paragraph{Models.}
We employed a consistent set of model backbones across different tasks to minimise model interference, except for \textit{temporal action localisation} task (detailed in section~\ref{sec:tal}). We followed the commonly used setup and selected the backbone I3D \cite{kay2017kinetics} as our video model. To handle audio-related aspects, we chose the VGGish \cite{hershey2017cnn} as our audio model. Additionally, for directional binaural feature extraction, we utilised the ResNet-18 model \cite{he2015deep}.
% 
% As specific for each task, an additional fully-connected layer will put on top on feature output by backbone to handling task-specified prediction.
%% TODO
A linear layer is positioned after the backbones to carry out each specific task based on backbone output features.

It is important to note that a simple concatenation of all modalities features can diminish the potential information derived from multi-modality \cite{wang2020makes}. 
Therefore, instead of solely concatenating modality features, we leverage a hierarchical attention mechanism for multi-modality integration. In this approach, the directional binaural feature serves as an attention query to direct focused attention towards the audio feature, enabling it to encapsulate the directional information into the audio feature. At the same time, the audio feature is also leveraged by acting as a query itself, enabling it to attentively interact with the video feature. This mechanism allows for creating a synergistic representation of the underlying data that integrates the features of all modalities. For more details and in-depth analysis, please refer to the supplementary material section \ref{sef:modalityfusion}.

\vspace{-3mm}
\paragraph{Training and verification setup.}
For each temporal action localisation model, we follow their original training settings. For I3D, VGGish, and ResNet-18 networks, the training settings are 200 epochs with the parameters described in \cite{Peng2022Balanced}. The training process utilises the AdamW optimiser with a learning rate of $1 \times 10^{-5}$ and a decay rate of 0.1 at the 80th and 120th epochs. We also apply data augmentation techniques such as rotation, scaling, and colour jittering.
The dataset was divided into training, validation, and test sets, following an 80/10/10 split. To ensure a balanced representation of scene categories, the examples were stratified probabilistically across the sets.

\begin{table}
    % \vspace{-1mm}
	% \caption{ \textbf{Video classification results.} The results obtained using only the \du view serve as a baseline for each respective modality combination. V:Video, A:Audio, D:Directional audio time delay.}
	\centering
	\caption{Video classification performance across different views (Ego: egocentric binocular view, Front: third-person front view, and \du: \du~view) and data modalities (V: Video, A: Audio, D: Directional binaural delay). Reported in Avg. Prec. (\%).}
    % \hspace{0.2pt}
    \label{tab:cls}
	\small

\resizebox{\columnwidth}{!}{
    \begin{tabular}{l|c|c|c}
     \toprule
   \multirow{2}{*}{Selected Views}   & \multicolumn{3}{c}{Modalities} \\
     &   V &  V + A &  V + A + D\\
      \midrule
      % \multicolumn{3}{c}{Single Video Type} \\
      % \midrule
     \multirow{1}{*}{Egocentric Only }
      & 51.95 \e{0.0}    & 55.24 \e{0.0}   &     58.92  \e{0.0}     \\  
  
      \multirow{1}{*}{Front Only}
      & 54.05 \p{2.1} & 65.33  \p{10.1} & 67.19 \p{8.3}  \\

      \multirow{1}{*}{360$^{\circ}$ Only}
    & 56.33 \p{4.4} & 67.14 \p{11.9} & 70.95 \p{12.0}         \\  
     
      \multirow{1}{*}{360$^{\circ}$ + Egocentric}
      & 58.99 \p{7.0} & 70.48 \p{15.2} & 72.11 \p{13.2}     \\  

      \multirow{1}{*}{360$^{\circ}$ + Front} 
      &59.70 \p{7.8} & 75.06 \p{19.8} & 77.69 \p{18.8}   \\

      \multirow{1}{*}{360$^{\circ}$ + Front + Ego}
     & \textbf{63.73 \p{11.8} }& \textbf{77.32 \p{22.1}} & \textbf{80.62 \p{21.7}}    \\  

% ---------------------- Color ------------------------------
   % \midrule

     \bottomrule
    \end{tabular}
    }
    
% \vspace{-0.3cm}
% \end{table*}
% \end{wraptable}
\end{table}

\subsection{Video Scene Classification}
\label{sec:classification}
Video scene classification assigns scene labels to videos based on their frames, enabling analysis of visual content and determining the subject matter.

% ----------------------------------- 

\begin{table*}[!t]
    \centering
     \caption{Temporal action localisation results. Baseline extractors are used in \cite{zhang2022actionformer, tang2023temporalmaxer, shi2023tridet, chen2020vggsound}.
    The mAP$@\sigma$ represents the mean average precision (\%) at a threshold of $\sigma$.  The best performance is achieved by employing \textit{V+A+D} modalities with extractors pre-trained on \textit{360+x}. }

\resizebox{\textwidth}{!}{
    \begin{tabular}{@{}c|c|cccc|cccc|cccc@{}}
     \toprule
     
    \multirow{3}{*}{Extractors} & \multirow{3}{*}{Modalities} & \multicolumn{4}{c|}{Actionformer~\cite{zhang2022actionformer}} & \multicolumn{4}{c|}{TemporalMaxer~\cite{tang2023temporalmaxer}} & \multicolumn{4}{c}{TriDet~\cite{shi2023tridet}} \\
    & & mAP & mAP & mAP & \multirow{2}{*}{Avg.} & mAP & mAP & mAP & \multirow{2}{*}{Avg.} & mAP & mAP & mAP & \multirow{2}{*}{Avg.\ }\\
    & & $@$0.5 & $@$0.75 & $@$0.95 &  & $@$0.5 & $@$0.75 & $@$0.95 &  & $@$0.5 & $@$0.75 & $@$0.95 &\\

      \midrule
      
       \multirow{2}{*}{\makecell[c]{Baseline \\ Extractors}} 
        & V 
       & $11.9$ \e{0.0}& $7.8$\e{0.0} & $3.3$\e{0.0} & $7.7$\e{0.0} & 
        $13.1$\e{0.0} & $8.8$ \e{0.0}& $3.7$\e{0.0} & $8.6$\e{0.0} &
        $16.7$\e{0.0} & $10.1$\e{0.0} & $4.8$ \e{0.0}& $10.5$\e{0.0}
       \\
       
       & V + A 
       & $19.1$ \p{7.2}& $11.3$\p{3.5} & $4.2$\p{0.9} & $11.5$\p{3.8}& 
        $21.0$\p{7.9} & $14.8$\p{6.0} & $5.6$\p{1.9} & $13.8$\p{5.2}& 
        $23.6$\p{6.9} & $17.2$\p{7.1} & $6.4$\p{1.6} & $15.7$\p{5.2}\\
       
      % & V + A + D 
      %  & -- & -- & -- & -- 
      %  & -- & -- & -- & -- 
      %  & -- & -- & -- & -- \\
       
      \midrule
      
       \multirow{3}{*}{\makecell[c]{Pre-trained \\ on \textit{360+x}}} 
      & V 
       & $16.4$\p{4.5} & $9.8$ \p{2.0}& $3.9$\p{0.6} & $10.0$\p{2.3} & 
      $20.4$\p{7.3} & $14.3$\p{5.5} & $5.2$\p{1.5} & $13.3$\p{4.7} &  
      $21.1$\p{4.4} & $15.3$\p{5.2} & $5.5$ \p{0.7}& $14.0$\p{3.5} \\
       
       & V + A 
       & $23.6$\p{11.7} & $16.9$\p{9.1} & $5.7$\p{2.4} & $15.4$\p{7.7} & 
        $25.8$\p{12.7} & $18.0$\p{9.2} & $6.4$\p{2.7} & $16.7$\p{8.1} & 
        $26.4$\p{8.7} & $18.5$\p{8.4} & $6.9$\p{2.1} & $17.3$\p{6.8}\\
       
       & V + A + D 
       & $\boldsymbol{24.9}$ \p{13.0}& $\boldsymbol{17.4}$ \p{9.6}& $\boldsymbol{6.1}$ \p{2.8}& $\boldsymbol{16.1}$ \p{8.4} &
        $\boldsymbol{26.6}$ \p{13.5}& $\boldsymbol{18.3}$ \p{9.5}& $\boldsymbol{6.5}$ \p{2.8}& $\boldsymbol{17.1}$\p{8.5} &
       $\boldsymbol{27.1}$ \p{10.4}& $\boldsymbol{18.7}$ \p{8.6}& $\boldsymbol{7.0}$ \p{2.2}& $\boldsymbol{17.6}$\p{7.1}
       \\
       
     \bottomrule
    \end{tabular}
    }
% \vspace{-1mm}
\label{tab:tal_combined}
\end{table*}

\vspace{-3mm}
\paragraph{Single view vs. multi-view.} 
First, we are interested in the influence of different combinations of video views on the classification performance. The results, representing each combination, are summarised in Table \ref{tab:cls}.
The results for single views are presented in the first three rows, indicating that using a single \du~panoramic view outperforms using either an egocentric binocular view or a third-person front view only. When employing multiple views, it is noted that better performance can be achieved compared to using a single view. Specifically, utilising all three views leads to the best performance. Such a performance can be attributed to the fact that although these three views describe the same scene, each different view offers a unique perspective that contributes to a more comprehensive understanding of the scene, resulting in improved performance.

\vspace{-3mm}
\paragraph{Single-modality vs. multi-modality and more.} 
We further investigate the impact of modalities on the model's performance. Various combinations of modalities are analysed, and the results are summarised in Table \ref{tab:cls} on a column-wise basis. In particular, the first column represents the visual modality alone, the second column combines video with audio, and the last column incorporates visual, audio, and directional binaural information modalities.

The inclusion of additional modalities leads to average precision improvements. For example, when all three views are utilised, incorporating more modalities results in improvements of \textit{13.59\%} and \textit{16.89\%}, respectively. This underscores the benefits of leveraging multiple modalities for a more comprehensive understanding of the scene and enhancing overall performance.

\subsection{Temporal Action Localisation}
\label{sec:tal}
% \vspace{-3mm}
Temporal Action Localisation (TAL) is a video understanding task that involves the dense identification and temporal segmentation of activities within a video stream over a specific time period. Current TAL approaches typically employ a two-stage paradigm \cite{wang2021proposal, zhang2022actionformer}. 
The first stage extracts features from the entire video, and the second stage predicts temporal segmentation based on these features.

\vspace{-2mm}
\paragraph{Feature extractors.}
Baseline extractors are widely utilised for various datasets, \eg ActivityNet \cite{caba2015activitynet} and Ego4D \cite{grauman2022ego4d}, on the TAL task. 
The baseline video features are obtained from an I3D model pre-trained on the Kinetics400 dataset \cite{kay2017kinetics}. The baseline audio features are derived from the pre-classification layer following activation of the VGGish model, pre-trained on AudioSet \cite{audioset}. There is no baseline extractor for \textit{directional binaural delay} feature, so the V+A+D modality was not included accordingly.
For a fair comparison, we reused our video classification models in section \ref{sec:classification} as \textit{Pre-trained on 360+x} extractors, following the same baseline extraction setup for both video and audio features. Additionally, the ResNet-18 feature extractor was used for \textit{directional binaural delay} feature extraction.

% This model takes input clips of 64 frames and produces output tensors with 1024-d features and a clip stride of 4 frames at 25 fps. As a result, the feature size is Tv x 1024, where Tv = 25/4. 
% The resulting audio feature tensor has a dimensionality of 128 and corresponds to the same time interval as the video, yielding features of size Tv x 128

% For comparative purposes, video classification models from the aforementioned subsection were reused for feature extraction as 'Pre-trained on \textit{360+x}' extractors, adhering to the baseline extraction setup for both video and audio features. Additionally, the  ResNet-18 (1D) feature extractor was also reused for \textit{directional audio time delay} feature extraction, resulting in features of size Tv x 128. 

\vspace{-2mm}
\paragraph{Experimental results.}
We provide a concise overview of the performance comparison for various temporal action localisation methods, including ActionFormer \cite{zhang2022actionformer}, TriDet \cite{shi2023tridet} and TemporalMaxer \cite{tang2023temporalmaxer}, between the baseline extractors and our \textit{Pre-trained on 360+x} extractors. The summarised results are presented in  Table \ref{tab:tal_combined}, from which we can see that the introduction of additional modalities (\ie audio and direction binaural delay) has a prominent positive impact on the TAL task, leading to performance improvements for both sets of extractors. This result highlights the importance of leveraging multiple modalities in enhancing the accuracy and effectiveness of temporal activity localisation techniques. Using our custom extractors can provide additional improvements, as the baseline extractors may not be optimised for our specific binocular or \du~views. Additional results on variations of views can be found in the supplementary material section \ref{sec:tal_views}.

\begin{table}
	\centering
    % \vspace{-1mm}  ''D'':
\caption{Q-to-Video retrieval results. The superscript* indicates modalities are co-trained. Recall reported with rank in $\{1, 5, 10\}$. }
\label{tab:ret}
\resizebox{\columnwidth}{!}{
\begin{tabular}{@{}l|ccc@{}}
     \toprule
    Query Modality	& R1 (\%) 	& R5 (\%) & R10 (\%)  \\
      \midrule
        A   &      $39.14$ \e{0.0} &   $62.76 	$  \e{0.0}&    $79.21$\e{0.0}\\  
      A + D &        $44.30$ \p{5.16} &     $66.92$  \p{4.16}&    $84.78$  \p{5.57} \\
    (A + D)$^*$  &   \bm{$55.88$}\p{16.74}  &    \bm{$72.53$} \p{9.77} &  \bm{$86.6$}  \p{7.39}  \\
     \bottomrule
\end{tabular}
}
% \vspace{-1mm}
\end{table}

\subsection{Cross-modality Retrieval}
\label{sec:retrieval}
% \vspace{-3mm}
In this context, we focus on a series of retrieval tasks that across modalities including audio, video and directional binaural delay. 
In a modality-specific retrieval scenario, the query modality (Q) serves as the input for retrieving the key modality (K) in the Q-to-K retrieval task. The performance evaluation metric R$\theta$ represents the recall at ranks $\theta$.

% _{\theta=\{1,5,10\}}

%-------------------------------------------
% \input{asset/table/ssl+cls}
% \captionsetup[table]{labelformat=simple, labelsep=period}
% set up labelformat and labelsep for subtable
% \captionsetup[subtable]{labelformat=simple}
\renewcommand{\subtablename}{Table \thetable} 
\renewcommand\thesubtable{\hspace{-0.5mm}}  % \thetable

\begin{table*}	
	% \centering
 % \left
	\begin{subtable}[t]{0.45\columnwidth}
 	\centering
	\subcaption{Models with different pre-train methods were fine-tuned and tested on video classification. The experiments use all three video views. Reported in Avg. Prec. (\%). }
\label{tab:ssl+cls}
		% \centering
\resizebox{1.0\columnwidth}{!}{
% \hspace{-4.2cm}
    \begin{tabular}{@{}l|ccc@{}}
     \toprule
   \multirow{2}{*}{ Pre-train Method } &   \multicolumn{3}{c}{Modalities}  \\
     & V & V + A & V + A + D  \\
      \midrule
      % \pm
      From Scratch &
     63.73 \e{0.0} & 77.32 \e{0.0} & 80.62 \e{0.0} 
      %& \multirow{3}{*}{360$^{\circ}$  + Front +  Binocular}
     \\  
        
      % \midrule
      Video Pace~\cite{Wang20} & 69.27 \p{5.5} & 79.56 \p{2.2} & 81.97 \p{1.3} \\  
 
      % \midrule
    Clip Order~\cite{xu2019self} & 69.91 \p{6.2} & 80.14 \p{2.8} & 82.18 \p{1.6} \\  

      % \midrule
       % Video Pace + Clip Order 
       VP~\cite{Wang20} + CO~\cite{xu2019self}
     & \boldsymbol{$76.84$~\p{13.1}} & \boldsymbol{$82.66$~\p{5.3}} & 
     \boldsymbol{$83.32$~\p{2.7}}\\
      % \midrule
     \bottomrule
    \end{tabular}
    }
	\end{subtable}
	\;
	% \quad
	% \quad
\setcounter{table}{\numexpr\value{table}+1}   %%  Add one to the counter
	\begin{subtable}[t]{0.515\columnwidth}
 	\centering
	 \subcaption{Comparison between supervised pre-trained extractors with SSL pre-trained counterparts on TAL task. The experiments use all three video views with modalities (V+A+D). }
\label{tab:ssl+tal}
		% \centering
  % \hspace{-3cm}
	\resizebox{1.0\columnwidth}{!}{
    \begin{tabular}{@{}l|ccc|c@{}}
     \toprule
      \makecell[l]{Pre-train Method}
      & \makecell[c]{mAP\\$@$0.5} & \makecell[c]{mAP\\$@$0.75} & \makecell[c]{mAP\\$@$0.95} & Avg.  \\
      \midrule
       Supervised
       & $27.1$ \e{0.0} & $18.7$ \e{0.0} & $7.0$ \e{0.0} & $17.6$ \e{0.0}   \\
       % \midrule
       Video Pace~\cite{Wang20}
       & $29.4$ \p{2.3} & $19.6$\p{0.9} & $7.4$\p{0.4} & $18.8$ \p{1.2}  \\
        % \midrule
   Clip Order~\cite{xu2019self}
       &$28.9$ \p{1.8}& $19.3$ \p{0.6}& $7.3$ \p{0.3}& $18.5$ \p{0.9} \\
        % \midrule 
     % Video Pace + Clip Order 
    VP~\cite{Wang20} + CO~\cite{xu2019self}
       &\boldsymbol{ $30.3$ }\p{3.2}& \boldsymbol{$20.2$}\p{1.5} & \boldsymbol{$7.9$}\p{0.9} & \boldsymbol{$19.5$~\p{1.9}}  \\
     \bottomrule
    \end{tabular}
    }
		% \caption{Caption 2}\label{table:1b}
	\end{subtable}
	% \caption{Main table caption}\label{table:1}
\end{table*}

% \renewcommand{\thesubtable}{}  
%-------------------------------------------

%-------------------------------------------
\begin{table*}%[!h]  % t
\begin{minipage}{1.0\textwidth}
	\centering
 % \vspace{2mm}
	\caption{Following original setup of THUMOS14 dataset~\cite{THUMOS14}, our dataset adaptation task uses video modality only.  } 
\label{tab:thumos}
\vspace{-2mm}

\resizebox{\textwidth}{!}{
\centering
    \begin{tabular}{@{}l|ccccc|c@{}}
     \toprule
    Feature Extractor & mAP$@0.3$ 	& mAP$@0.4$ & mAP$@0.5$ & mAP$@0.6$& mAP$0.7$& Avg. \\
      \midrule
        Kinetics400~\cite{kay2017kinetics} (\textit{Pre-train})  &     $83.7 	$ \e{0.0} &  $80.2 	$\e{0.0}  &  $72.8 $\e{0.0}  &  $ 62.4 	$ \e{0.0} &  $47.4 	$\e{0.0}  &  $69.5$\e{0.0} \\  
        % 360+x (V+A+D jointly \textit{PT})&  $	80.5 	$  &  $78.4 	$  &  $71.7 	$  &  $63.1 $  &  $	51.9 	$  &  $69.1$\\  
      \textit{360+x} (\textit{Pre-train}) &      $84.5$\p{0.8} & $81.0$ \p{0.8}& $73.4$ \p{0.6}& $65.9$ \p{3.5}& $54.6$ \p{7.2}& $71.9$  \p{2.4} \\
      Kinetics400~\cite{kay2017kinetics} (\textit{Pre-train}) and  360+x (\textit{Fine-tune}) 	&   $\boldsymbol{85.3}$\p{1.6} & $\boldsymbol{81.8}$\p{1.6} & $\boldsymbol{74.9}$\p{2.1} & $\boldsymbol{68.1}$\p{5.7} & $\boldsymbol{58.2}$\p{10.8} & $\boldsymbol{73.7}$\p{4.2} \\
     \bottomrule
    \end{tabular}
    }
 % \vspace{-3mm}
% \end{table*}
\end{minipage}%

\end{table*}

%-------------------------------------------

\vspace{-1mm}
\paragraph{Q-to-Video retrieval results.} 

Table \ref{tab:ret} illustrates the retrieval results for the Query modality retrieve videos. In this table, \textit{A+D} denotes a set of independently trained audio and directional binaural features employed as query features. Moreover, \textit{(A+D)*} signifies the collaborative training of these features instead of treating them independently. The inter-modality retrieval results shown in Table \ref{tab:ret} clearly show the modality compliance quality of the \textit{360+x} dataset.
Besides Q-to-Video retrieval, we also performed Q-to-Audio and Q-to-Directional binaural delay experiments, details can be found in the supplementary material section \ref{sec:moreretrieval}.

\subsection{Self-supervised Representation Learning}
% \vspace{-3mm}
% Self-supervised learning is becoming increasingly attractive due to its great potential to leverage a large amount of unlabeled data to prepare models with generic representations for downstream tasks. 

\paragraph{Experiment setup.}
In this section, we investigated the impact of different self-supervised learning (SSL) methods using two engaging video pretext tasks: video pace (VP) prediction \cite{Wang20}  and clip order (CO) shuffle prediction \cite{xu2019self}. The VP task challenges the model to determine the pace of a video, while the CO task asks the model to rearrange shuffled video clips into their correct chronological order.
The original VP and CO primarily concentrated on video data, but to capitalise on the advantages of multi-modality, we expanded these approaches to include audio and directional binaural delay modalities. This extension was done to align modality with the temporal coherence and dynamics observed in the video. For more comprehensive explanations, please refer to the supplementary material section \ref{explainssl}.

\vspace{-1mm}
\paragraph{Experimental results.}

We first examined the impact of self-supervised learning models for video classification. Table \ref{tab:ssl+cls}\hspace{0.5mm} demonstrates the consistent precision gains achieved by utilising SSL pre-trained models. Notably, leveraging both video pace and clip order SSL techniques resulted in an average performance improvement of \textit{$\sim$ 7\%}.

% We utilize the  tasks,  for the comparison.

% Here we explore the effectiveness of applying models pre-trained on the proposed \textit{360+x} dataset for adaptation to other datasets, such as the THUMOS14 . In this section, we utilise the TriDet framework } to perform TAL task and follow the original modality setup in THUMOS14 to employ only video modality for the comparison. 

% We follow the experimental framework outlined in \cite{shi2023tridet}. When integrating with the THUMOS14 dataset, all video feature extraction procedures are executed in accordance with the I3D architecture \cite{carreira2017quo}. 

We proceeded to perform experiments using SSL pre-trained models as feature extractors for the temporal action localisation task incorporating all three modalities (V+A+D) with the TriDet framework \cite{shi2023tridet}. Since a training-from-scratch model cannot serve as the first-stage extractor, we employed the supervised extractors from section \ref{sec:classification} as a comparison. 
The summarised results in Table \ref{tab:ssl+tal}\hspace{0.5mm} indicate that pre-training with video pace (VP) or clip order (CO) individually leads to an average performance improvement of  \textit{$\sim$ 1.2\%} and \textit{$\sim$ 0.9\%} respectively on average, compared to the supervised baseline. The combination of both SSL methods yields the highest performance gain of \textit{$\sim$ 1.9\%}.

\subsection{Pre-training Model for Dataset Adaptation}
% and EpicKitchen \cite{EpicKitchen}
This section explores the efficacy of leveraging models pre-trained on the \textit{360+x} dataset for adaptation to other datasets like THUMOS14 \cite{THUMOS14}. By adhering to THUMOS14 setup, the experiments use TriDet framework \cite{shi2023tridet} for conducting Temporal Action Localisation (TAL).

The performance of this experiment, specifically the mean average precision (mAP) scores covering IoU thresholds from 0.3 to 0.7, are presented in Table \ref{tab:thumos}. As outlined by the results, 
exclusive reliance on \textit{360+x} video data for training showcases the potential for enhanced performance as compared to training solely based on the Kinetics400 dataset \cite{kay2017kinetics}. Remarkably, this performance improvement becomes more prominent at higher IoU thresholds. 
The utmost optimal performance, however, emerges through a two-step approach, commencing with pre-training on the Kinetics400 dataset followed by fine-tuning on the \textit{360+x} dataset with an average $\sim$ \textit{4.2\%} improvement compared to solely Kinetics400 pre-trained extractor. This finding showcases that the employment of the \textit{360+x} dataset for feature extractor training can be beneficial for dataset adaptation in sub-stream tasks.
More results on dataset integration are available in the supplementary material section \ref{sec:migration}.

\section{Conclusions}
% \vspace{-0.05cm}

In this work, we studied the problem of panoptic scene understanding and presented, to our knowledge, the first-of-its-kind dataset -- \textit{360+x} to support the study. The proposed \textit{360+x} is a large-scale multi-modal dataset that consists of several different viewpoints (\eg egocentric, third-person-view, and panoramic view) and covers various real-world activities in real daily life. With the most possibly available perspectives describing a real-world scene, \textit{360+x} aims to support the research in understanding the world around us in a way that humans understand (and even beyond).
Additionally, we also presented a benchmark study of several scene understanding tasks based on this newly collected dataset, with a comparison to other existing datasets. Extensive experimental analysis validated the effectiveness of each of the perspectives within our dataset, and also suggested interesting insights, confirming that with more viewpoints or data modalities, the understanding of a scene could be more comprehensive. Surprisingly, models trained without manual annotation (\ie self-supervised learning) on our dataset even perform better than those trained with human annotations in a fully supervised manner.
We hope this new dataset could bring in new directions towards scene understanding and look forward to the research on them.

% \vspace{-0.1cm}
\subsection*{Acknowledgement}
% \vspace{-0.1cm}
This project was partially supported by the Ramsay Research Fund, and the Royal Society Short Industry Fellowship (SIF\textbackslash R1\textbackslash231009).
Y. Hou and C. Qu were partially supported by the CSC grant (No.202308060328) and Allsee Technologies Ltd., respectively.
The computations described in this research were performed using the Baskerville Tier 2 HPC service\footnote{https://www.baskerville.ac.uk/} (funded by EP/T022221/1 and EP/W032244/1) and is operated by Advanced Research Computing at the University of Birmingham. 
% Baskerville was funded by the EPSRC and UKRI through the World Class Labs scheme (EP/T022221/1) and the Digital Research Infrastructure programme (EP/W032244/1) and is operated by Advanced Research Computing at the University of Birmingham.
% The computations in this research were performed using the Baskerville Tier 2 HPC service\footnote{https://www.baskerville.ac.uk/}. Baskerville was 
%  funded by EPSRC and UKRI through World Class Labs (EP/T022221/1) and Digital Research Infrastructure (EP/W032244/1) and is operated by Advanced Research Computing at the University of Birmingham.

% \newpage

{
    \small
    \bibliographystyle{ieeenat_fullname}
    \bibliography{main}  % .bbl
    % \bibliography{main.bbl}  % .bbl
}

\newpage

% \input{supp}

\clearpage
\maketitlesupplementary

% Reset the Section and using alphabet
\setcounter{section}{0}
\renewcommand{\thesection}{\Alph{section}}   % Alphabet Section

\maketitle
\appendix

\section*{Introduction}
This document provides supplementary materials for the
main paper. Specifically, section~\ref{supp:organization} describes the data organisation in detail, while section \ref{scene} explains the procedure used to select the scene labels and the temporal segmentation labels. More statistics of the proposed dataset are presented in section~\ref{AdditionalDA}. The ethical use of the dataset and the author's statement are discussed in section \ref{ethical}. 
Self-supervised methods and modality feature fusion methods employed in our work are introduced in section \ref{explainssl} and section \ref{fusion}, respectively. 
Additional experimental results are presented in section \ref{experience}, and more samples from the dataset are shown in section \ref{moreexample}. The social impact of the proposed dataset and the limitations of this work are analysed in section \ref{social} and section \ref{limit}, respectively. Potential future work is discussed in section \ref{future}.

% \paragraph{Website.} 
% The dataset and benchmark are publicly available online, specifically at the project website (\href{https://x360dataset.github.io/}{https://x360dataset.github.io}) and the GitHub repository  (\href{https://github.com/x360dataset/x360dataset-dev}{https://github.com/x360dataset/x360dataset-dev}). 

\vspace{-3mm}
\paragraph{License.} 

The \textit{360+x} dataset is licensed under the Creative Commons Attribution-NonCommercial-ShareAlike 4.0 International Public License. %To view a copy, visit \href{http://creativecommons.org/licenses/by-nc-sa/4.0/}{here}. 

\vspace{-3mm}
\paragraph{Author statement.}
The authors acknowledge that they are fully responsible for any potential violations of rights, ethical issues, or legal disputes related to their work. The authors further confirm that they have obtained all necessary permissions and licenses for the data used in the research.

\section{\textit{360+x} Dataset Organisation}
\label{supp:organization}

For each data instance, we provide a comprehensive set of views, including:

\begin{itemize}[wide]
  \item \du\  panoramic view
  \item Third-person front view 
  \item Egocentric binocular view
  \item Egocentric monocular view
\end{itemize}

For each view, we offer a variety of data modalities and the original file, allowing for a more comprehensive understanding of the scene, which is structured as follows:

\begin{itemize}[wide]
  \item Video\footnote{A set of continuous frames without audio.}
  \item Multi-channel audio
  \item Directional binaural delay 
  \item Temporal segments label 
\end{itemize}

Along with the data instance,  we also provide accompanying metadata including scene category labels, textual scene descriptions, weather conditions, capture time, and GPS information. This provides an opportunity for exploring a comprehensive understanding of the scene from various angles.

\paragraph{Accessibility.}

Large-scale data collection can present challenges for researchers due to limitations in hardware resources such as storage and computing power. To address this, we offer a three-step solution:

\begin{itemize}[wide]
  \item Partitioned data: We provide standardised mini-sets of data for quick overviews and initial experimentation, allowing researchers to explore the dataset without being overwhelmed by its size.
  \item Reduced-resolution: We offer reduced-resolution versions of our extracted frame-by-frame images, which can be used to speed up exploration of the data in the early stages of research. The original high-resolution images are also available for those who require them.
  \item Pre-computed features: We provide pre-computed features such as video and audio features, which have been extracted using the methods described in the main paper. These features offer a convenient and efficient way for researchers to access and analyse the data without having to perform extensive processing.
\end{itemize}

\section{Selection of Scene Label and Temporal Segmentation Label}
\label{scene}

The scene labels in \textit{360+x} dataset aim to represent common real-world environments and activities people routinely experience in daily life. During data collection, we strived to capture diverse scenarios across different locations that resemble natural experiences. The categories emerged organically from the range of spaces and events we were able to access and record.

For the classification of database, it is generally based on scenes \cite{zhou2017scene,zhou2017places,xiao2010sun,zhou2017scene} or action behaviours \cite{soomro2012ucf101,kuehne2011hmdb,caba2015activitynet,monfort2019moments}. However, considering that scene locations and activities often overlap, for example, `\textit{speaking}' can occur in `\textit{dining \& food outlets}' or `\textit{indoor residential spaces}', and even in the same location `\textit{campus}' may have various actions such as `\textit{walking}' and `\textit{speaking}'. Our multi-modal data set is based on video recordings of natural behaviours in natural scenes. Each video contains rich naturally occurring behavioural information and scene information, to annotate the video more completely and efficiently, we divide the scene and behavioural actions into two layers of labels: scene labels and temporal segmentation labels. 

Scene labels are based on the place where the scene occurs. We learn from the places dataset \cite{zhou2017places}, which extracts 401 scenes based on wordnet~\cite{miller1995wordnet}. However, those scenes are not all common in daily life scenes, such as `\textit{archaeological excavation}', `\textit{server room}', \etc The division is also more detailed, such as `indoor residential spaces' can have multiple categories: `\textit{bedroom}', `\textit{living room}', `\textit{dining room}', `\textit{attic}', \etc Therefore, in order to more accurately fit daily life, we put these 401 directories into the large language model~\cite{openai2023chatgpt} for classification and summary, and then through manual screening, we finally obtained 28 categories including indoor and outdoor. After the scene categories were confirmed, we collected several videos for each category, considering a balanced contribution of weather, lightness and captured locations.

Temporal segmentation labels are the behavioural activities that occur in the scene. We obtained the time segmentation tags of the \textit{360+x} database based on the activity level standard of ActivityNet \cite{caba2015activitynet} and combined them with the actual activities in the collected videos. Then we sampled about 50 videos from each directory and performed label pre-annotation. After about two rounds of pre-annotation, we analysed the differences between labels and the length of timeline coverage of each annotation, and then generated a temporal segmentation labels dictionary. To capture the diversity and granularity of activities within each category, we defined a total of 38 action instance labels covering specific actions and behaviours. Finally, we selected three professional annotators to annotate all the videos in the database according to the dictionary.

% We recognize there may be some ambiguity and overlap between the location-based categories (e.g. Pub, Kitchen) and activity-based ones (e.g. Social, Celebration). To clarify, location categories refer specifically to the physical setting, irrespective of the activities within. In contrast, activity categories denote the nature of human behaviours and interactions, regardless of the location. For instance, our dataset does not contain examples of working from home in living rooms or kitchens, which would blend location and activity labels. By focusing on ordinary events in their conventional environments, we aimed to minimize ambiguity and capture the essence of each category.

% We also avoided ambiguous blends of multiple categories within one scenario. The recordings emphasize commonplace occurrences in typical settings. Through concentrating on regular daily events in their expected environments, we sought to reduce confusion between the location-based and activity-based labels.

% Although we aimed to minimize sampling bias during data collection, we acknowledge that some biases may still exist in the current dataset version. For example, the videos were primarily recorded in urban or college town areas, which underrepresents rural environments. The scenes also do not fully capture the diversity of geographic regions across different countries and cultures. Additionally, the volunteer data collectors may unconsciously select scenarios that align with their own preferences and experiences, introducing individual biases.

\begin{figure}
    \centering
    \begin{subfigure}[b]{0.49\textwidth}
        \centering
        \includegraphics[width=\textwidth]{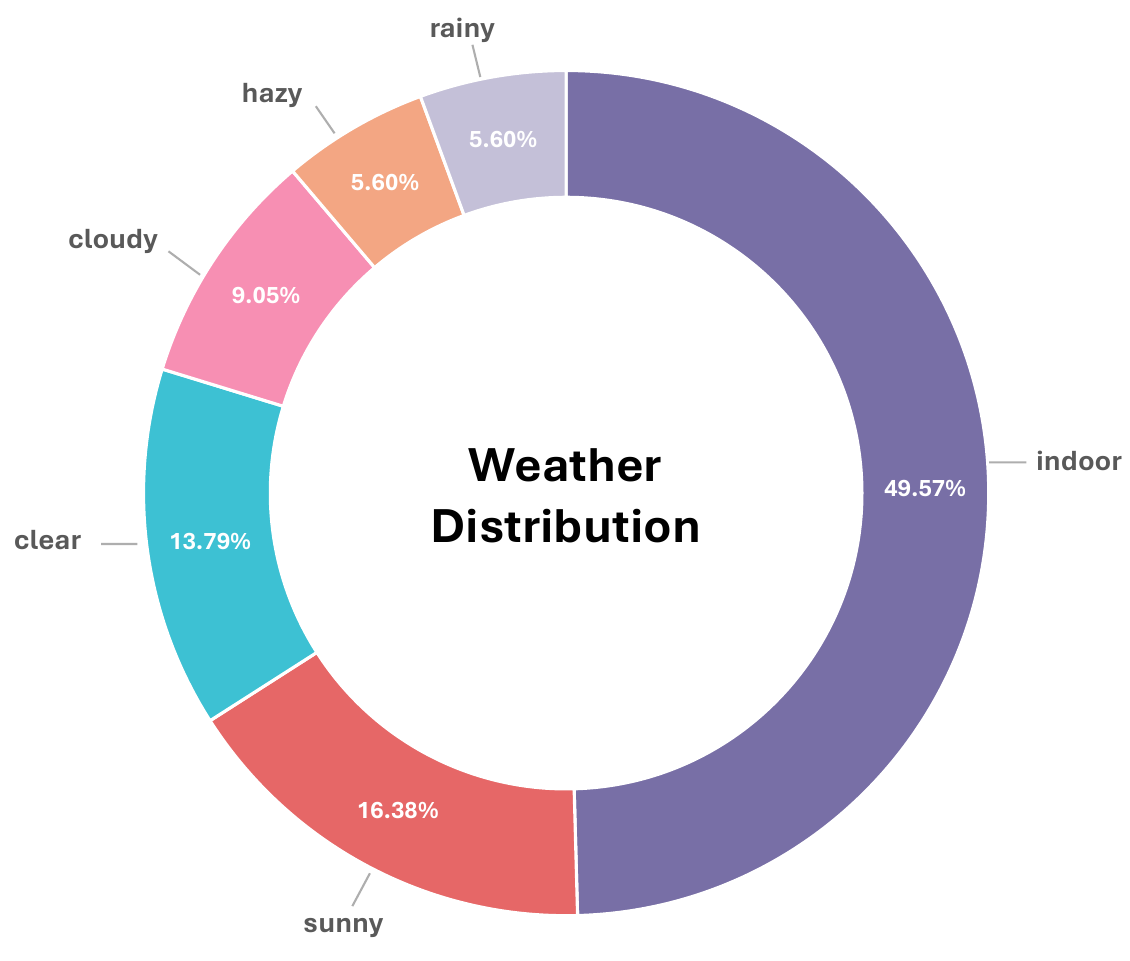}
        \caption{Overall weather distribution}
        \label{fig:weather}
    \end{subfigure}
    \hfill
    \begin{subfigure}[b]{0.49\textwidth}
        \centering
        \includegraphics[width=\textwidth]{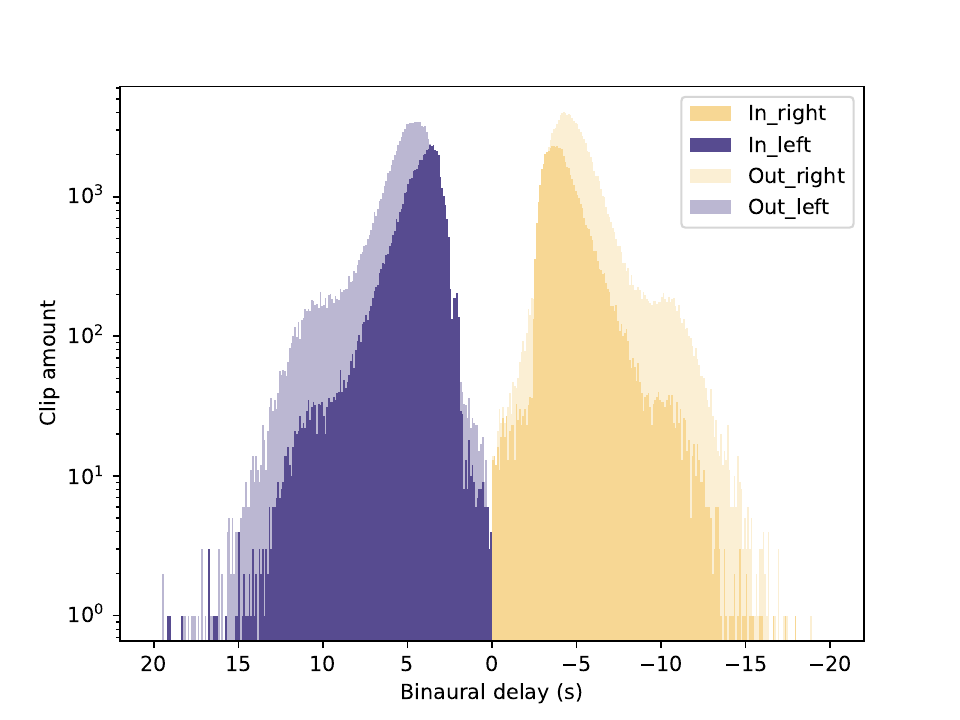}
        \caption{Overall binaural delay}
        \label{fig:time_delay}
    \end{subfigure}
    \caption{Additional dataset indoor/outdoor statistics.
    % , (a) weather distribution, (b) overall binaural delay.
    \vspace{-0.5cm}
    }
    \label{fig:TSL_and_complexity}
\end{figure}

\section{Additional Dataset Statistics}
\label{AdditionalDA}

Beyond the action and scene categories mentioned in the main paper section \ref{sec:3.3DataAnnotation}, we also include weather tags. As illustrated in Figure \ref{fig:weather}, we collected data from both outdoor and indoor environments. For those \textit{purely} indoor scenes that cannot tell any weather conditions, we label them as `\textit{indoor}' tag, while for outdoor scenes or some indoor scenes that can tell the weather, we further categorise them into `\textit{sunny}', `\textit{clear}', `\textit{cloudy}', `\textit{hazy}' and `\textit{rainy}'. Figure \ref{fig:time_delay} represents the balanced clip histogram distribution of binaural delay in both indoor and outdoor environment.

   % L D R U
    % \includegraphics[width=0.97\textwidth, trim=10 25 88 27, clip]

%--------------------------------------------
\begin{figure*}[!t]

    \centering
    
    \includegraphics[width=\textwidth]{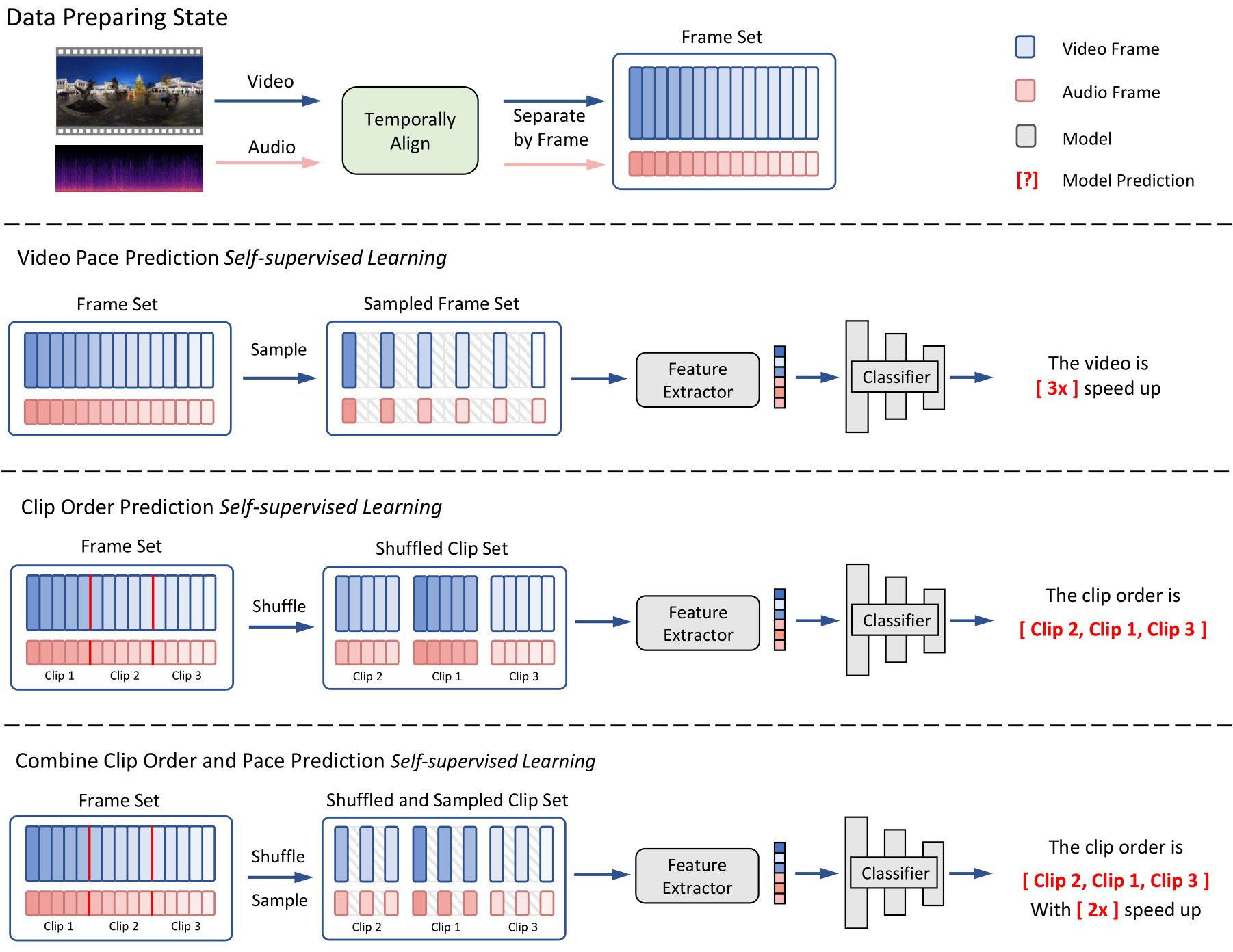}
    \caption{Elucidation of the self-supervised learning (SSL) techniques employed in our study: within SSL, audio is treated in tandem with video frames. To illustrate, when the video speed is augmented by a factor of 2, the audio sample rate is attenuated by 2 (thus speeding it up) to maintain synchronisation. Correspondingly, if the sequence of video clips is rearranged, the audio clips undergo a commensurate reshuffling. The processing of ITD data mirrors this approach used for audio data.}
\label{fig:ssl}  
\end{figure*}

%--------------------------------------------

\section{Privacy and Ethics}
\label{ethical}

We acknowledge data collectors have ethical obligations and standards to uphold when conducting data collection efforts. While specifics vary per site, three common obligations and guidelines have been followed:

\begin{enumerate}
\item Compliance with legal terms and consortium conditions of use, specifically for research purposes only.
\item Protection of participant confidentiality and privacy. 
\item Avoidance of sensitive areas to prevent any potential breaches of confidentiality.
\end{enumerate}

% The collecting partner holds consent forms and/or release forms for all videos.  The data has been manually de-identified to remove personally identifiable information (PII) and reviewed pre-release.

\paragraph{Sensitive information processing.}
To protect the privacy of individuals, we use an automated face-blurring tool, \textit{Deface}\footnote{https://github.com/ORB-HD/deface}, to redact personally identifiable information (PII) from the videos. \textit{Deface} employs the CenterFace \cite{xu2020centerface} face detection model to identify facial regions in frames, then applies Gaussian blurring to mask each detected face.

While completely removing faces could maximise privacy, blurred faces retain some visual information and context. The blurring parameters were tuned to balance privacy protection and data utility based on established practices \cite{cheng2021can}. All videos were manually reviewed post-redaction to catch any errors or missings detection. 

Despite our efforts to maintain efficiency and consistency, certain limitations exist. Factors such as occlusion, lighting, and face angle can affect face detection accuracy, and the blurring strength may be too weak or too strong in some instances. Additionally, our process does not address other forms of personally identifiable information like voices and text. While not perfect, our approach does reduce the privacy risk compared to fully visible faces, and allows the altered data to remain valuable for research purposes.

% ----------------------
\begin{figure*}[!t]

    \centering
    % left down right top
    \includegraphics[width=\textwidth, trim=30 210 30 10, clip]{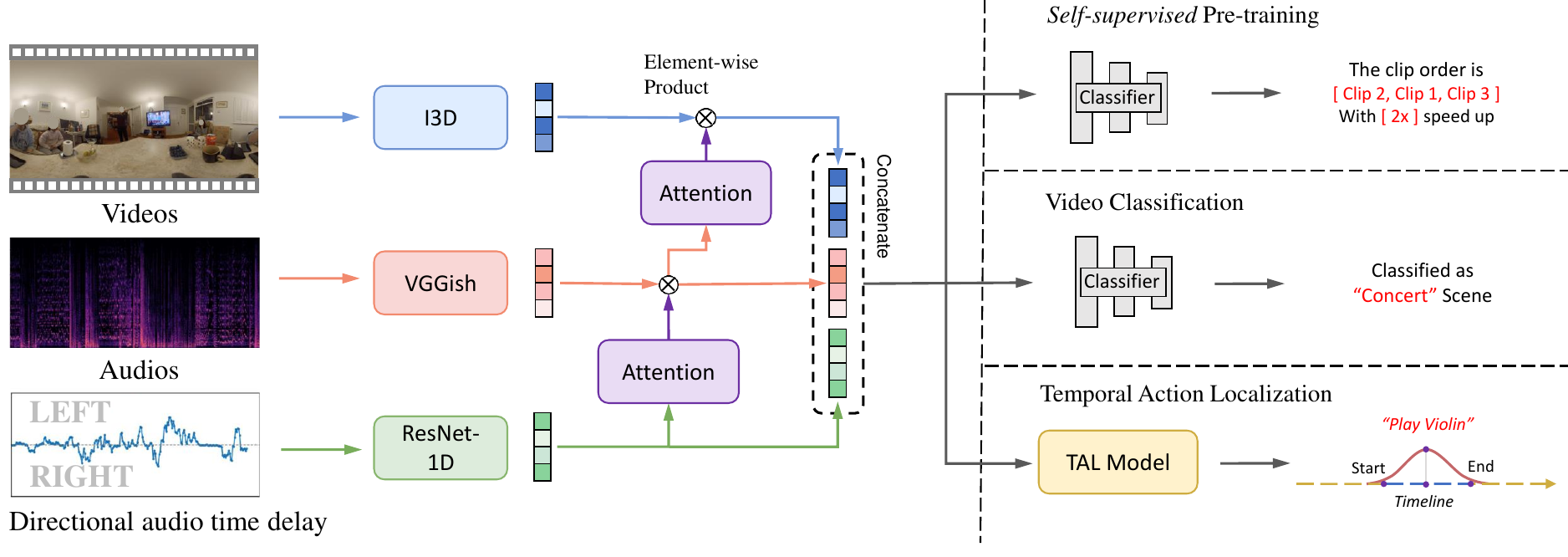}
    \caption{Illustration of Modality Fusion: The features from video, audio, and ITD are extracted utilizing I3D, VGGish, and ResNet-1D correspondingly. Subsequently, these features are concatenated for each sub-task. }

\label{fig:fusion}
\end{figure*}

% ----------------------

\section{Explain of Self-supervised Learning}
\label{explainssl}

In this study, we utilise self-supervised learning (SSL) techniques proposed in video pace (VP) prediction \cite{Wang20} and clip order (CO) shuffle prediction \cite{xu2019self} to pre-train models for enhanced feature learning and subsequent task performance. These two methods are originally tailored for video data, and involve using speed perturbation or clip order permutation on the visual content.

However, our dataset provides more modalities beyond merely video. 
To fully leverage the power of self-supervised learning, we extend these methods to incorporate more modalities (\ie audios and direction binaural delay). Figure \ref{fig:ssl} depicts how SSL methods can be applied to both video and audio modalities, while ensuring synchronisation between them.  For example, if the video playback speed is altered (\eg $\times2$), the corresponding audio sample rate is changed accordingly (\ie $\times0.5$) to maintain synchronisation. Similarly, when the sequence order of video clips is shuffled, the order of audio clips is also rearranged identically to preserve alignment. The direction binaural delay data, which contains spatial audio information, undergoes similar synchronised transformations during SSL pre-training as the audio data. By treating all three modalities (\ie video, audio, and direction binaural delay) jointly and applying transformations consistently across them, we enable cross-modal coordination and representation learning.

It is noteworthy that the VP and CO primarily focus on leveraging temporal information as training guidance, applying it either globally (pace) or locally (clip) to offer distinct interventions to this temporal data. By combining these interventions, there is potential to enhance the model's capability to capture global and local temporal dependencies simultaneously. This integration, depicted in Figure \ref{fig:ssl}, is delineated as `combine clip order and pace prediction' or varied pace clip order (VP+PO) shuffle. This integration is highlighted in our experiments detailed in the main paper Tables \ref{tab:ssl+cls}\hspace{0.06cm} and \ref{tab:ssl+tal}\hspace{0.06cm}, where noted benefits become evident.

In summary, a core aspect of our self-supervised multimodal learning approach is ensuring aligned cross-modal augmentations and fusing representations across video, audio, and spatial audio domains. This provides a strong foundation for the multi-modal benchmarks in our work.

\section{Explain of  Modalities Fusion}
\label{fusion}
Simply concatenating the modalities without proper fusion can lead to a reduction in the benefits of multi-modal learning, as pointed out in \cite{wang2020makes}. Therefore, instead of solely concatenating modality features, we leverage a hierarchical attention mechanism for multi-modality integration as depicted in Figure \ref{fig:fusion}. To simplify the illustration, we use V - video, A - audio, and D - direction binaural delay data as simplified symbols representing each modality.

In nature of multi-modality,  the direction binaural delay data contains spatial audio information, and audios can indicate the rich movement region to the videos. We design the hierarchical attention with D as an attention query to direct focused attention towards A. Afterwards, A is also leveraged as a query to attentively interact with V. The experimental supports for selecting A as the attention medium is also presented in section \ref{results}. This hierarchical design enables the encapsulation of directional and spatial information into audio and video modalities, creating a synergistic representation of the underlying data that integrates the features across modalities.

\section{More Experiment Results}
\label{experience}

\subsection{Temporal Action Localisation}
\label{sec:tal_views}
As a supplement to section \ref{sec:tal} in the main paper, we expand the experiments to variations of views, as detailed in Table \ref{table:TAL_extra}. The results therein show a trend consistent with those observed in Table \ref{tab:cls} in the main paper, indicating that the utilisation of multiple views contributes positively.

\begin{table}[h]

	\centering
	% \caption{\small TAL results for different view combinations using TriDet, with extractors being \textit{I3D} pretrained on \textit{360+x}. The lines with a gray background were reported in the main paper.}
	\caption{\small TAL results for different views using TriDet, with extractors being \textit{I3D} pretrained on \textit{360+x}. The lines with a \textcolor{gray}{grey background} were reported in the main paper.}
    \label{table:TAL_extra}
	\small
\vspace{-0.3cm}

\resizebox{\columnwidth}{!}{
    \begin{tabular}{@{}l|cccc|cccc|cccc@{}}
        \toprule
     
        \multirow{3}{*}{Selected View} & \multicolumn{4}{c}{V}& \multicolumn{4}{c}{V+A}& \multicolumn{4}{c}{V+A+D} \\
         & mAP & mAP & mAP & \multirow{2}{*}{Avg.\ }& mAP & mAP & mAP & \multirow{2}{*}{Avg.\ }& mAP & mAP & mAP & \multirow{2}{*}{Avg.\ }\\
         & $@$0.5 & $@$0.75 & $@$0.95 && $@$0.5 & $@$0.75 & $@$0.95 && $@$0.5 & $@$0.75 & $@$0.95 &\\
        
         \midrule
         
           \multirow{1}{*}{\makecell[c]{Egocentric Only}} &
     $ 12.5$ & $ 9.8$ & $4.3$ & $8.9$ \!\e{0.0}&$ 16.2$ & $ 12.3$ & $4.6$ & $11.0$ \!\e{0.0}&$ 16.9$ & $ 12.7$ & $4.7$ & $11.4$ \!\e{0.0}

 \\
            
           \multirow{1}{*}{\makecell[c]{Front Only}}  &
          $ 19.7$ & $ 14.4$ & $5.2$ & $13.1$ \!\p{4.2}&$ 24.5$ & $ 17.6$ & $6.1$ & $16.1$ \!\p{5.1}&$ 25.6$ & $ 18.0$ & $6.2$ & $16.6$ \!\p{5.2}
\\
             \cellcolor{gray!25} 
              \multirow{1}{*}{\makecell[c]{\du Only}} & \cellcolor{gray!25} 
        $ 21.1$ & \cellcolor{gray!25}  $ 15.3$ & \cellcolor{gray!25}  $5.5$ & \cellcolor{gray!25}  $14.0$ \!\p{5.1}& \cellcolor{gray!25} $ 26.4$ & \cellcolor{gray!25}  $ 18.5$ & \cellcolor{gray!25}  $6.9$ & \cellcolor{gray!25}  $17.3$ \!\p{6.3}& \cellcolor{gray!25} $ 27.1$ & \cellcolor{gray!25}  $ 18.7$ & \cellcolor{gray!25}  $7.0$ & \cellcolor{gray!25}  $17.6$ \!\p{6.2}

        % { \multirow{1}{*}{\makecell[c]{\du Only}} &
        % $ 21.1$ & $ 15.3$ & $5.5$ & $14.0$ \!\p{5.1}&$ 26.4$ & $ 18.5$ & $6.9$ & $17.3$ \!\p{6.3}&$ 27.1$ & $ 18.7$ & $7.0$ & $17.6$ \!\p{6.2} }
\\

           \multirow{1}{*}{\makecell[c]{\du + Egocentric}} &
         $ 21.4$ & $ 15.8$ & $5.7$ & $14.3$ \!\p{5.4}&$ 27.3$ & $ 19.2$ & $7.2$ & $17.9$ \!\p{6.9}&$ 27.8$ & $ 19.6$ & $7.2$ & $18.2$ \!\p{6.8}
\\

           \multirow{1}{*}{\makecell[c]{\du + Front}} &
           $ 24.2$ & $ 16.8$ & $6.1$ & $15.7$ \!\p{6.8}&$ 28.1$ & $ 20.3$ & $7.3$ & $18.6$ \!\p{7.6}&$ 28.2$ & $ 20.8$ & $7.3$ & $18.8$ \!\p{7.4}
\\

           \multirow{1}{*}{\makecell[c]{\du + Front + Ego}} &
       \boldsymbol{  $ 24.6$ }& \boldsymbol{$ 17.1$} & \boldsymbol{$6.3$ }& \boldsymbol{$16.0$ }\!\textbf{\p{7.1}}& \boldsymbol{$ 28.2$} & \boldsymbol{$ 20.6$ }& \boldsymbol{$7.3$ }&\boldsymbol{ $18.7$ }\!\p{7.7}&\boldsymbol{$ 28.8$ }& \boldsymbol{$ 21.0$}& \boldsymbol{$7.4$ }&\boldsymbol{ $19.1$} \!\textbf{\p{7.7}} \\
     \bottomrule
    \end{tabular}
    }
    
\vspace{-0.2cm}
\end{table}

\subsection{Modality Fusion}
\label{sef:modalityfusion}
We also explored alternative modality fusion approaches, such as direct concatenation of modalities, concatenation followed by a linear layer, concatenation followed by self-attention, and varied hierarchical structures of hierarchical attention. The performance of these fusion methods on Temporal Action Localisation is systematically compared and presented in Table \ref{table:TAL_ablation}, suggesting the effectiveness of our presented hierarchical attention approach.

\begin{table}[h]
	\centering
	% \caption{ \small TAL resultsusing TriDet, with extractors being \textit{I3D} pretrained on \textit{360+x}. The experiments are conducted under 360+Egocentric+F and V+A+D. The notation X$\to$Y represents X as the query and Y as the key-value pair in the attention mechanism.}
	\caption{ \small TAL with TriDet, \textit{I3D} pretrained on \textit{360+x}, under the setting 360+Egocentric+F and V+A+D. X$\to$Y: X as the query and Y as the key-value pair in the attention mechanism.}
    \label{table:TAL_ablation}
	\small
\vspace{-0.3cm}

\resizebox{\columnwidth}{!}{
    \begin{tabular}{@{}l|cccc@{}}
        \toprule
     
        \multirow{1}{*}{Feature Fusion}

         & mAP$@$0.5  & mAP$@$0.75  & mAP$@$0.95  & \multirow{1}{*}{Avg.\ }\\
         % & & & \\

 \midrule
          
 Concatenation&$ 19.2$\!\e{$0.0$} & $ 14.6$\!\e{$0.0$} & $ 5.3$\!\e{$0.0$} & $ 13.0$\!\e{$0.0$} \\
Concat + Linear Layer&$ 21.2$\!\p{$2.0$} & $ 15.1$\!\p{$0.5$} & $ 5.5$\!\p{$0.2$} & $ 13.9$\!\p{$0.9$}   \\
Concat + Self-Attention&$ 26.9$\!\p{$7.7$} & $ 18.9$\!\p{$4.3$} & $ 6.8$\!\p{$1.5$} & $ 17.5$\!\p{$4.5$}  \\
D$\to$V + Concat A&$ 17.8$\!\m{$1.4$}& $ 13.8$\!\m{$0.8$}& $ 5.2$\!\m{$0.1$}& $ 12.3$\!\m{$0.8$}\\
D$\to$A + Concat V&$ 24.6$\!\p{$5.4$} & $ 17.2$\!\p{$2.6$} & $ 6.2$\!\p{$0.9$} & $ 16.0$\!\p{$3.0$}   \\
A$\to$D + Concat V&$ 20.5$\!\p{$1.3$} & $ 14.9$\!\p{$0.3$} & $ 5.7$\!\p{$0.4$} & $ 13.7$\!\p{$0.7$}   \\
A$\to$V + Concat D&$ 28.3$\!\p{$9.1$} & $ 20.6$\!\p{$6.0$} & $ 7.3$\!\p{$2.0$} & $ 18.7$\!\p{$5.7$}  \\

\midrule

Hierarchical Attention, D$\to$A, A$\to$V&\boldsymbol{$28.8$}\!\textbf{\p{$9.6$}}&\boldsymbol{$21.0$}\!\textbf{\p{$6.4$}} &\boldsymbol{$7.4$}\!\textbf{\p{$2.1$}} &\boldsymbol{$19.1$}\!\textbf{\p{$6.0$}}

\\

     \bottomrule
    \end{tabular}
    }
    
\vspace{-0.2cm}
\end{table}

\subsection{Cross-modality Retrieval}
\label{sec:moreretrieval}

As we mentioned in the main paper section \ref{sec:retrieval}, we are embarking on a series of retrieval tasks that traverse the audio, video and directional time delay modalities. This section provides more experimental results on Query-to-Audio and Query-to-Directional information results.

\paragraph{Q-to-Audio retrieval results.} 

Table \ref{tab:audioretrival} illustrates the retrieval results for the retrieving audios. In this table, the notation \textit{V+D} represents a set of video and directional binaural features that are trained independently. Additionally, the superscript \textit{*} indicates that these features are collaboratively trained rather than being treated separately. 

The query \textit{V+D} exhibits superior audio retrieval performance, surpassing the use of videos alone. Additionally, the suppression of \textit{(V+D)*} suggests that the modalities V and D are not directly related, which forms the foundation for designing our hierarchical attention mechanism that employs audio modality as the attention medium.
\label{results}

% \vspace{-3mm}
\begin{table}[h]
	\centering
    % \vspace{-1mm}  ''D'':
\caption{Q-to-Audio retrieval results. The superscript* indicates modalities are co-trained. Recall reported with rank in $\{1, 5, 10\}$. }
\resizebox{\columnwidth}{!}{
\begin{tabular}{@{}l|ccc@{}}
     \toprule
    Query Modality	& R1 (\%) 	& R5 (\%) & R10 (\%)  \\
      \midrule
       V   &      $54.17$ \e{0.00}  &   $68.32 	$ \e{0.00} &    $80.72$ \e{0.00}\\  
      V + D &      \boldsymbol{$66.36 $} \p{12.19} &    \boldsymbol{$76.78  $} \p{8.46} &  \boldsymbol{$88.59$} \p{7.87}   \\
    (V + D)*  &   $59.21$ \p{5.04} &   $72.65	$  \p{4.33} &    $86.84$ \p{6.21}   \\
     \bottomrule
\end{tabular}
}
\label{tab:audioretrival}
\vspace{-0.5cm}
\end{table}

% --------------------------

% --------------------------

\paragraph{Q-to-Directional feature retrieval results.} 
Table \ref{tab:directionalretrival} illustrates the retrieval results for the Query modality retrieve directional features. In this table, the notation \textit{V+A} represents video and audio, respectively. The query \textit{(V+A)*} exhibits better directional feature retrieval performance than other queries. The effective retrieval results across modalities demonstrate the high quality and compliance with the modalities of the \textit{360+x} dataset.

% -----------------

\begin{table}[h]
	\centering
    % \vspace{-1mm}  ''D'':
\caption{Q-to-Directional binaural delay retrieval results. The superscript* indicates modalities are co-trained. Recall reported with rank in $\{1, 5, 10\}$. }
\vspace{-0.1cm}
\label{tab:ret}
\resizebox{\columnwidth}{!}{
\begin{tabular}{@{}l|ccc@{}}
        \toprule
    Query Modality	& R1 (\%) 	& R5 (\%) & R10 (\%)  \\
      \midrule
        V   &      $6.02	$ \e{0.00} &   $17.64	$ \e{0.00} &    $25.93$\e{0.00}\\  
      V + A &        $ 	54.15	 $ \p{48.13} &     $76.10 $ \p{58.46} &    $	90.32$ \p{64.39}   \\
    (V + A)*  &   \boldsymbol{$67.26	$} \p{61.24} &    \boldsymbol{$89.47 $} \p{71.83}  &  \boldsymbol{$94.26$}  \p{68.33}  \\
     \bottomrule
    \end{tabular}
    }
\label{tab:directionalretrival}
% \vspace{-1mm}
\vspace{-0.2cm}
\end{table}

\begin{table*}

\begin{minipage}{1.0\textwidth}
% \begin{table*}
	\centering
		\caption{The test outcomes for the \textbf{\textit{verb}} sub-task within the EPIC-Kitchens dataset \cite{damen2018scaling}. We utilise the ego-centric monocular modality for training as the sole source of feature extraction. PT: pre-train, FT: Fine-tune.}
\label{tab:kitchen}
 \vspace{-2mm}
\resizebox{\textwidth}{!}{
 \begin{tabular}{@{}l|c|c|c|c|c|c@{}}
     \toprule
    Feature Extractor & mAP$@0.3$ 	& mAP$@0.4$ & mAP$@0.5$ & mAP$@0.6$& mAP$0.7$& Avg. \\
      \midrule
       EPIC-Kitchens dataset \cite{EpicKitchen}  &     $28.6 	$\e{0.0}   &  $27.4 	$ \e{0.0}  &  $26.1$ \e{0.0}  &  $ 	24.2 	$\e{0.0}   &  $20.8 	$ \e{0.0}  &  $25.4$ \e{0.0}  \\   
        \textit{360+x}   Dataset	&  $	28.1 $ \m{0.5} &  $27.1	$\m{0.3}  &  $25.9 $ \m{0.2}  &  $	24.3 	$ \p{0.1}  &  $21.2	$ \p{0.4}  &  $25.3$\m{0.1} \\ 
        
      \textit{360+x} (PT), Epic-Kitchens (FT)	 &        \boldsymbol{$28.8$} \p{0.2}  &  \boldsymbol{$27.8$} \p{0.4}   &  \boldsymbol{$26.5$} \p{0.4}  &  \boldsymbol{$24.9$} \p{0.7}   &  \boldsymbol{$ 21.7 $}\p{0.9}   &  \boldsymbol{$25.9$}\p{0.5}   \\

     \bottomrule
    \end{tabular}
}
% \vspace{-0.1cm}
% \end{table*}
\end{minipage}
% \hfill
\end{table*}

\begin{table*}
% EpicKitchen [1] 	
% EpicKitchen [1] Pretrain, 360+x Fine-tune (Ego-centric Only) 	
% 360+x (V+A+D Jointly) Pretrain, EpicKitchen [1] Fine-tune 	
% 360+x (Ego-centric Only) Pretrain, EpicKitchen [1] Fine-tune 	
 % \vspace{2mm}
\begin{minipage}{1.0\textwidth}
% \begin{table*}
	\centering
		\caption{The test outcomes for the \textbf{\textit{noun}} sub-task within the EPIC-Kitchens dataset \cite{damen2018scaling}. We utilise the ego-centric monocular modality for training as the sole source of feature extraction.}
\label{tab:kitchen2}
 \vspace{-2mm}
\resizebox{\textwidth}{!}{
 \begin{tabular}{@{}l|c|c|c|c|c|c@{}}
     \toprule
    Feature Extractor & mAP$@0.3$ 	& mAP$@0.4$ & mAP$@0.5$ & mAP$@0.6$& mAP$0.7$& Avg. \\
      \midrule
       EPIC-Kitchens dataset \cite{EpicKitchen}  &     $27.4 $\e{0.0}  &  $26.3 	$ \e{0.0}  &  $24.6$\e{0.0}  &  $ 	22.2$\e{0.0}  &  $18.3$\e{0.0} &  $23.8$\e{0.0}  \\   
        \textit{360+x}  Dataset	&  $	26.9 $ \m{0.5} &  $	26.0 	$\m{0.3}  &  $24.4 	$ \m{0.2} &  $22.3 $\p{0.1}  &  $	18.6 	$\p{0.3}  &  $23.7$\m{0.1}\\  
      \textit{360+x} (PT), Epic-Kitchens (FT)	 &        \boldsymbol{$27.9 	$} \p{0.5}  &  \boldsymbol{$26.9 	$} \p{0.6} &  \boldsymbol{$25.4$} \p{0.8} &  \boldsymbol{$23.2$} \p{1.0} &  \boldsymbol{$ 	19.3	$}  \p{1.0}&  \boldsymbol{$24.5$}\p{0.7}  \\
    % 	 	 	 	24.4
     \bottomrule
    \end{tabular}
    }
\end{minipage}

\end{table*}

\subsection{Migration of the Dataset Pre-training Model}
\label{sec:migration}
Regarding the integration with the  EPIC-Kitchens \cite{damen2018scaling} dataset, we follow the experiment setup in \cite{zhang2022actionformer} and deploy the SlowFast architecture \cite{feichtenhofer2019slowfast} for feature extraction. The outcomes of the experimentation, centred around the \textit{verb} and \textit{noun} sub-tasks within the EPIC-Kitchens dataset, are concisely displayed in Table \ref{tab:kitchen} and Table \ref{tab:kitchen2}. These tables provide a comprehensive overview of mean average precision (mAP) scores across a spectrum of IoU thresholds, spanning from 0.1 to 0.5.

In accordance with the EPIC-Kitchens ~\cite{damen2018scaling}, which offers a large amount of monocular egocentric data, we solely employ monocular egocentric information from the \textit{360+x}  for this section, thereby ensuring a consistent and reliable basis for experimental analysis.  Examining Table \ref{tab:kitchen} and Table \ref{tab:kitchen2}, the \textit{360+x} dataset extractor does not perform as well as the EPIC-Kitchens model when trained only with EPIC-Kitchens. This is likely due to the fact that the EPIC-Kitchens model is better suited for the EPIC-Kitchens dataset.  However, pre-training with the \textit{360+x} dataset followed by fine-tuning on EPIC-Kitchens \cite{damen2018scaling} results in enhanced performance when compared with training solely on the EPIC-Kitchens dataset. This observation suggests that despite the disparate data formats inherent in the two datasets, pre-training on the \textit{360+x} dataset holds the potential to contribute to improved performance within the EPIC-Kitchens context \cite{damen2018scaling}.

\subsection{Transformer-Based Backbone}
We used I3D as our backbone as it was widely adopted in video understanding tasks in the literature. However, we further explore \textit{more contemporary Transformer-based} models as our backbone, \eg VideoMAE~\cite{tong2022videomae}, pretrained on the Kinetics dataset, akin to the I3D model setting in the main paper. Table \ref{table:VideoMAE} reports the performance on temporal action localisation using VideoMAE. Compared to the results in Table \ref{tab:tal_combined} in the main paper (\ie the greyed line I3D in Table~\ref{table:VideoMAE}), Transformer shows better performance. Additionally, this experiment further validates the impact/benefits of various views and modalities.

\begin{table}[h]
	\centering
	% \caption{ \small TAL results  using TriDet with extractors being \textit{VideoMAE} pretrained on \textit{kinetics}. The first lines was reported in the main paper using \textit{I3D} extractor as reference.}
	\caption{ \small TAL using TriDet with extractors being \textit{Transformer-based} model pretrained on \textit{kinetics}. The \textcolor{gray}{greyed line} was reported in the main paper using \textit{I3D} extractor, for reference.}
    \label{table:VideoMAE}
	\small
\vspace{-0.3cm}

\resizebox{\columnwidth}{!}{
    \begin{tabular}{@{}l|cccc|cccc@{}}
        \toprule
     
        \multirow{3}{*}{Selected View} & \multicolumn{4}{c}{V}& \multicolumn{4}{c}{V+A} \\
         & mAP & mAP & mAP & \multirow{2}{*}{Avg.\ }& mAP & mAP & mAP & \multirow{2}{*}{Avg.\ }\\
         & $@$0.5 & $@$0.75 & $@$0.95 && $@$0.5 & $@$0.75 & $@$0.95&\\
        
         \midrule

          \multirow{1}{*}{\textcolor{Gray}{\makecell[c]{\du Only}, with I3D}} &
 \textcolor{Gray}{$ 16.7$\!\e{0.0}} &  \textcolor{Gray}{$ 10.1$\!\e{0.0} }&\textcolor{Gray}{  $ 4.8$\!\e{0.0} }&\textcolor{Gray}{  $ 10.5$\!\e{0.0} }&\textcolor{Gray}{ $ 23.6$\!\e{0.0} }&\textcolor{Gray}{  $ 17.2$\!\e{0.0} }& \textcolor{Gray}{ $ 6.4$\!\e{0.0} }&  \textcolor{Gray}{$ 15.7$\!\e{0.0}} 

          \\

 \midrule
          
%            \multirow{1}{*}{\makecell[c]{Egocentric Only}} &
%   $ 5.6$\!\m{11.1} & $ 4.2$\!\m{5.9} & $ 2.7$\!\m{2.1} & $ 4.2$\!\m{6.4} &$ 10.3$\!\m{13.3} & $ 7.3$\!\m{9.9} & $ 3.9$\!\m{2.5} & $ 7.2$\!\m{8.6} 

%  \\
            
%            \multirow{1}{*}{\makecell[c]{Front Only}}  &
%           $ 16.2$\!\m{0.5} & $ 12.8$\!\p{2.7} & $ 4.8$\!\e{0.0} & $ 11.3$\!\p{0.7} &$ 24.0$\!\p{0.4} & $ 16.9$\!\m{0.3} & $ 5.6$\!\m{0.8} & $ 15.5$\!\m{0.2} 

% \\
            
 \multirow{1}{*}{\makecell[c]{\du Only}} &
 $ 17.1$\!\p{$0.4$} & $ 13.4$\!\p{$3.3$} & $ 5.2$\!\p{$0.4$} & $ 11.9$\!\p{$1.4$} &$ 25.9$\!\p{$2.3$} & $ 18.5$\!\p{$1.3$} & $ 6.1$\!\m{$0.3$} & $ 16.8$\!\p{$1.1$}

 \\
        
        % { \multirow{1}{*}{\makecell[c]{\du Only}} &
        % $ 21.1$ & $ 15.3$ & $5.5$ & $14.0$ \!\p{5.1}&$ 26.4$ & $ 18.5$ & $6.9$ & $17.3$ \!\p{6.3}&$ 27.1$ & $ 18.7$ & $7.0$ & $17.6$ \!\p{6.2} }

           \multirow{1}{*}{\makecell[c]{\du + Egocentric}} &
       $ 16.9$\!\p{0.2} & $ 13.1$\!\p{3.0} & $ 5.0$\!\p{0.2} & $ 11.7$\!\p{1.1} &$ 26.4$\!\p{2.8} & $ 19.0$\!\p{1.8} & $ 6.2$\!\m{0.2}& $ 17.2$\!\p{1.5} 

\\

           \multirow{1}{*}{\makecell[c]{\du + Front}} &
           $ 19.5$\!\p{2.8} & $ 16.3$\!\p{6.2} & $ 5.6$\!\p{0.8} & $ 13.8$\!\p{3.3} &$ 27.6$\!\p{4.0} & $ 21.2$\!\p{4.0} & $ 6.5$\!\p{0.1} & $ 18.4$\!\p{2.7}

          \\

           \multirow{1}{*}{\makecell[c]{\du + Front + Ego}} &
       \boldsymbol{$19.2$}\!\textbf{\p{2.5} }&\boldsymbol{$15.8$}\!\textbf{\p{5.7}}&\boldsymbol{$5.4$}\!\textbf{\p{0.6} }&\boldsymbol{$13.5$}\!\textbf{\p{2.9}}&\boldsymbol{$27.8$}\!\textbf{\p{4.2}}&\boldsymbol{$21.7$}\!\textbf{\p{4.5} }&\boldsymbol{$6.6$}\!\textbf{\p{0.2}}&\boldsymbol{$18.7$}\!\textbf{\p{3.0}}

\\

     \bottomrule
    \end{tabular}
    }
    
\vspace{-0.3cm}
\end{table}

\section{More Data Examples}
Here we provide additional examples of the data (Figures~\ref{fig:head}  $\sim$
 ~\ref{fig:end}) to show a better understanding of the content and quality of the \textit{360+x} Dataset.

\label{moreexample}

\section{Social Impact}
\label{social}
Our contribution has the potential to positively impact \textit{scene understanding} through multi-modality learning. The proposed \textit{360+x} Dataset provides the research community with a multi-view perspective with rich modalities for \textit{scene understanding} accompanied by rigorous privacy and ethics standards. Additionally, it offers a diversity and density of activities and reproducible benchmarks for technical advances in scene understanding and beyond.

We acknowledge that large-scale data collection with inadequate oversight could raise privacy and ethical concerns. Therefore, we intend to hinder potential negative applications by making \textit{360+x} data available only for users who sign a license agreement with the statement enumerating the allowable uses of the data.

\section{Limitations}
\label{limit}
% The limitation of the \textit{360+x} dataset is mainly three-fold. 

Our dataset aims to encompass various aspects of daily life to reflect the real world, yet  we acknowledge that it still possesses certain biases and cannot fully represent all aspects of the real world. Despite our efforts to collect massive everyday videos from geographically and demographically diverse sources, the current 28 scenes and 15 cities are still far from complete coverage of the full spectrum of everyday life. Furthermore, while we have included footage from rural and field locations, the majority of the videos remain concentrated in urban or college town areas, resulting in a biased representation of reality.

Another limitation pertains to the potential for biases and noise in our data collection procedures. The unscripted nature of video capturing can introduce inconsistency noise since collectors might choose scenes based on their personal interests, leading to an incomplete or biased depiction of daily experiences. Additionally, the video capturing results are also susceptible to the location of the recorder, which may introduce geometrical bias.

Finally, there remains the potential for temporal labelling bias. While we have taken steps to minimise bias through multiple annotator merging, there still exists the possibility of variations in interpretations of the scene or temporal activities due to individual differences in knowledge backgrounds and natural language use. This can result in subtle yet potentially significant biases in the language-based narrations and action boards.

% \input{supp/sec/7_availablity}

% Finally, even though our temporal segmentation labelling is performed by multiple labellers and merged to minimize bias, there still objectively exist variations in the understanding of actions.

% \ref{fig:\theimagecounter}

\section{Future Work}
\label{future}
The \textit{360+x} dataset is a collaborative project aimed at driving forward the development of foundational AI research in the realm of panoramic multi-modal machine perception and scene understanding. We actively seek and encourage global collaborations with researchers and participants from diverse and underrepresented regions, as their contributions are critical for capturing the richness and diversity of daily life activities. Therefore, we have developed our data collection and annotation methods to be comprehensive and transparent, allowing researchers from diverse backgrounds to participate in expanding the diversity and quality of the dataset.

 % We adhere to strict privacy and confidentiality standards, ensuring that all data collection is conducted in compliance with legal terms and with the informed consent of subjects, who are aware that their data will only be used for scientific research purposes.

In addition to the current benchmarks, we plan to expand the scope of our dataset to encompass other video-audio scene understanding tasks such as audio-visual diarization, scene querying, pre/post conditions, and forecasting, which will further advance the state-of-the-art techniques in this field. However, our current dataset is lacking in spatial-temporal localisation of objects, actions, and audio sources, which we are currently working to address through the augmentation of our labelling process. Although we have made significant progress, the substantial annotation workload has postponed the completion of this task. Spatial annotations will be included in a future update.

% Also, due to occlusion, there may exist regions visible to the stereo camera but not the panoramic camera, and vice versa. To address potential label inconsistencies, we plan to provide separate labels for each camera view in a future version. We also intend to perform location alignment between the cameras to enable spatial correspondence.

To ensure the long-term utility of the dataset, we commit to providing regular updates and maintenance. This includes verifying and correcting any issues related to data accessibility and integrity, as well as expanding the dataset with new content to maintain its relevance with the latest advancements and challenges in academia and industry.

\newpage
\thispagestyle{empty}
\pagestyle{empty}    % 所有页都不显示页码

\verticalphotosHead{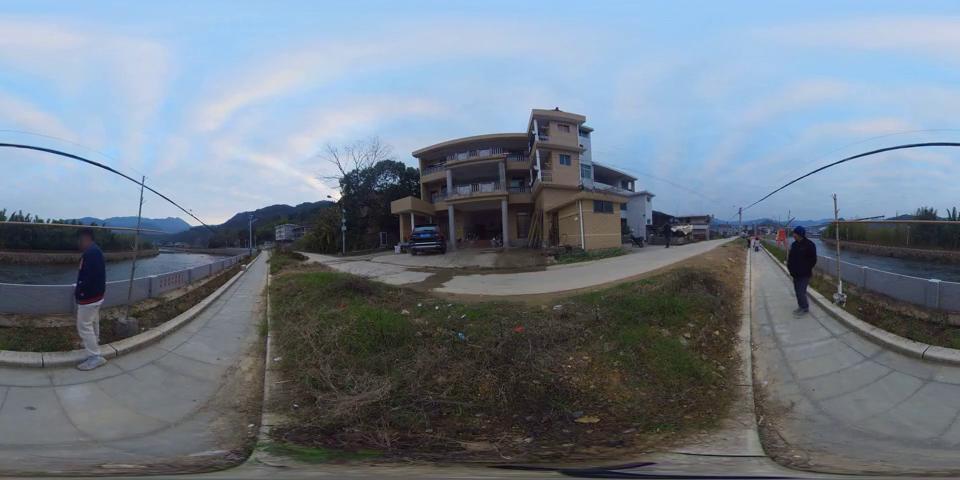}{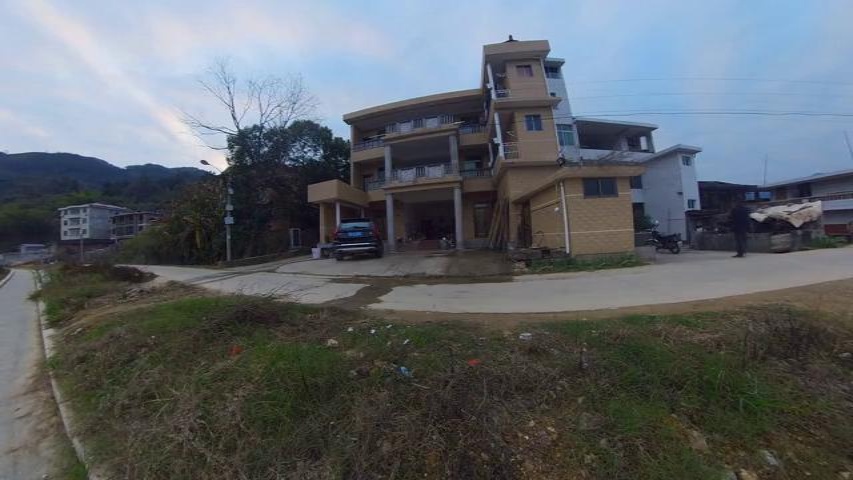}{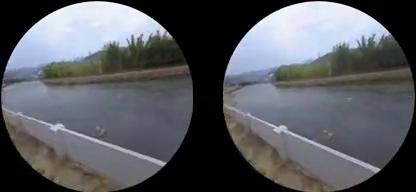}{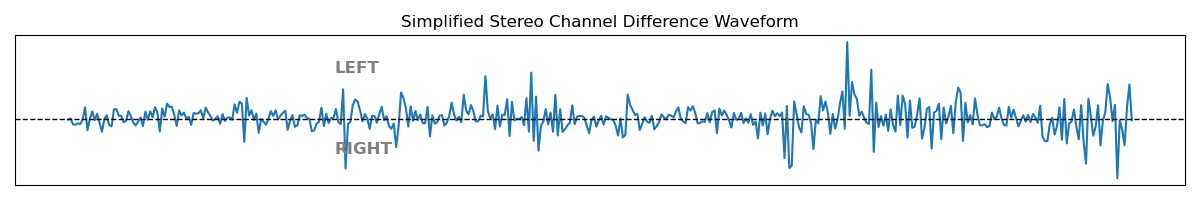}{Agriculture \& Rural}

% \label{Agriculture \& Rural}

\verticalphotos{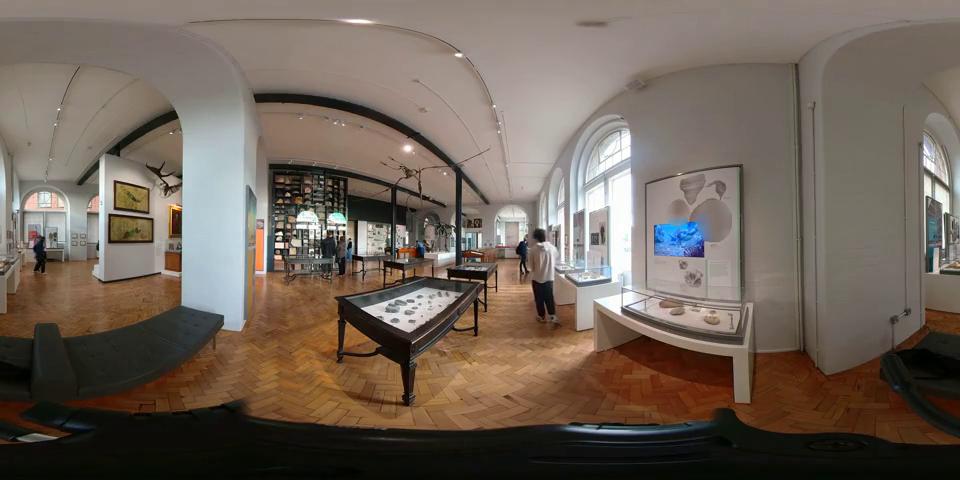}{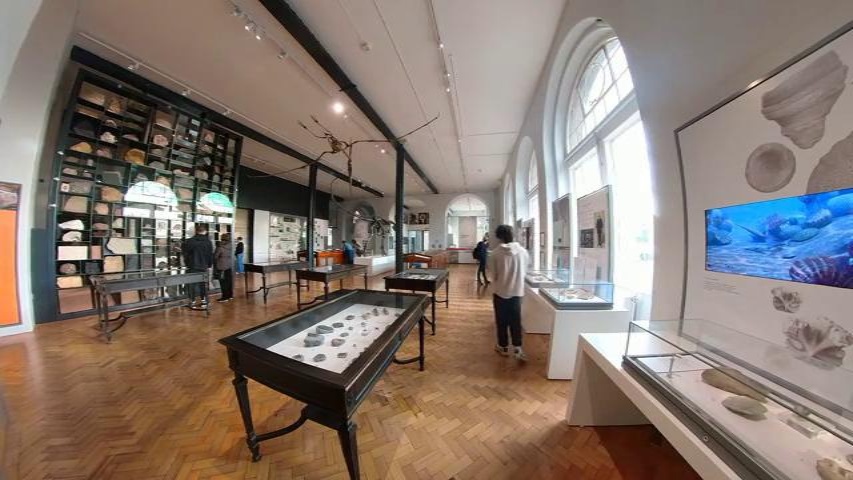}{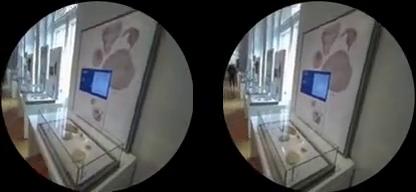}{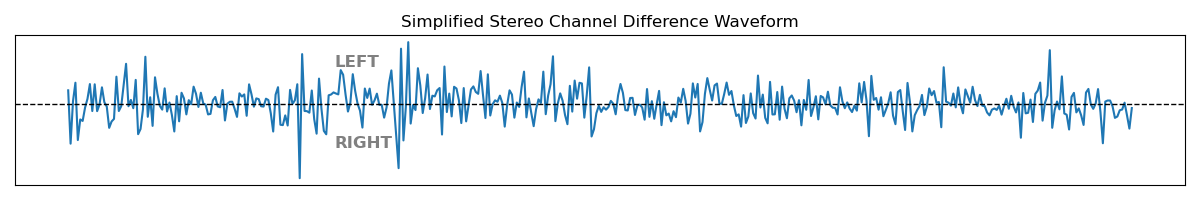}{Artistic Spaces}

\verticalphotos{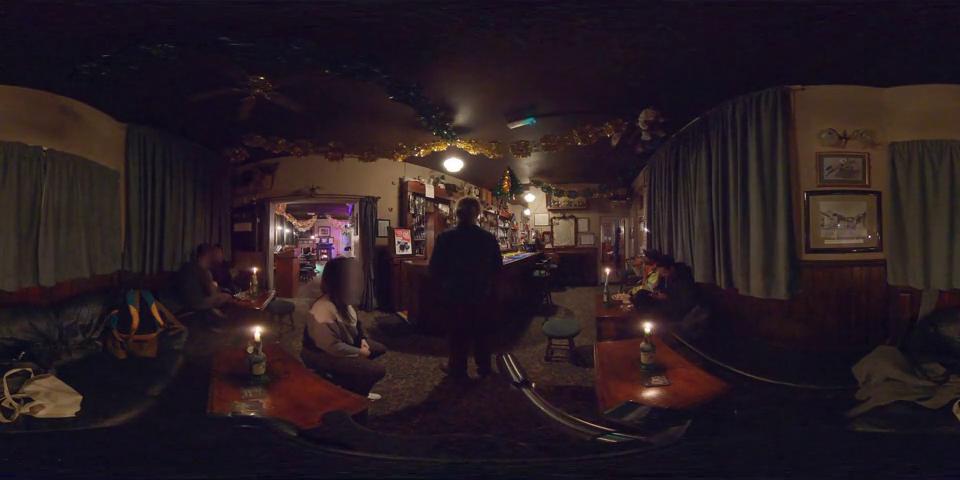}{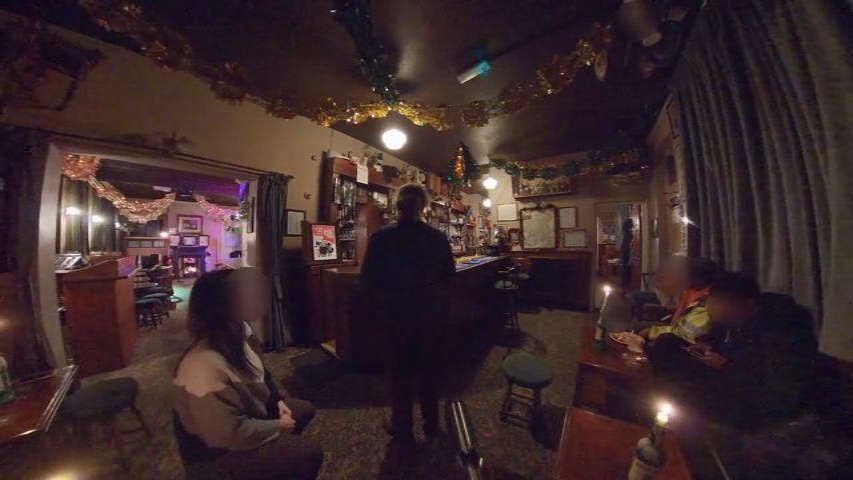}{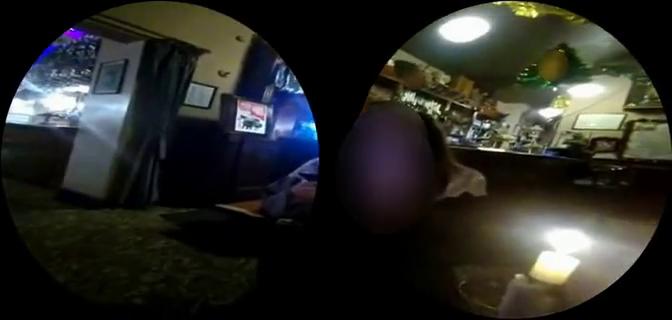}{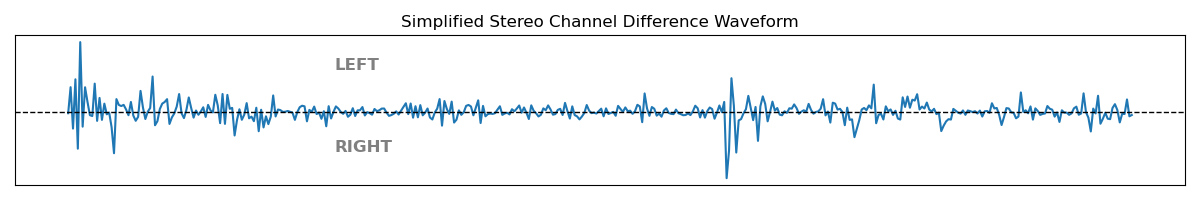}{Bars \& Nightlife}

\verticalphotos{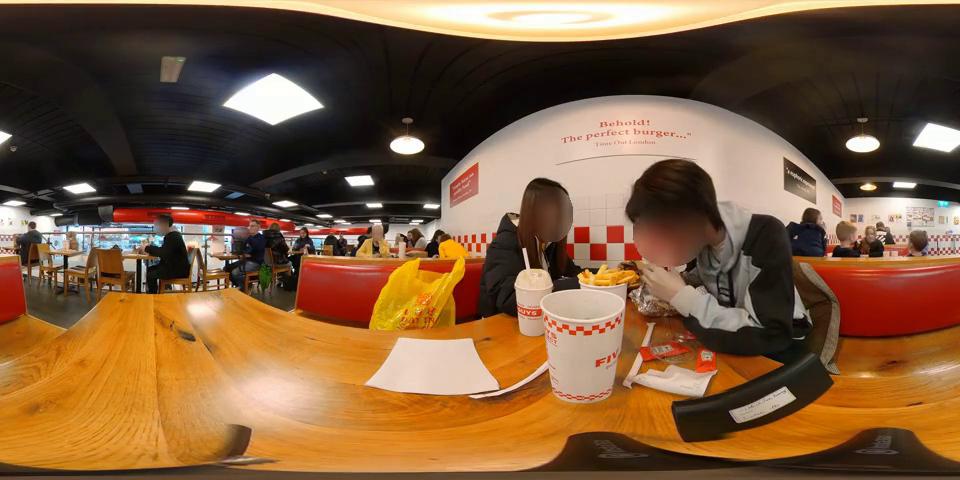}{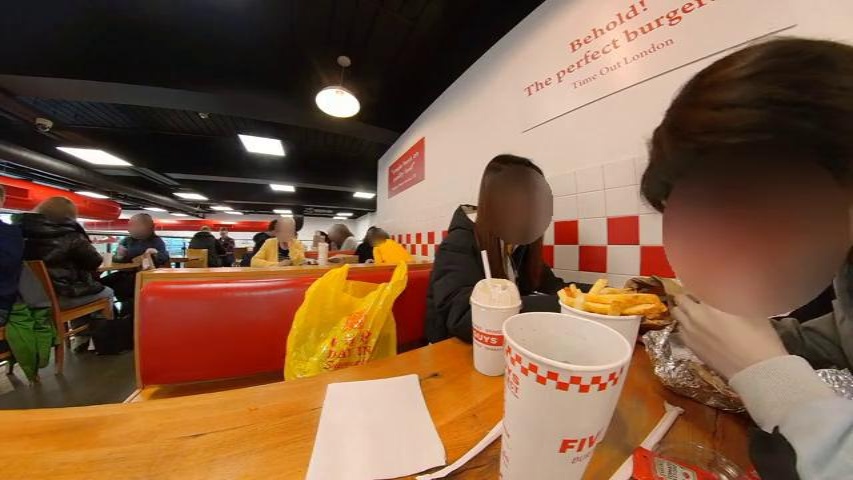}{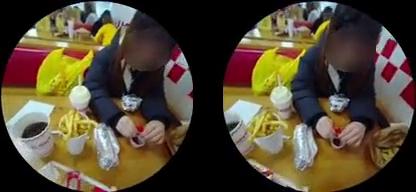}{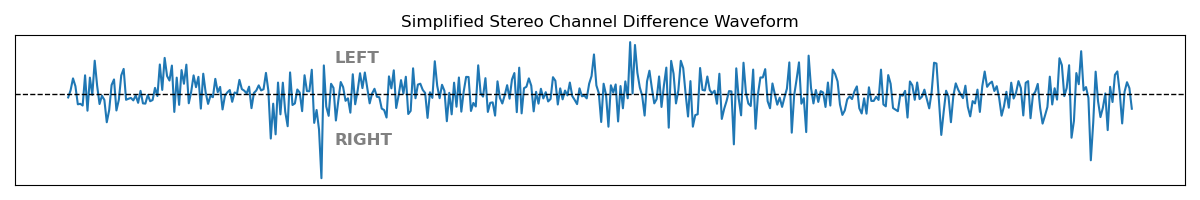}{Dining \& Food Outlets}

\verticalphotos{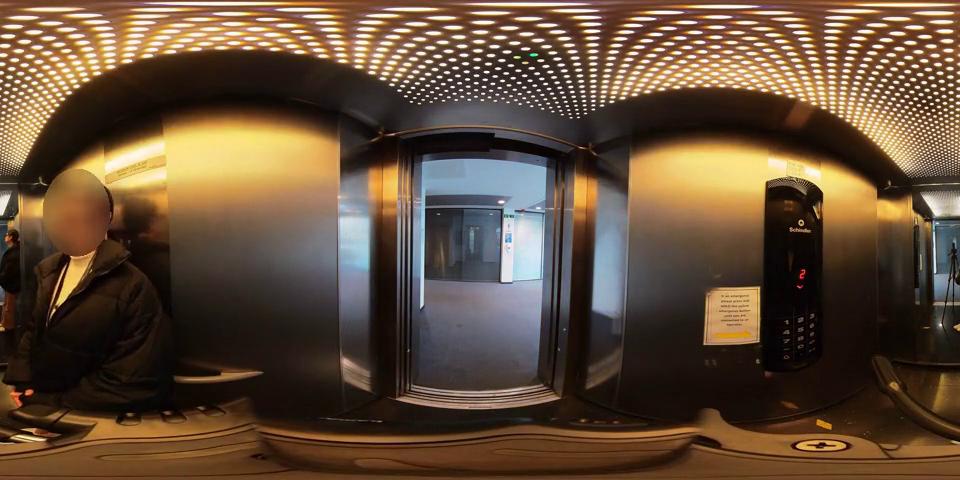}{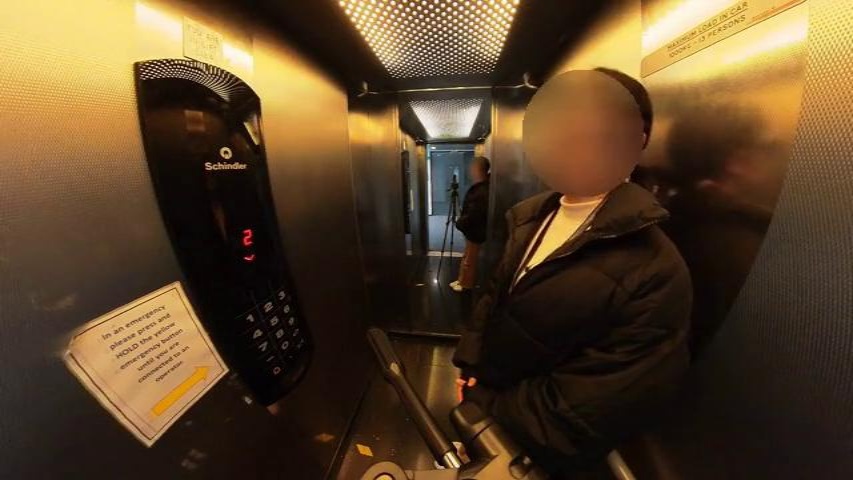}{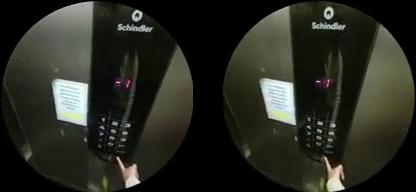}{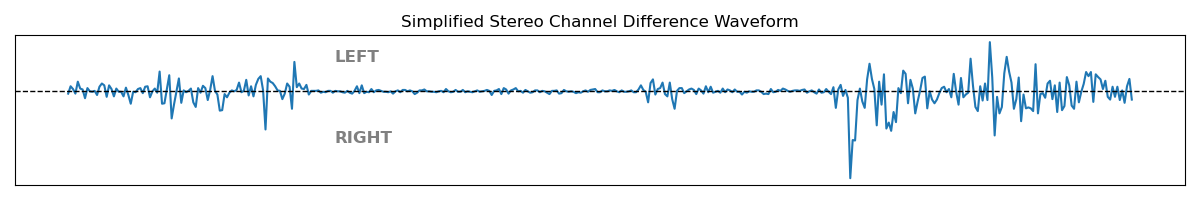}{Elevators \& Escalators}

\verticalphotos{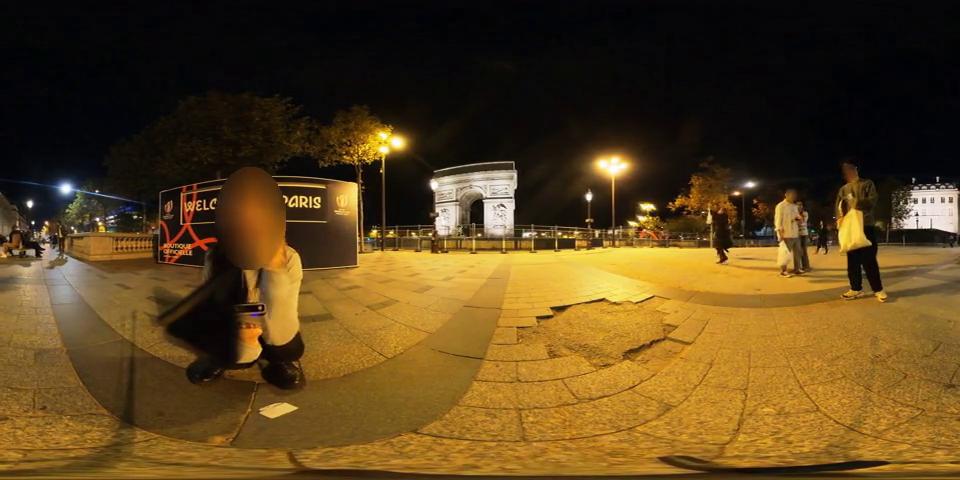}{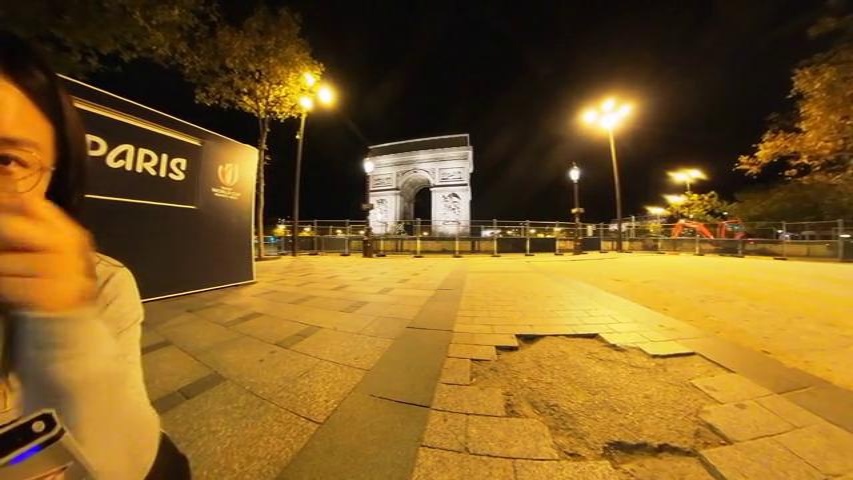}{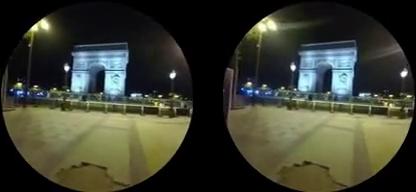}{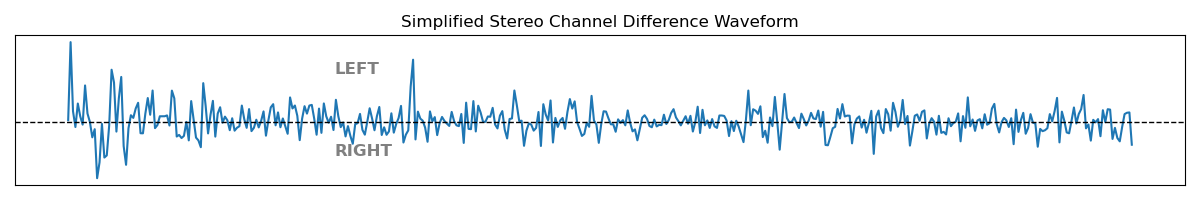}{Historic \& Religious Sites}

\verticalphotos{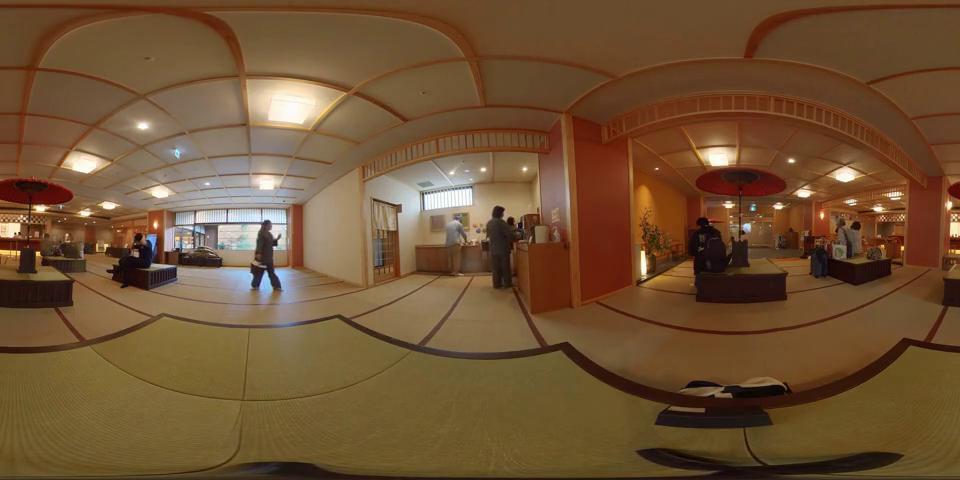}{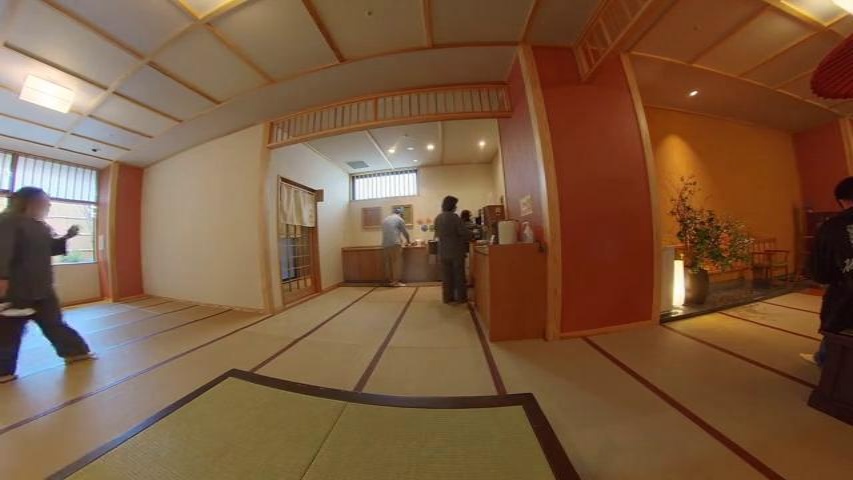}{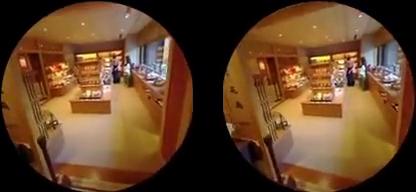}{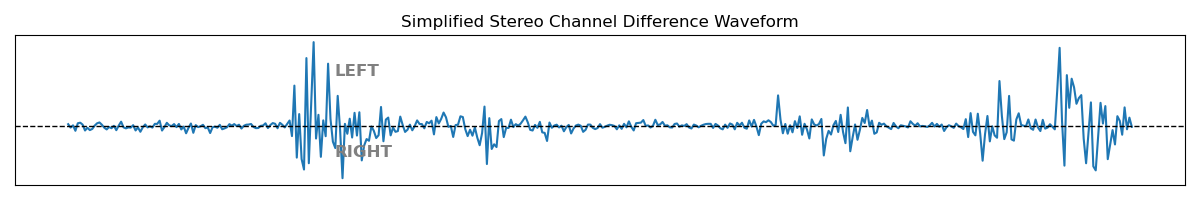}{Hotel \& Temporary Stay}

\verticalphotos{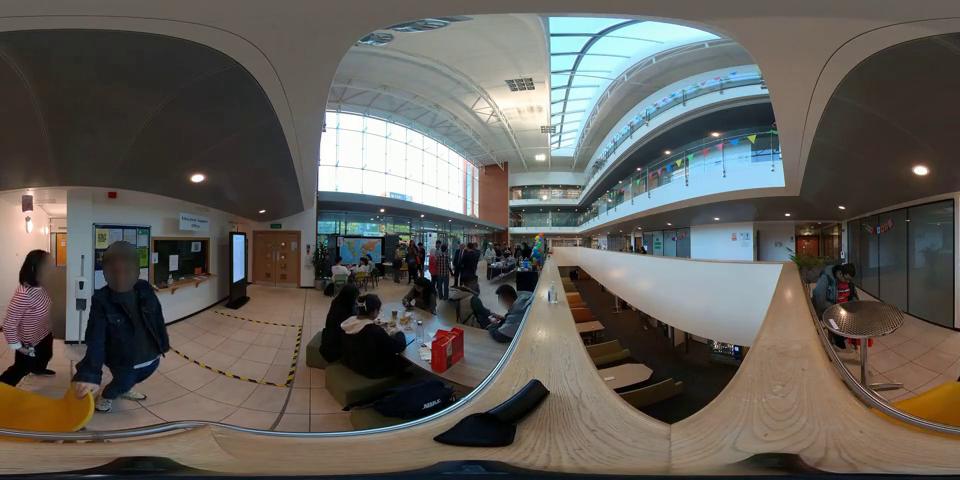}{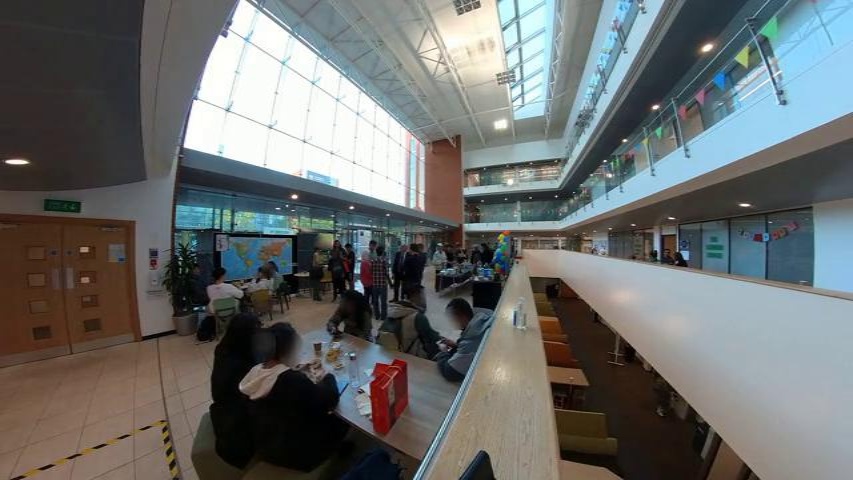}{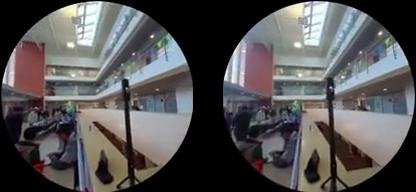}{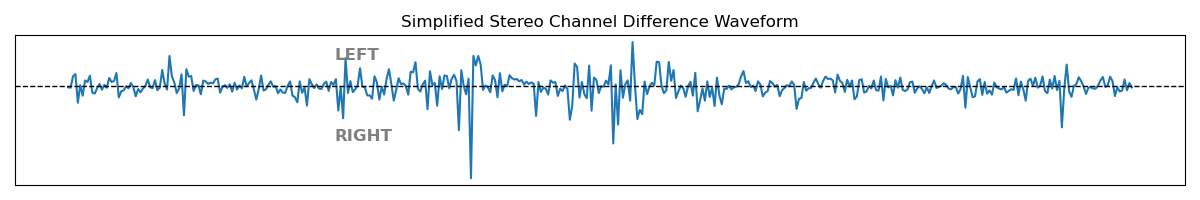}{Indoor Educational Spaces}

\verticalphotos{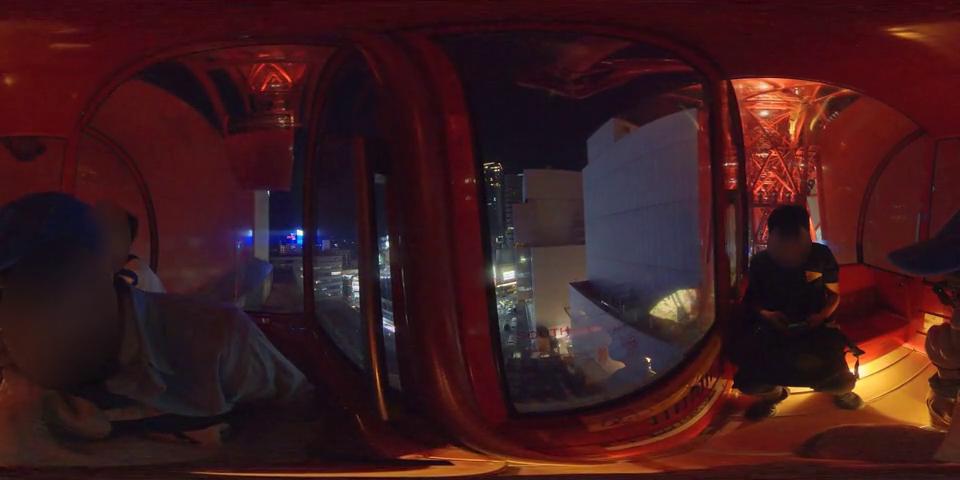}{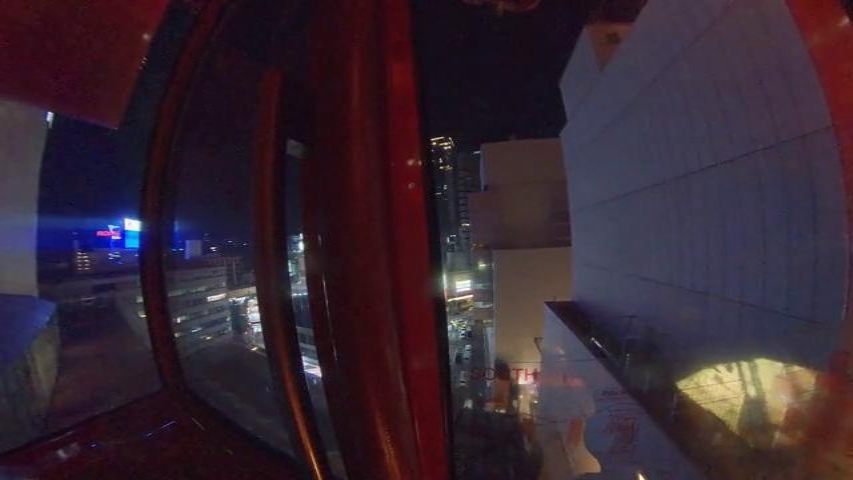}{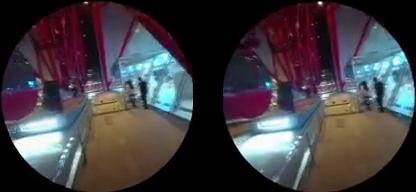}{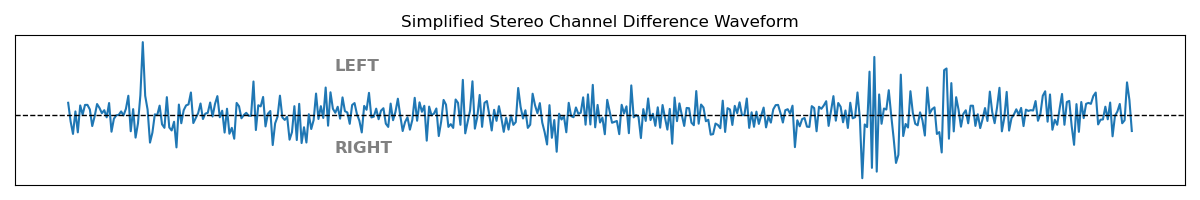}{Indoor Entertainment Venues}

\verticalphotos{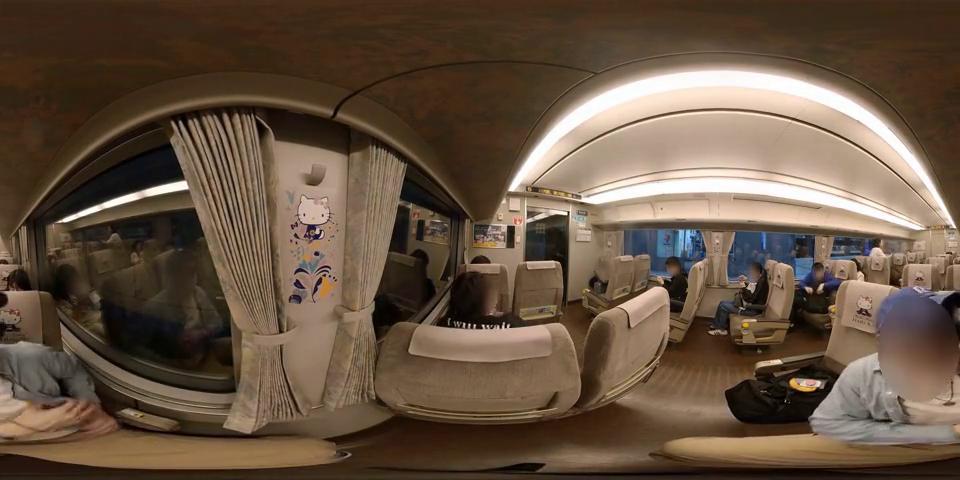}{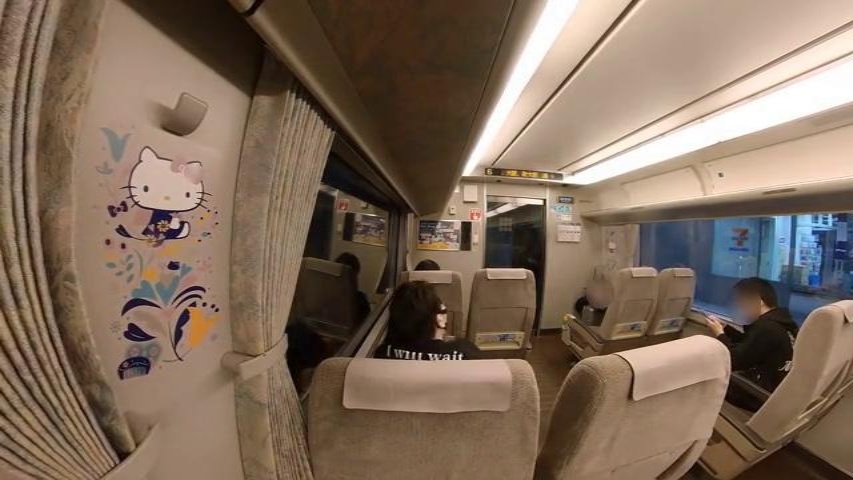}{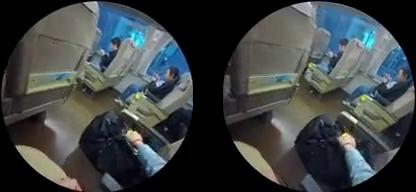}{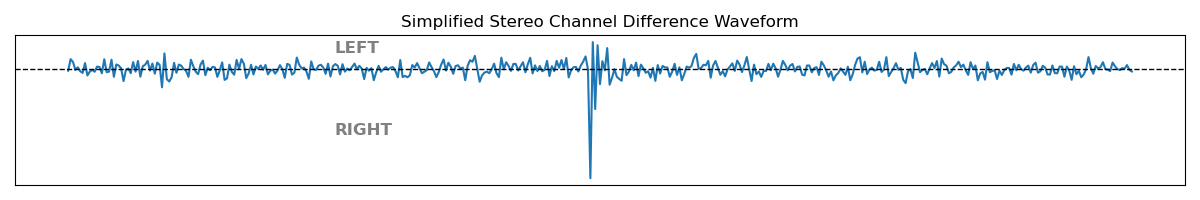}{Indoor Residential Spaces}

\verticalphotos{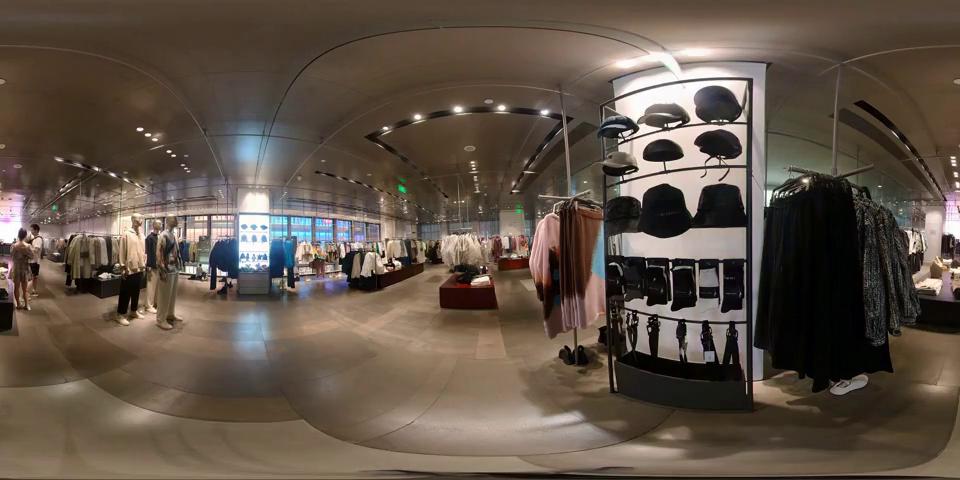}{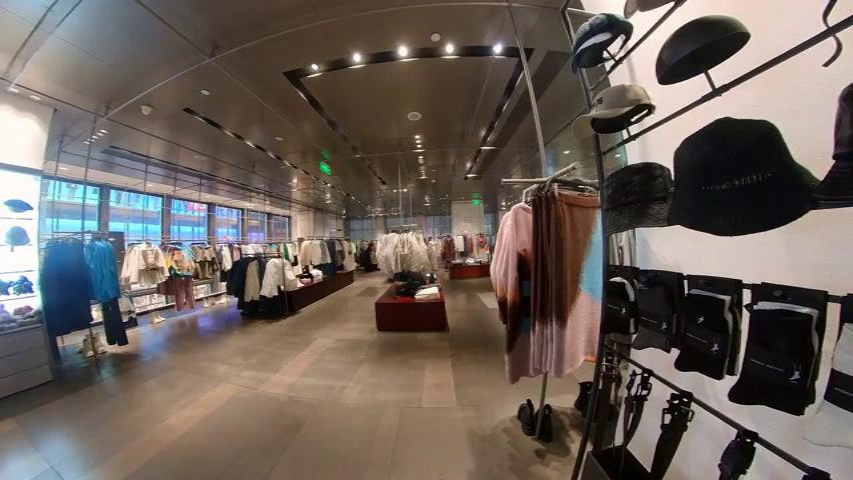}{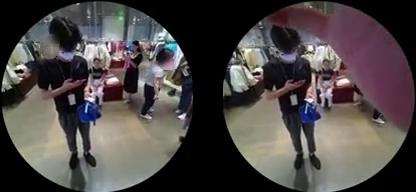}{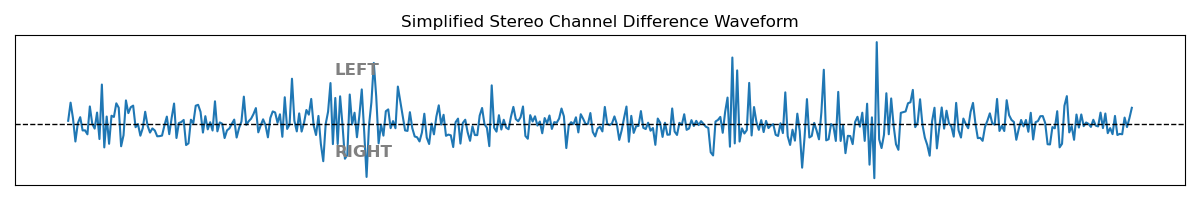}{Indoor Shops \& Retail \& Commercial}

\verticalphotos{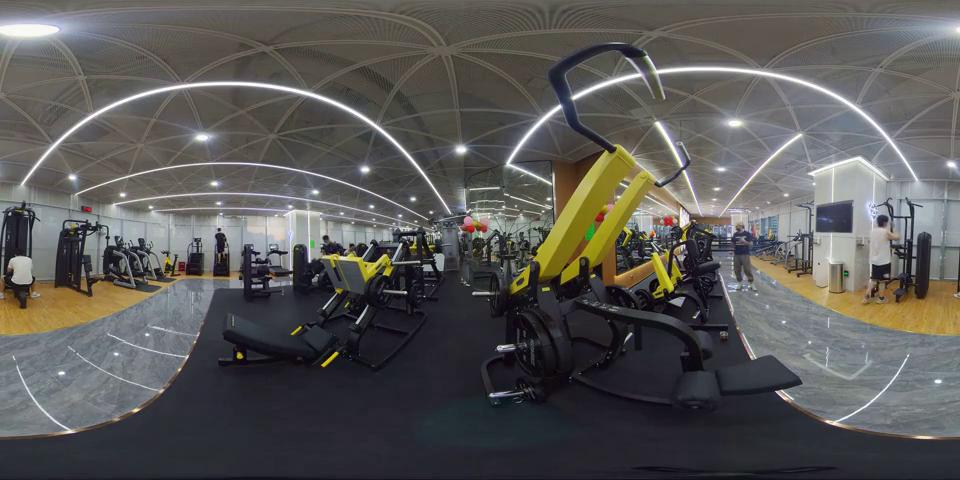}{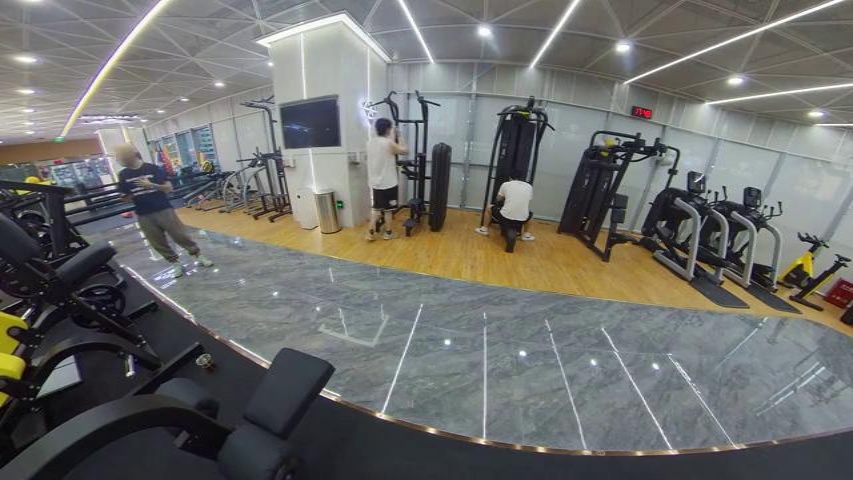}{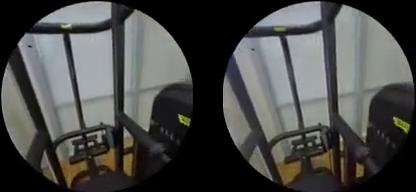}{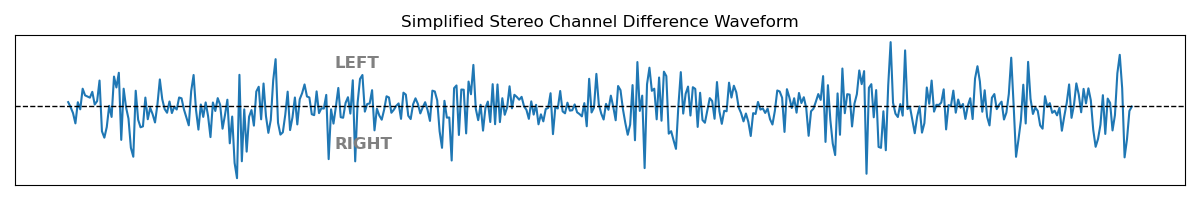}{Indoor Sports Venues}

\verticalphotos{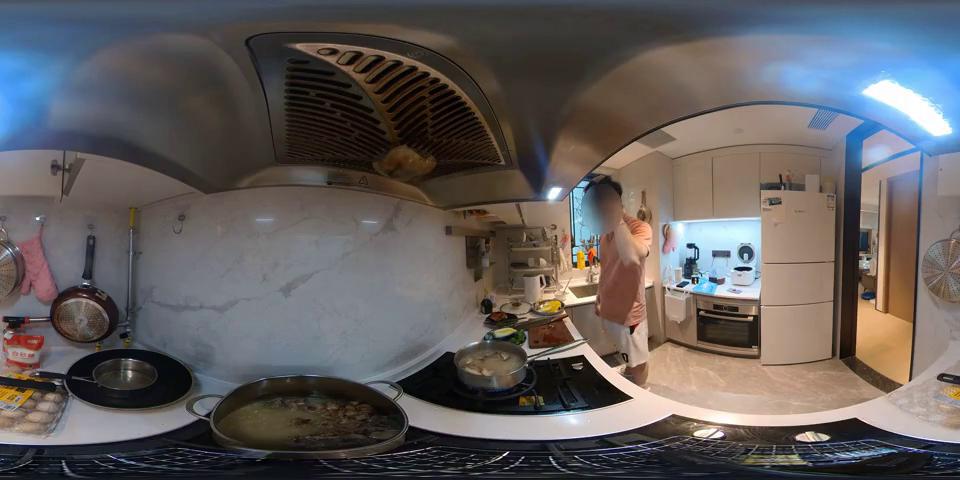}{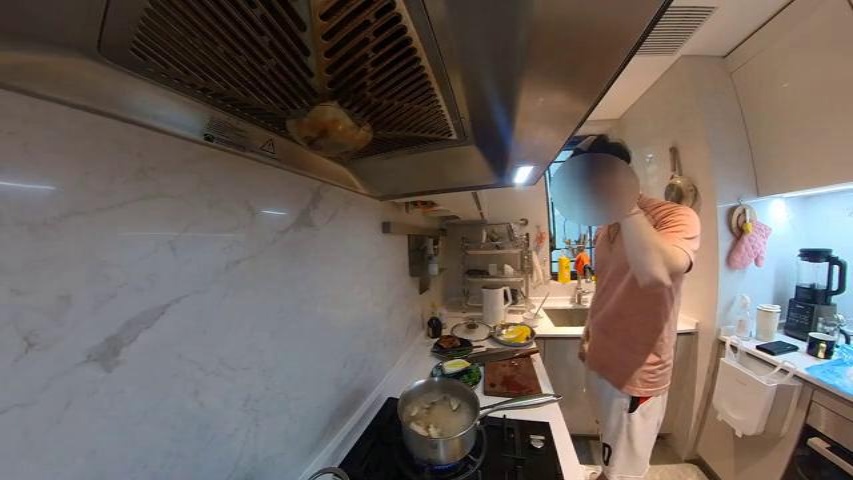}{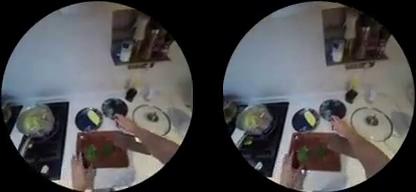}{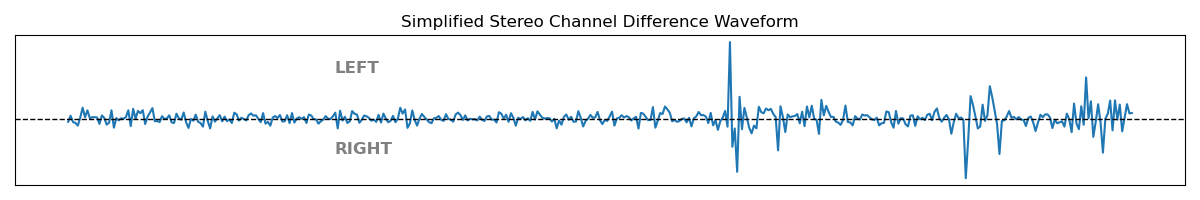}{Kitchen}

\verticalphotos{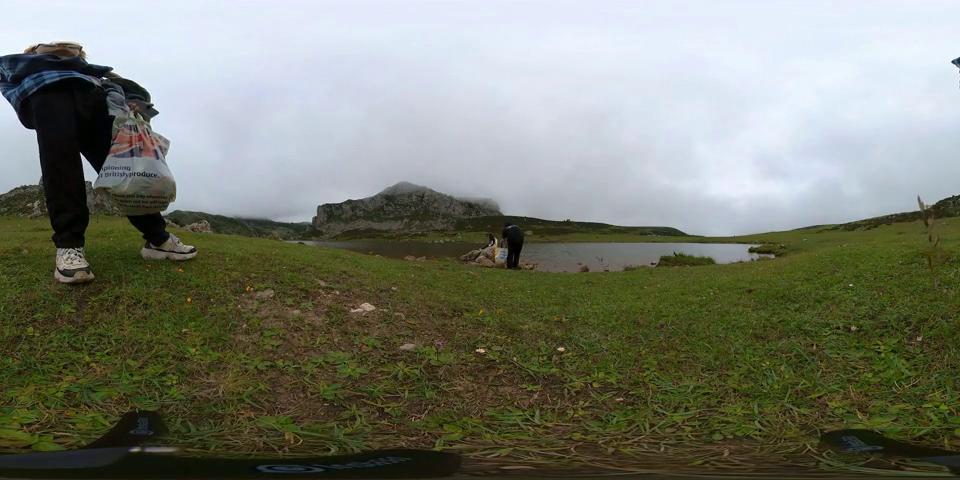}{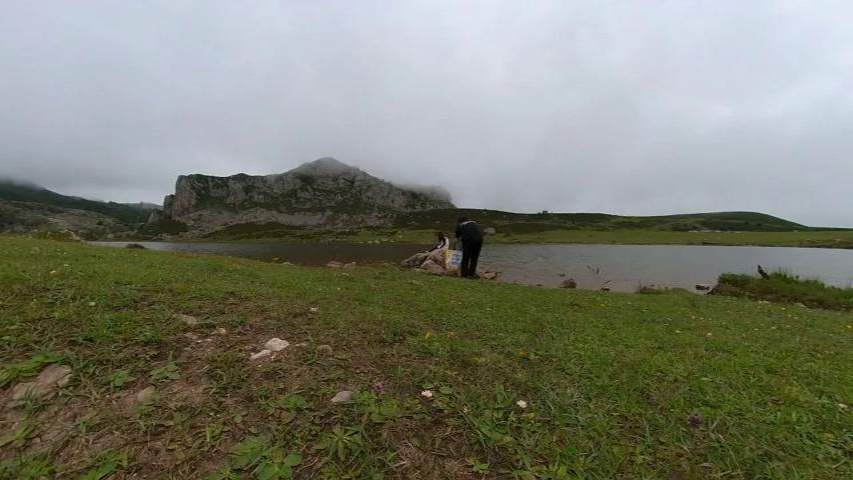}{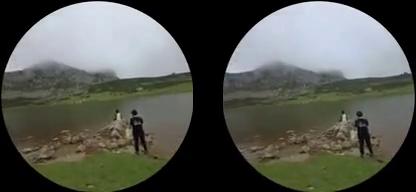}{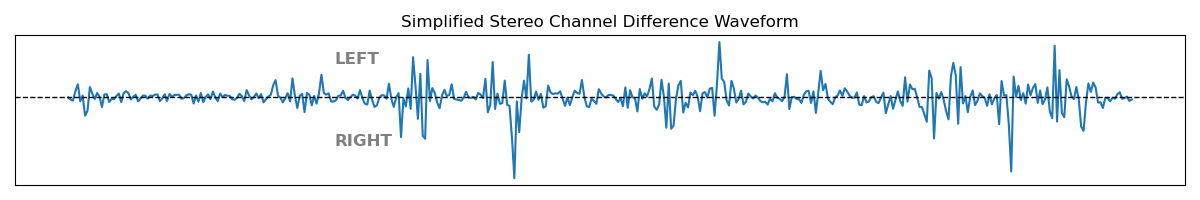}{Nature}

\verticalphotos{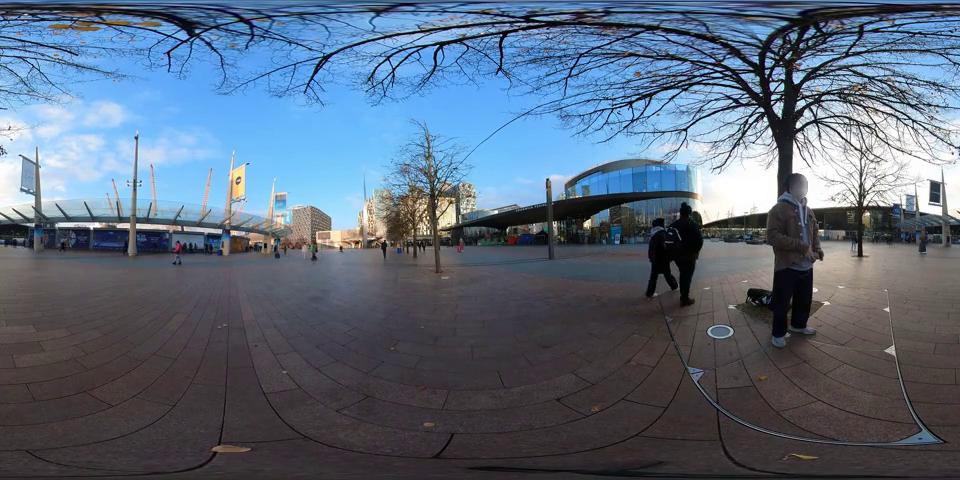}{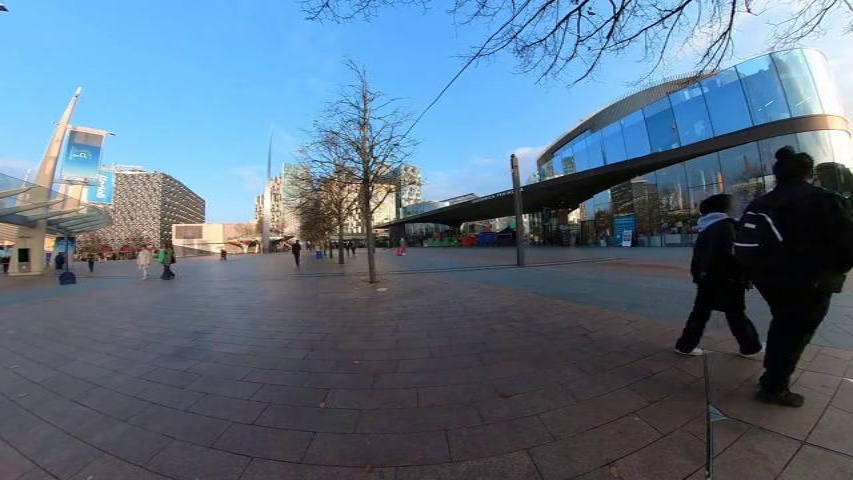}{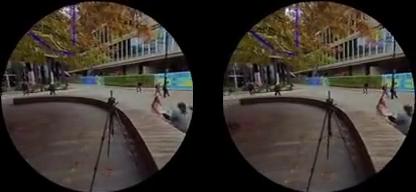}{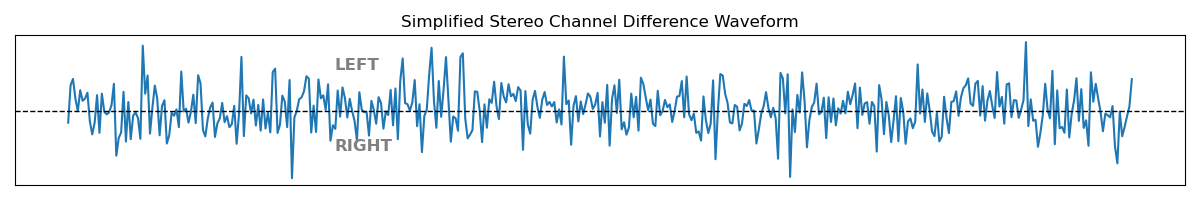}{Open Public Spaces}

\verticalphotos{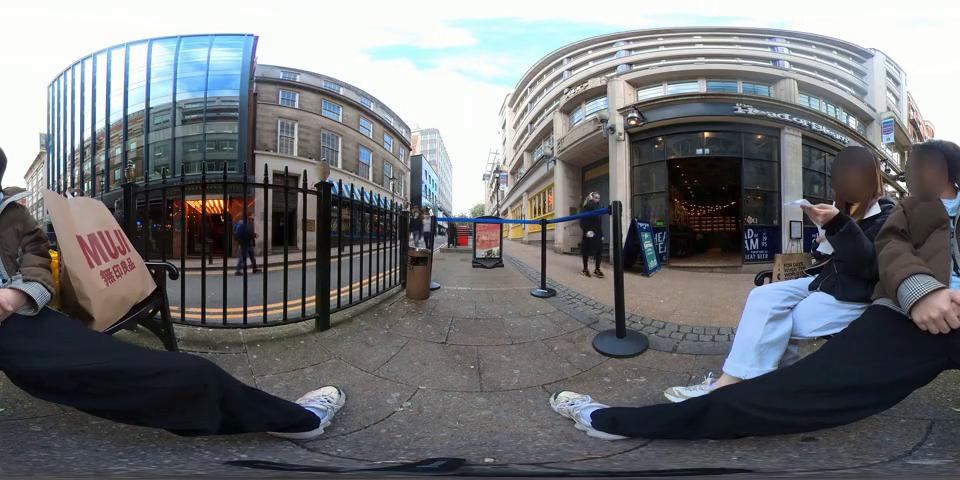}{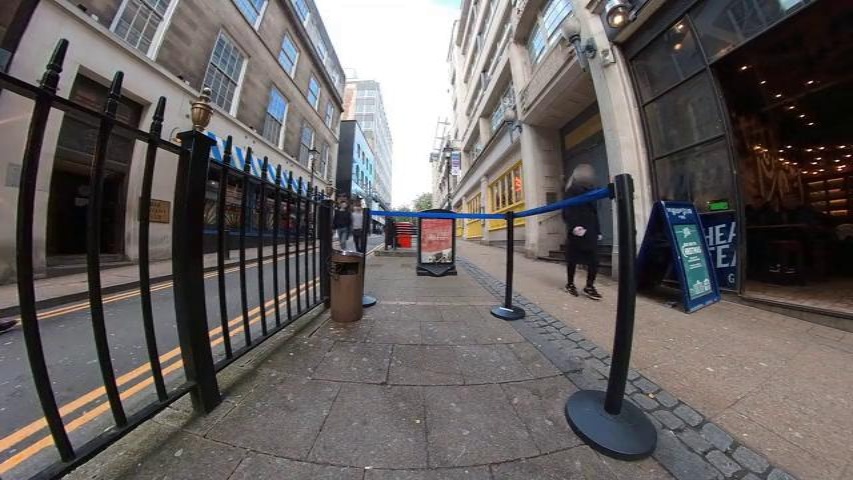}{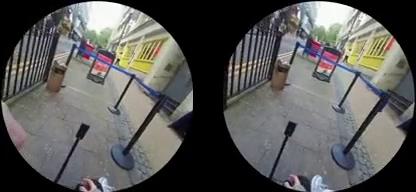}{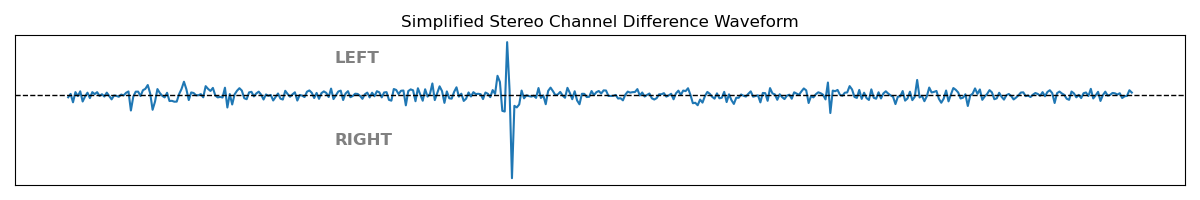}{Outdoor Commercial \& Markets Outside}

\verticalphotos{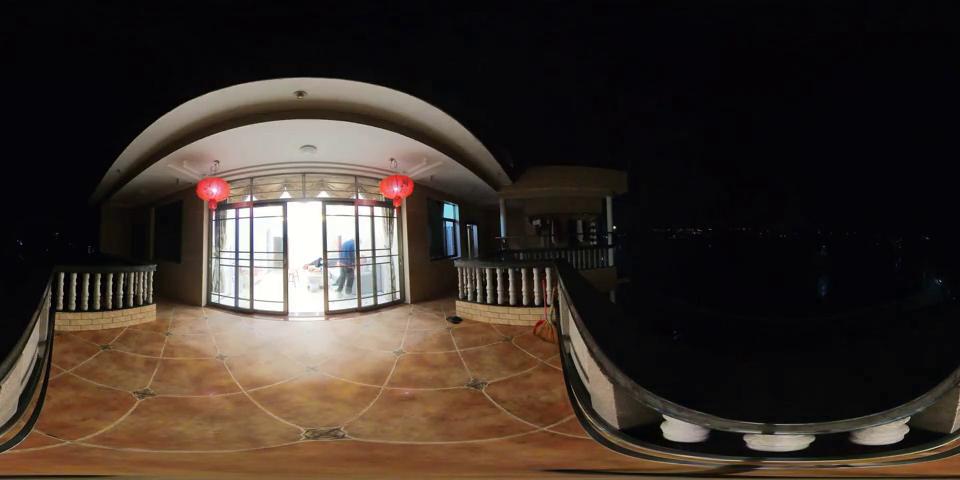}{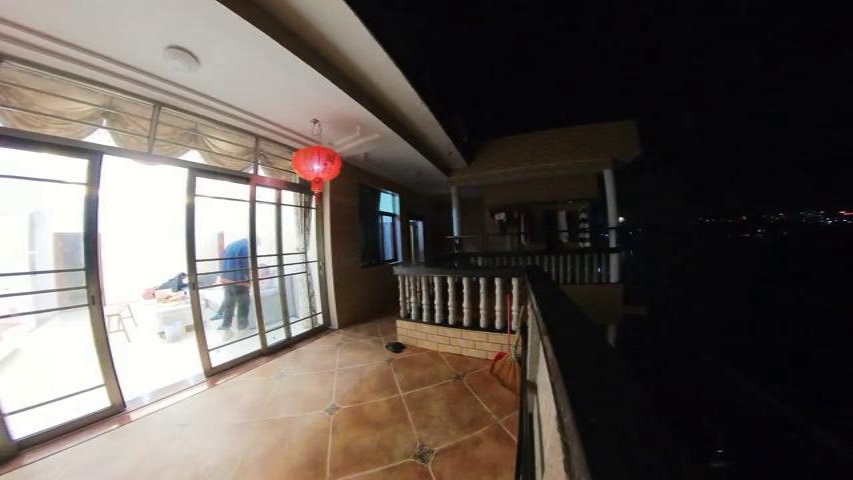}{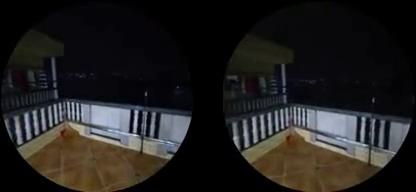}{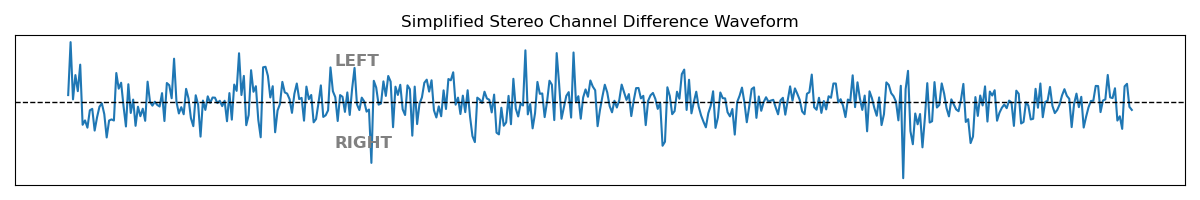}{Outdoor Residences \& Living}

\verticalphotos{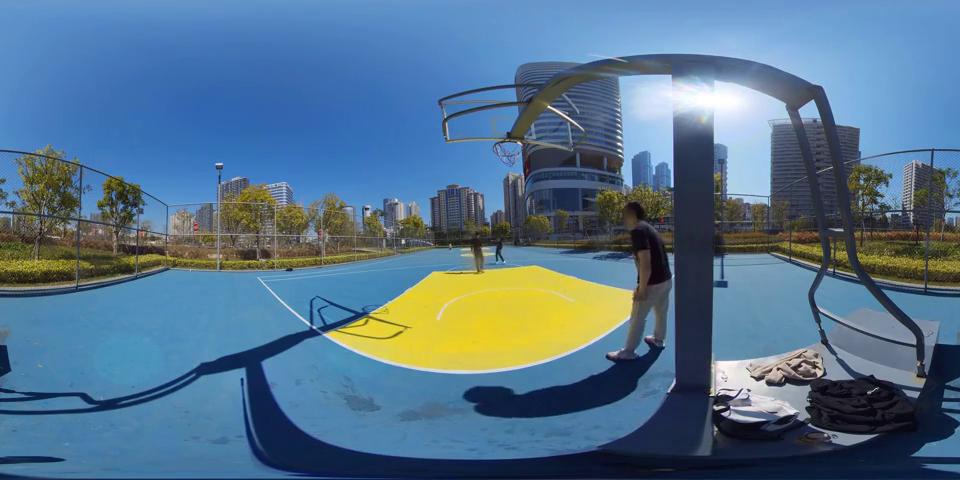}{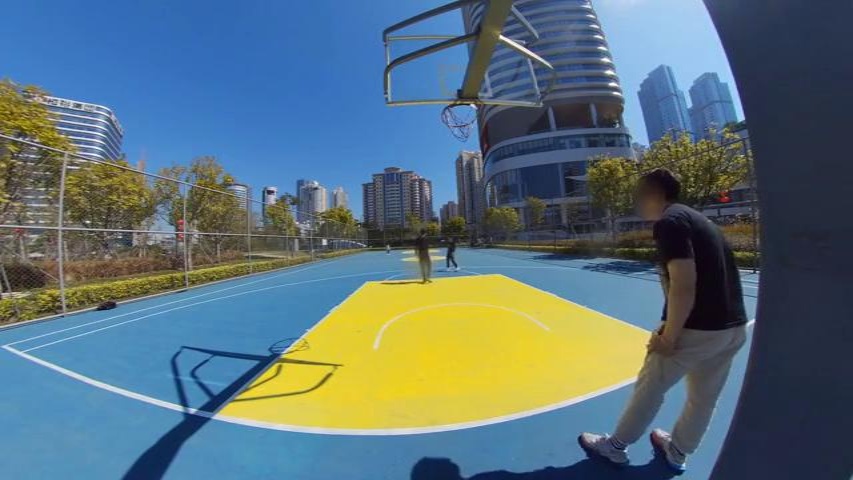}{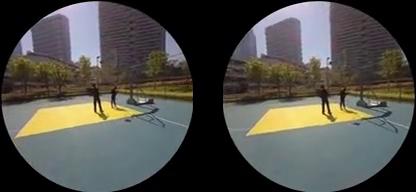}{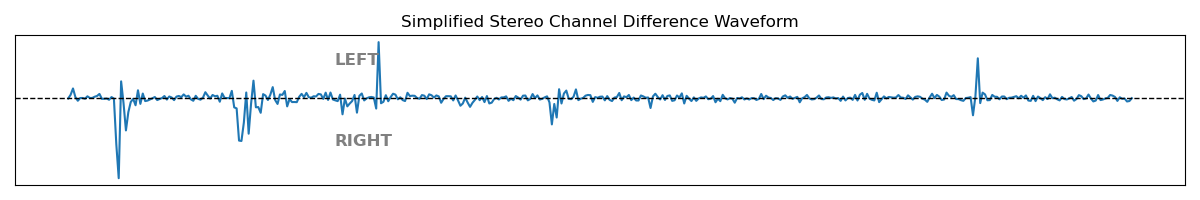}{Outdoor Sports \& Athletic Fields}

\verticalphotos{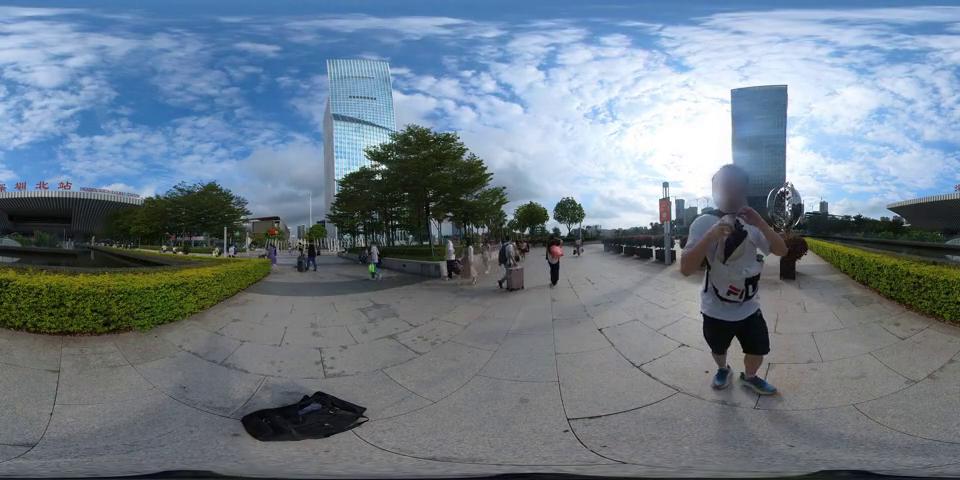}{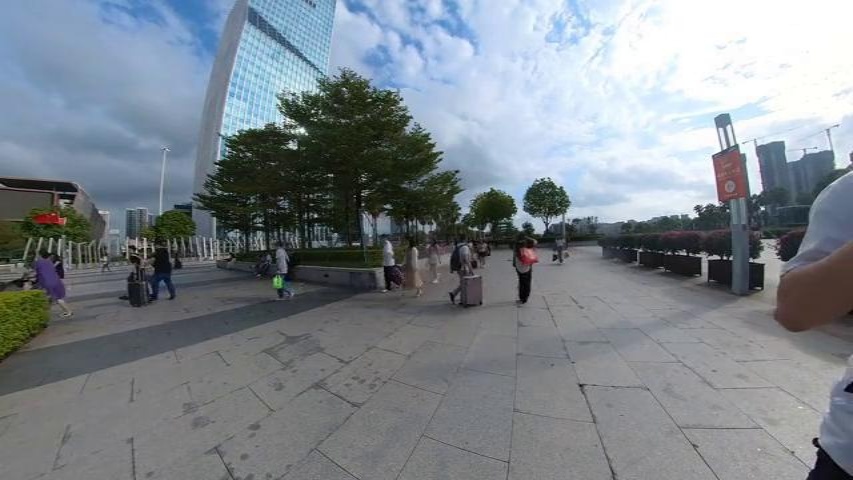}{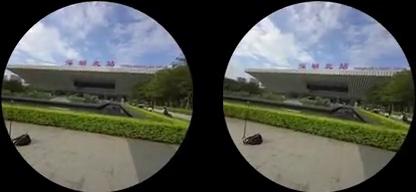}{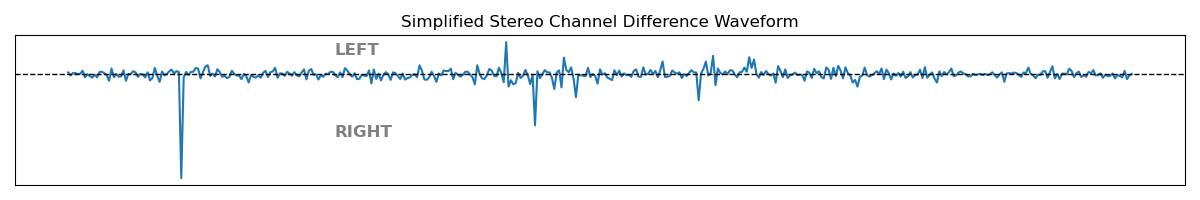}{Outdoor Transportation}

\verticalphotos{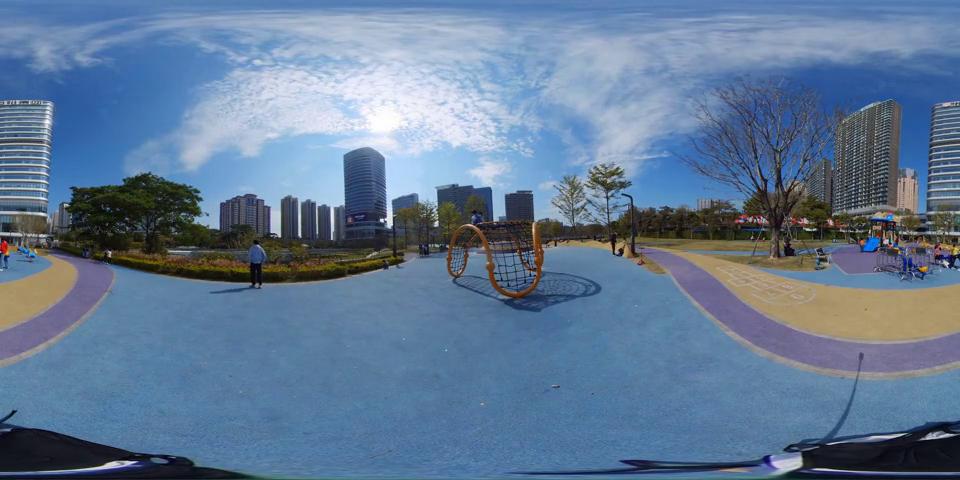}{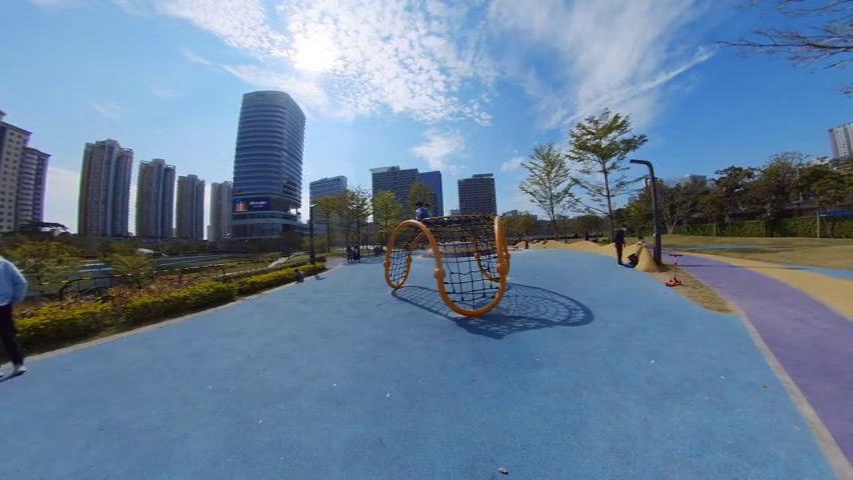}{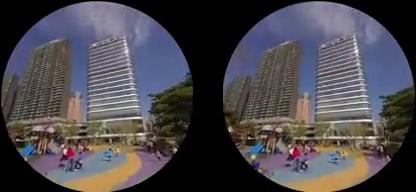}{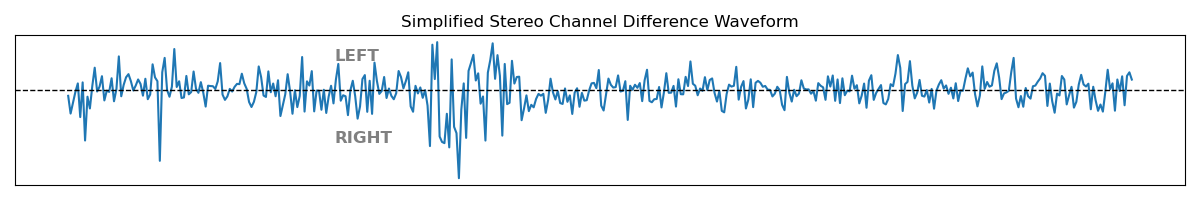}{Parks \& Recreational Areas}

\verticalphotos{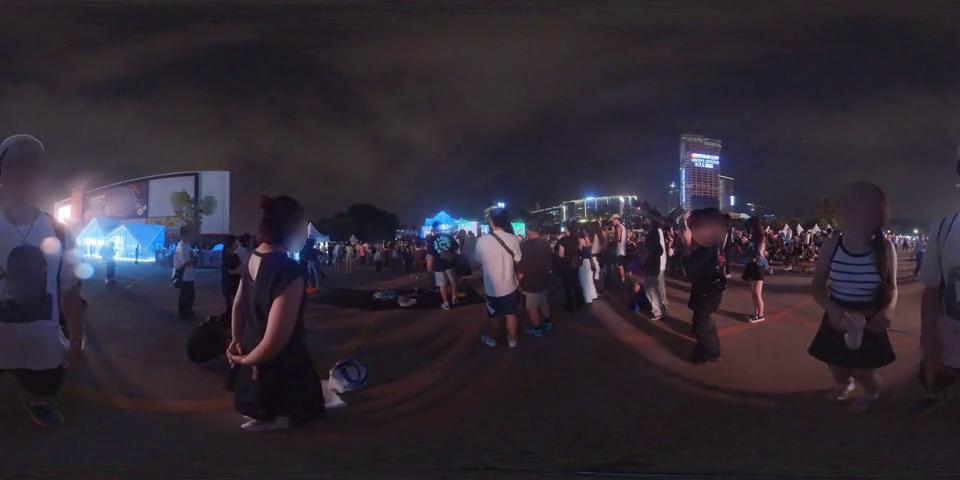}{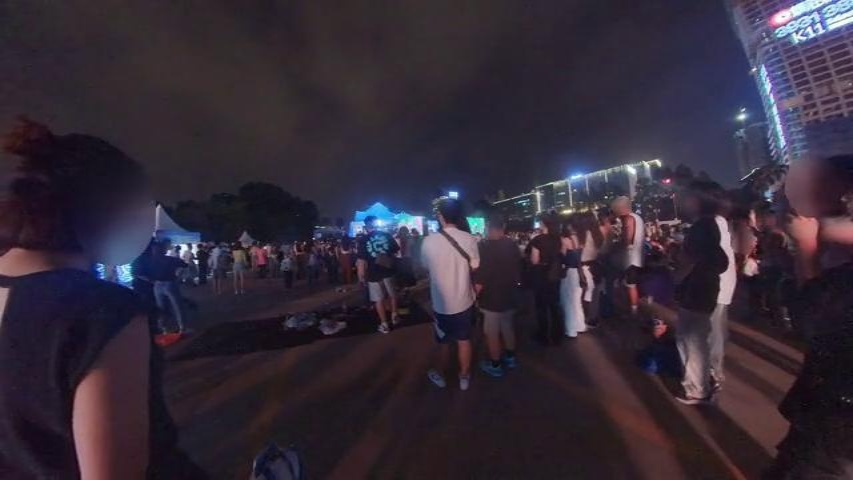}{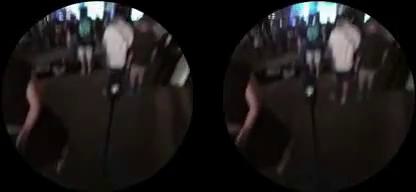}{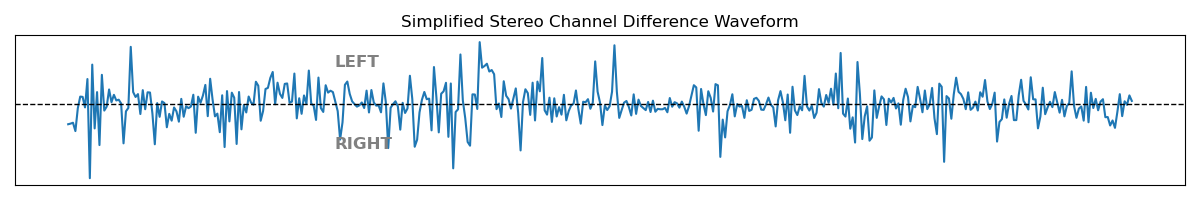}{Public Gathering \& Conference Spaces}

\verticalphotos{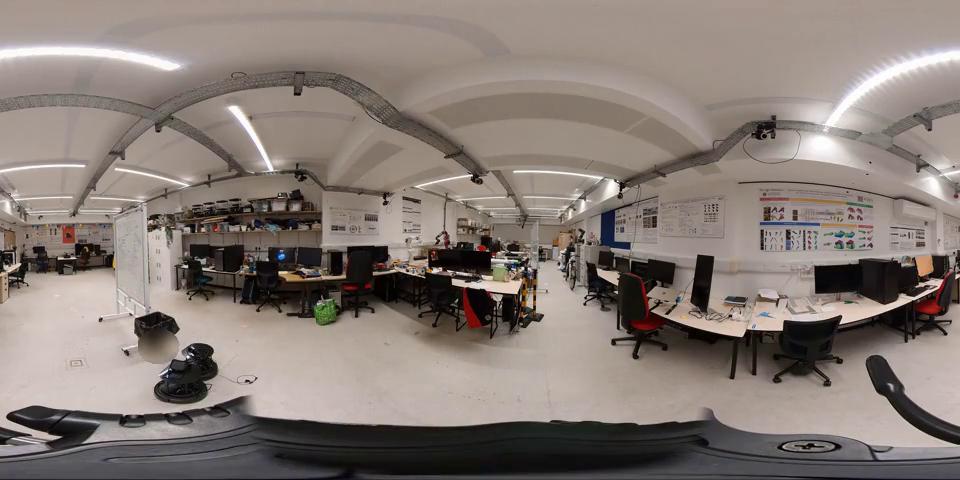}{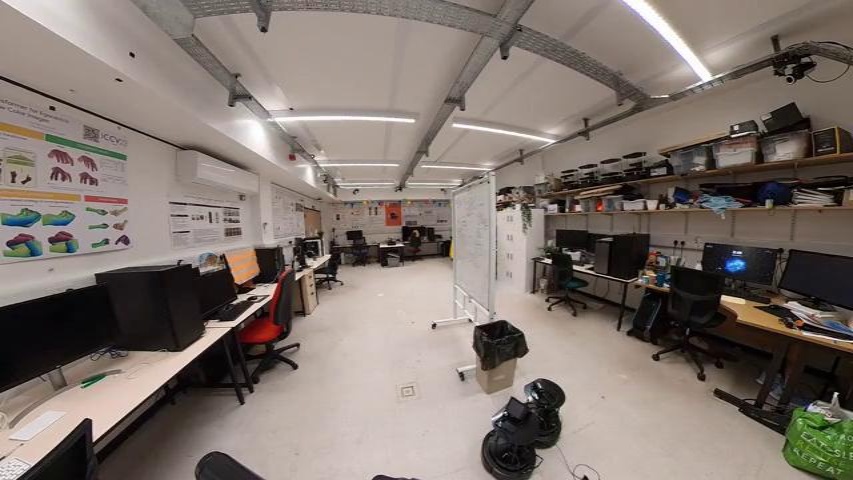}{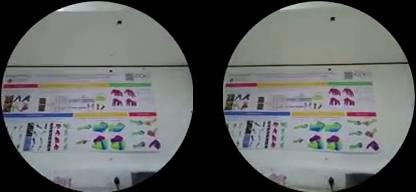}{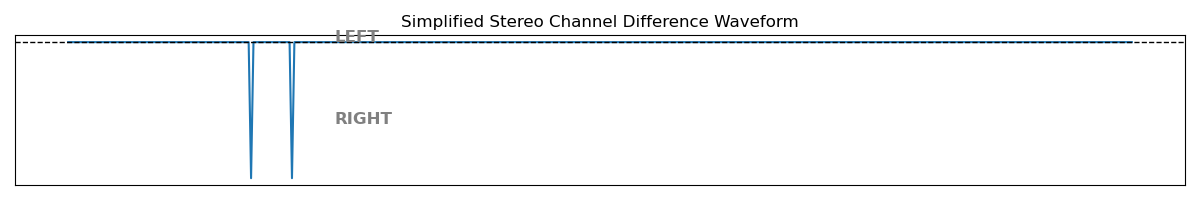}{Scientific Interior Space}

\verticalphotos{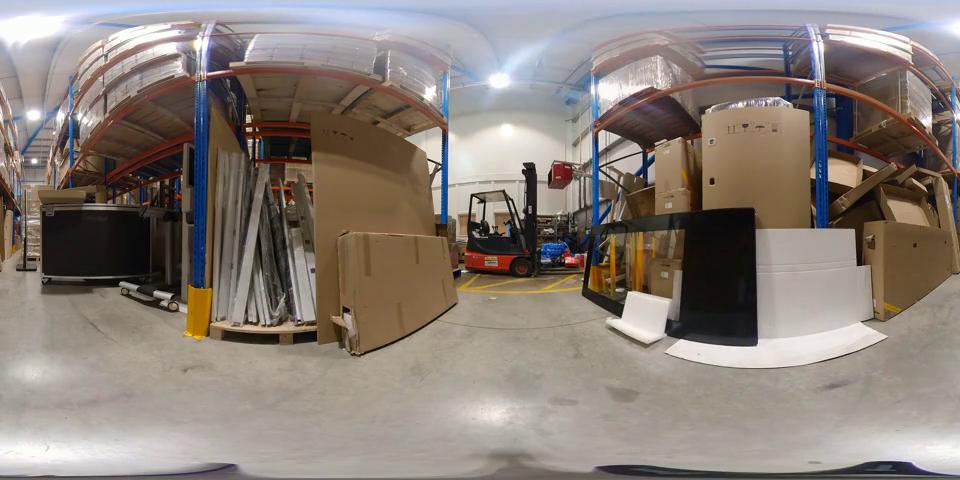}{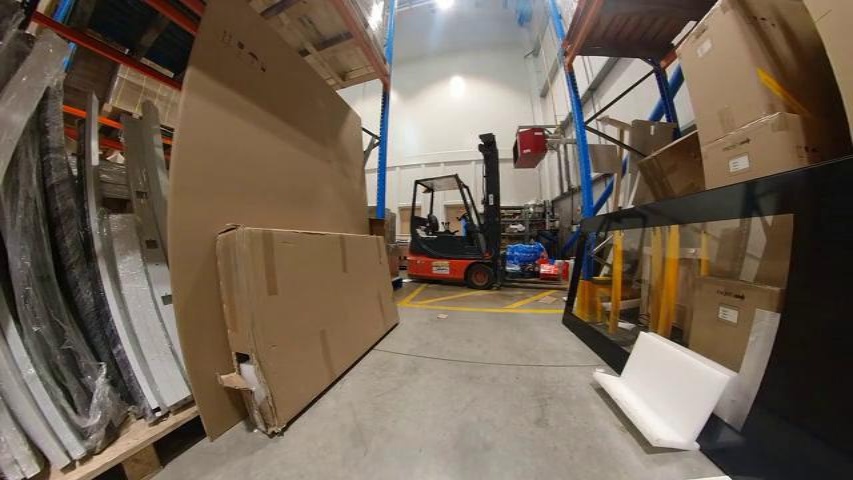}{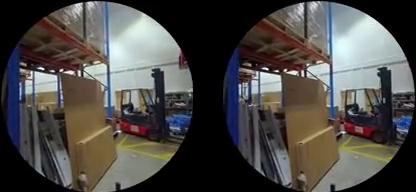}{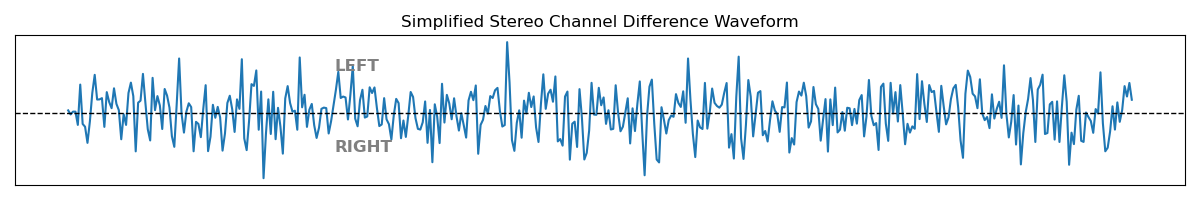}{Storage \& Utility}

\verticalphotos{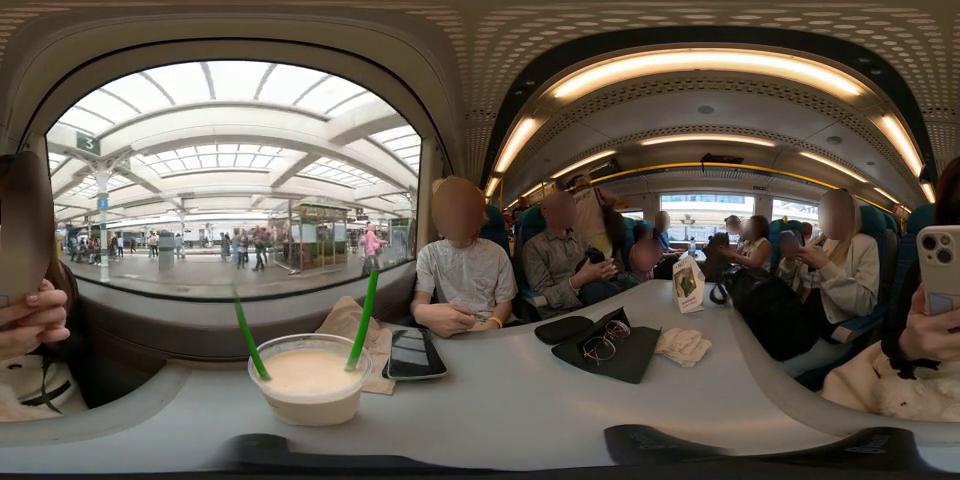}{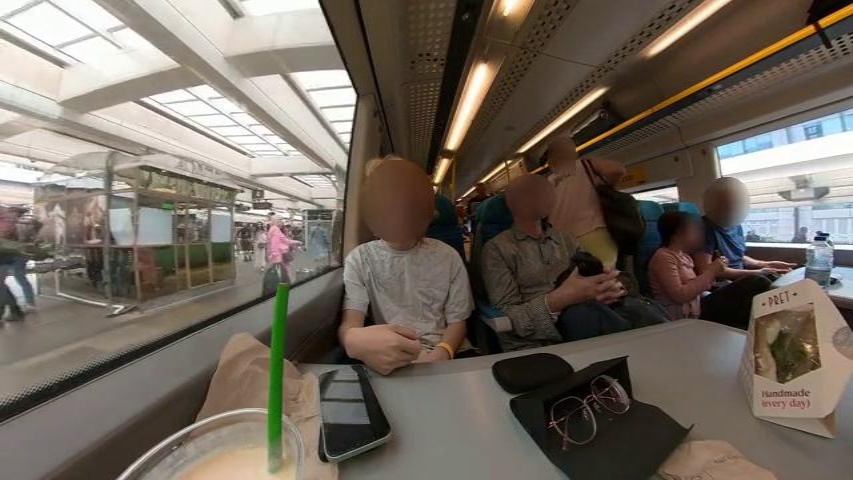}{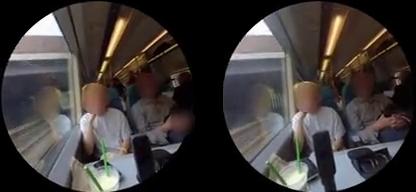}{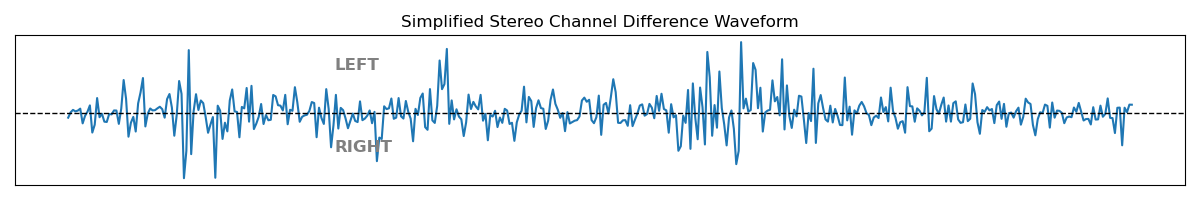}{Transportation Interiors}

\verticalphotos{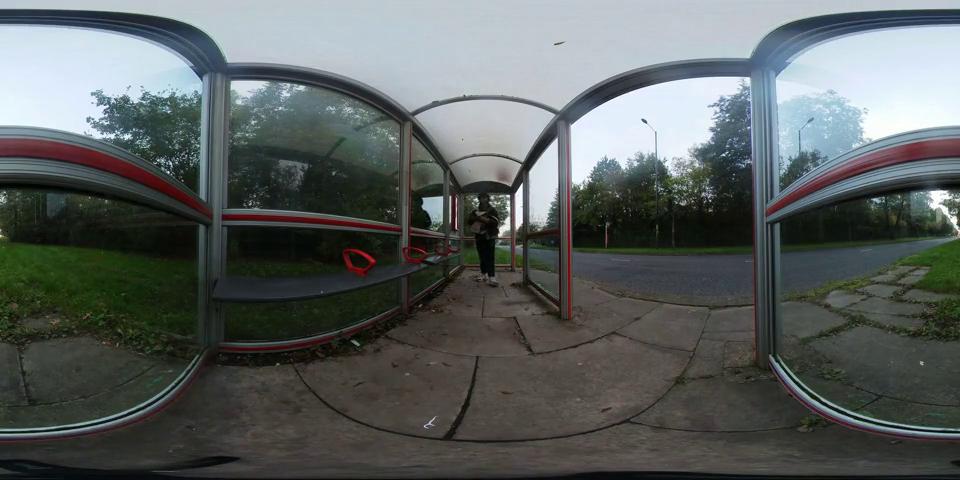}{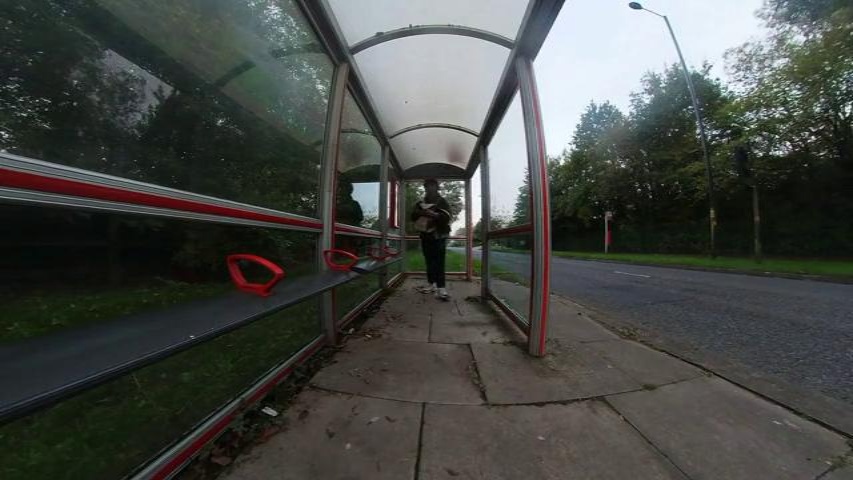}{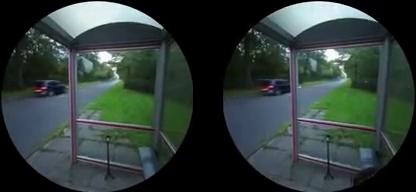}{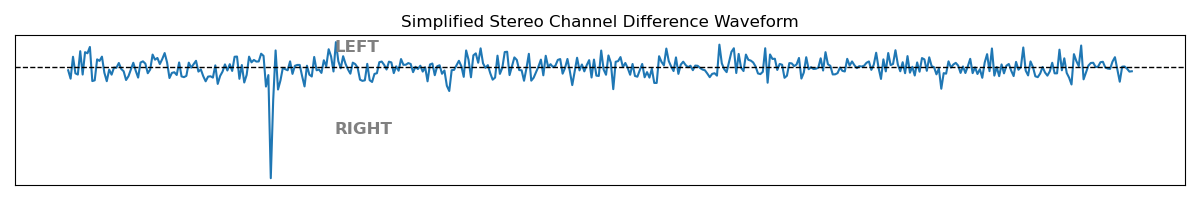}{Transportation Stops}

\verticalphotos{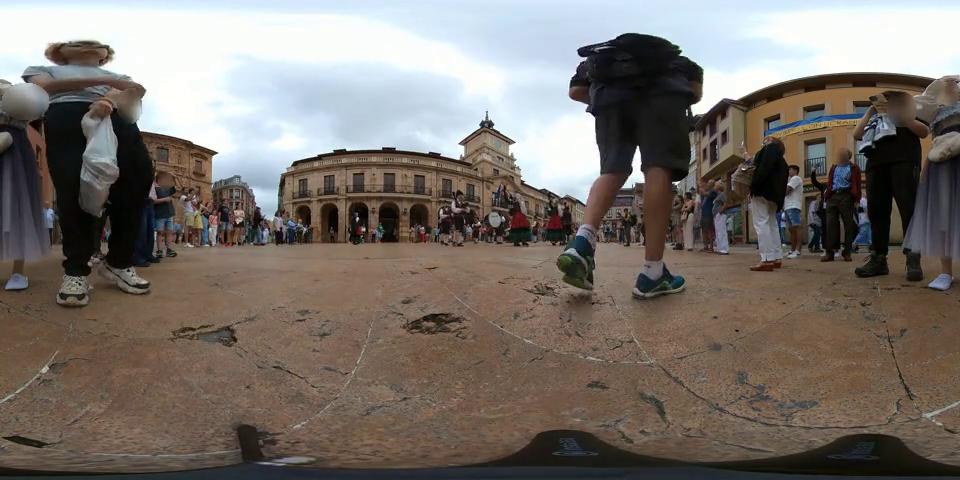}{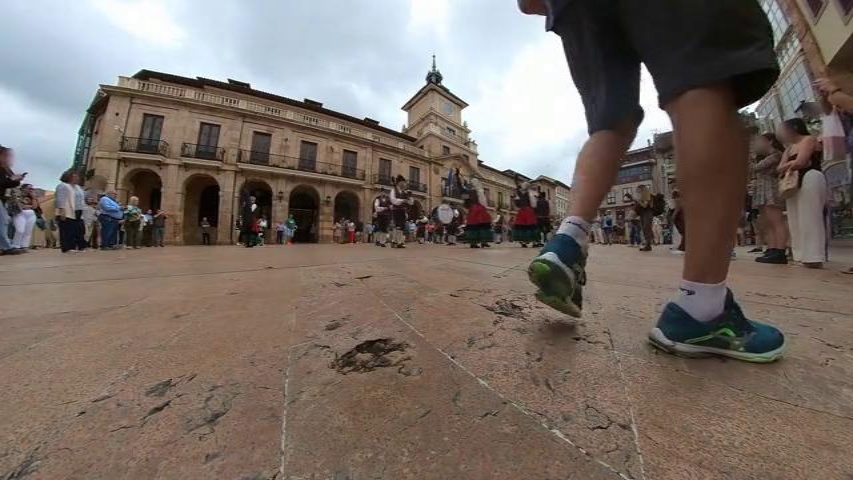}{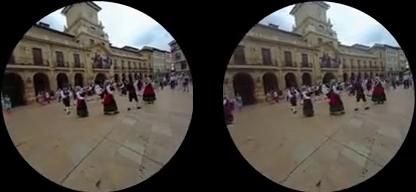}{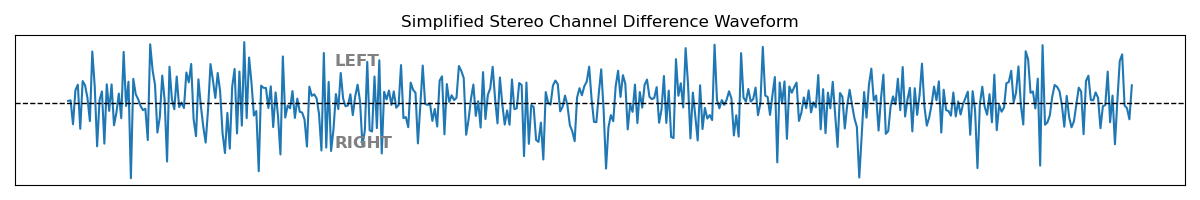}{Urban Constructions \& street}

\verticalphotos{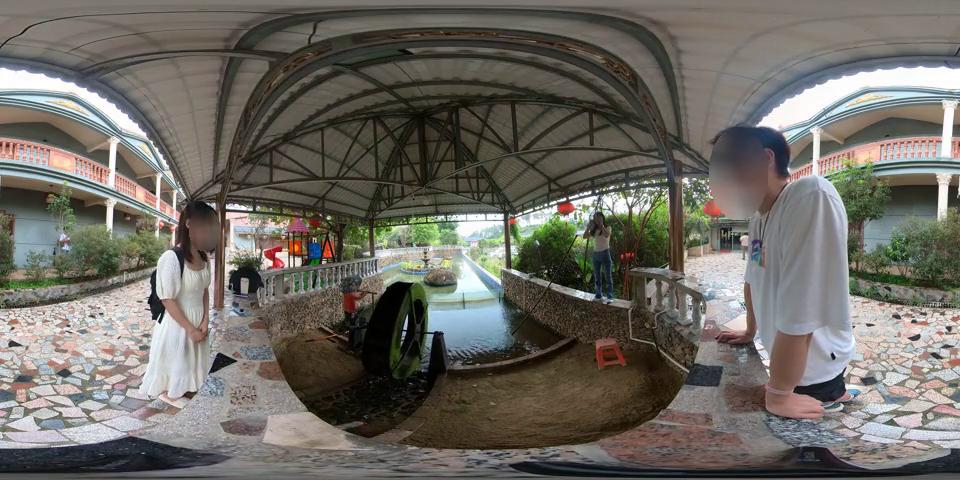}{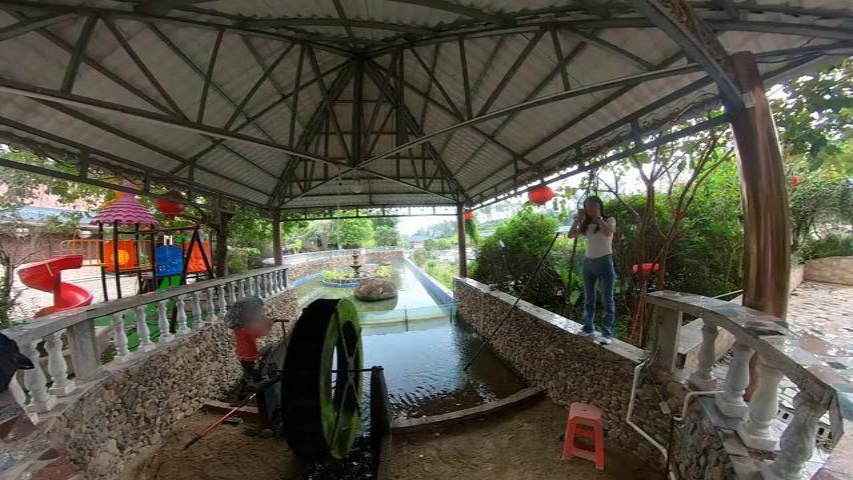}{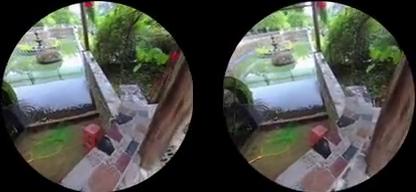}{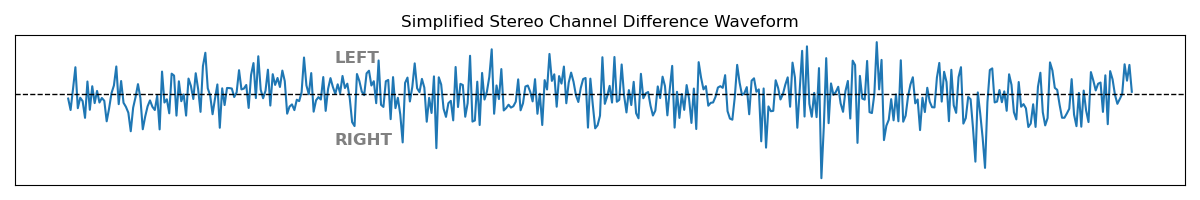}{Waterfronts \& Water Bodies}

\verticalphotosEnd{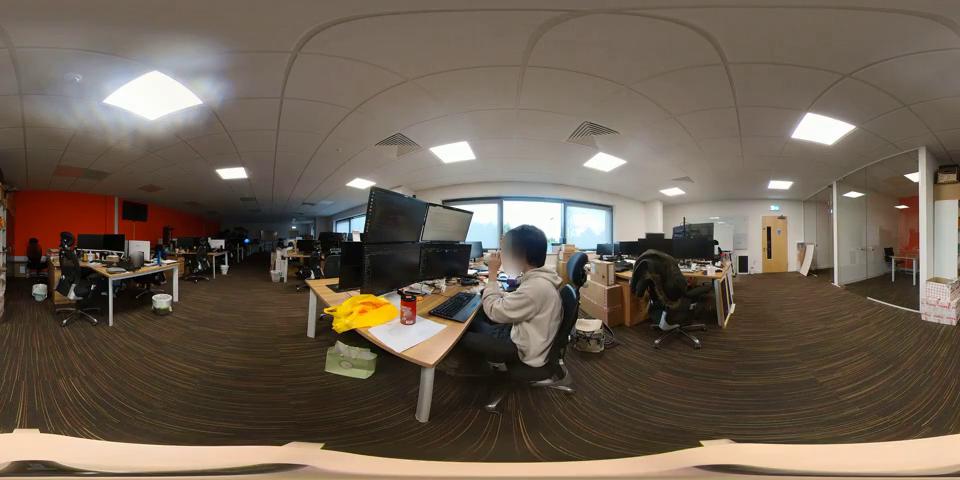}{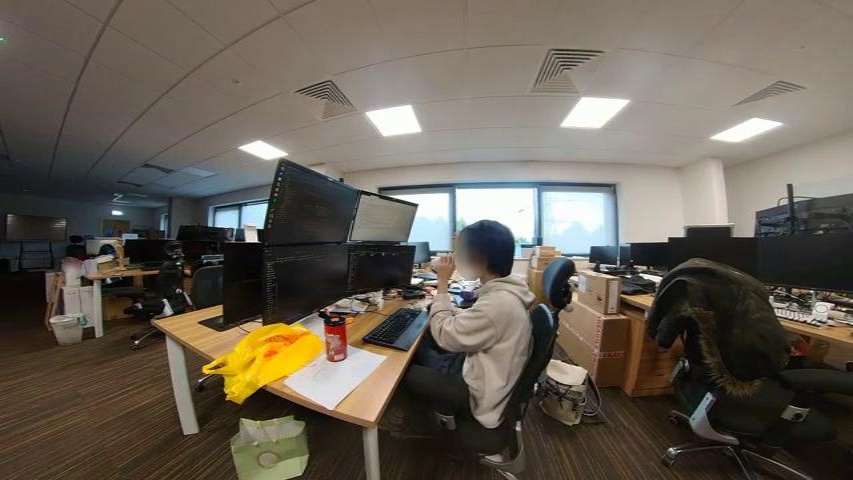}{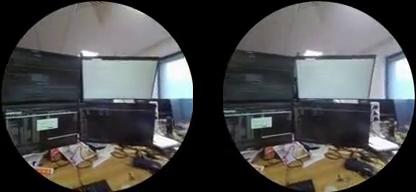}{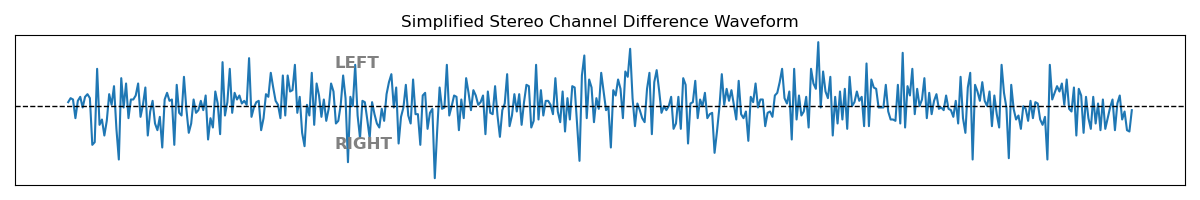}{Workspaces}

\end{document}